\documentclass[10pt,twocolumn,twoside]{IEEEtran}
\usepackage{booktabs}
\usepackage[svgnames,table]{xcolor}
\usepackage{microtype}
\usepackage{graphicx}
\usepackage{subfigure}
\usepackage{multirow}
\usepackage{booktabs} 
\usepackage{amsmath,amssymb,amsthm,epsfig,dsfont,color,graphicx,epstopdf,hhline,pgfplots}
\usepackage{algorithm,algpseudocode}
\usepackage{tabularx}

\newlength\figH
\newlength\figW

\newtheorem{definition}{Definition}
\newtheorem{theorem}{Theorem}

\usepackage{pgfplots}

\newcommand{\CC}[1]{\cellcolor{gray!#1}}

\definecolor{color1}{rgb}{1,0.498039215686275,0.0549019607843137}
\definecolor{color0}{rgb}{0.12156862745098,0.466666666666667,0.705882352941177}
\definecolor{color3}{rgb}{0.83921568627451,0.152941176470588,0.156862745098039}
\definecolor{color2}{rgb}{0.172549019607843,0.627450980392157,0.172549019607843}
\begin{document}
	\title{Adaptive Diffusions for Scalable\\ Learning over Graphs}
	\author{\textit{Dimitris Berberidis\textsuperscript{1}, Athanasios N. Nikolakopoulos\textsuperscript{2} and Georgios B. Giannakis\textsuperscript{1,2}}\\
		Dept. of ECE\textsuperscript{1} and Digital Tech. Center\textsuperscript{2}, University of Minnesota, Minneapolis, MN 55455, USA\\
		\thanks{Work was supported by NSF 171141, 1514056 and 1500713}
		\vspace*{-0.1cm}}
	
\maketitle
\begin{abstract}
Diffusion-based classifiers such as those relying on the Personalized PageRank and the Heat kernel, enjoy remarkable classification accuracy at modest computational requirements. 
Their performance however is affected by the extent to which the chosen diffusion captures a typically unknown label propagation mechanism, that can be specific to the underlying graph, and potentially different for each class. The present work introduces a disciplined, data-efficient approach to learning \textit{class-specific diffusion  functions 
adapted to the underlying network} topology. The novel learning approach leverages the notion of ``landing probabilities'' of class-specific random walks, 
which can be computed efficiently, thereby ensuring scalability to large graphs. 
This is supported by rigorous analysis of the properties of the model as well as the proposed algorithms.   
Furthermore, a robust version of the classifier facilitates learning even in noisy environments.    
	Classification tests on real networks demonstrate that adapting the diffusion function to the given graph and observed labels, significantly improves the performance over fixed diffusions; reaching -- and many times surpassing -- the classification accuracy of computationally heavier state-of-the-art competing methods, that rely on node embeddings and deep neural networks. 
\end{abstract}

\section{Introduction}\label{sec:intro}

\IEEEPARstart{T}{he} task of classifying nodes of a graph arises frequently in several applications on real-world networks, such as the ones derived from social interactions and biological dependencies. 
Graph-based semi-supervised learning (SSL) methods tackle this task  building on the premise that the true labels are distributed ``smoothly'' with respect to the underlying network, which then motivates leveraging the network structure  to increase the classification accuracy \cite{SSL}. 
Graph-based SSL has been pursued by various intertwined methods, including iterative label propagation \cite{label-propagation-and-quadratic-criterion,zhu2003semi,liu2012robust,kveton2010semi}, kernels on graphs \cite{merkurjev2016semi}, manifold
regularization \cite{belkin2006manifold},  graph partitioning \cite{ugander2013balanced,joachims2003transductive}, transductive learning \cite{talukdar2009new}, competitive infection models \cite{rosenfeld2017semi}, and bootstrapped label propagation \cite{buchnik2017bootstrapped}. SSL based on graph filters was discussed in \cite{sandryhaila2013discrete}, and further developed in \cite{Kovac2014semi} for bridge monitoring.
Recently, approaches based on node-embeddings \cite{perozzi2014deepwalk,grover2016node2vec,yang2016revisiting}, as well as deep-learning architectures \cite{kipf2016semi,atwood2016diffusion} have gained popularity, and were reported to have state-of-the-art performance. 

Many of the aforementioned methods are challenged by computational complexity and scalability issues that limit their applicability to large-scale networks. Random-walk-based diffusions present an efficient and effective alternative. Methods of this family diffuse probabilistically  the known labels through the graph, thereby ranking nodes according to weighted sums of variable-length landing
probabilities. Celebrated representatives include those based on the Personalized PageRank (PPR) and the Heat Kernel that were found to perform remarkably well in certain application domains \cite{kloster2014heat}, and have been nicely linked to particular network models \cite{kloumann2017block,avrachenkov2012generalized,kondor2002diffusion}.  {Spectral diffusions have been used for community detection~\cite{lemon,losp}, where local diffusion patterns are produced to approximate low-conductance communities, and adaptive PPR has been applied for prediction on a heterogeneous protein-function network~\cite{aptrank}. }

The effectiveness of diffusion-based classifiers can vary considerably depending on whether the chosen diffusion conforms with  the latent label propagation mechanism that might be, (i) particular to the target application or underlying network topology; and, (ii) different for each class. The present contribution\footnote{A preliminary version of the work has appeared in~\cite{AdadifBigData}.} alleviates these shortcomings and markedly improves the performance of random-walk-based classifiers by \textit{adapting the diffusion functions of every class} to both the network and the observed labels.  The resultant novel classifier relies on the notion of landing probabilities of \textit{short-length random walks} rooted at the observed nodes of each class. The small number of these landing probabilities can be extracted efficiently with a small number of sparse matrix-vector products, thus ensuring applicability to large-scale networks.  Theoretical analysis establishes that short random walks are in most cases sufficient for reliable classification. Furthermore, an algorithm is developed to identify (and potentially remove) outlying or anomalous samples jointly with adapting the diffusions.  We test our methods in terms of multiclass and multilabel classification accuracy, and confirm that it can achieve results competitive to  state-of-the-art methods, while also being considerably faster.

The rest of the paper is organized as follows. Section \ref{sec:problem} introduces random-walk based diffusions. Our novel approach along with relevant analytical results are the subjects of Section~\ref{sec:main}.  Section~\ref{sec:robust} presents a robust version of our algorithm, and Section \ref{sec:remarks} places our work in the context of related methods. Finally, Section~\ref{sec:numerical} presents experiments, while Section~\ref{sec:conclusions} concludes the paper and discusses future directions. 

\noindent{\textbf{Notation.}} 
 Bold lower-case letters denote column vectors (e.g., $\mathbf{v}$); bold upper-case letters denote matrices (e.g., $\mathbf{Q} $). Vectors $\mathbf{q}_j$  and $\mathbf{q}^\mathsf{T}_{i}$ denote the  $j^{\textit{th}}$ column and the  $i^{\textit{th}}$ row of $\mathbf{Q}$, respectively; whereas $Q_{ij}$ (or sometimes for clarity $[\mathbf{Q}]_{ij}$) denotes the $ij^\textit{th}$ entry of $\mathbf{Q}$. Vector $\mathbf{e}_K$ denotes the $K^{\textit{th}}$ canonical column vector; and  $\lVert \cdot \rVert$ denotes the Euclidean norm, unless stated otherwise. 
 
\section{Problem Statement and Modeling}\label{sec:problem}
	
	Consider a  graph $\mathcal{G}:=\{ \mathcal{V},\mathcal{E} \}$, where $\mathcal{V}$ is the set of $N$ nodes, and $\mathcal{E}$ the set of edges. Connectivity is captured by the weight matrix $\mathbf{W}$ having entries $W_{ij} > 0 $ if $(i,j) \in \mathcal{E}$. Associated with each node $i \in {\cal V}$ there is a discrete label $y_i\in\mathcal{Y}$. In SSL classification over graphs, a subset $\mathcal{L}\subset \mathcal{V}$ of nodes has available labels $\mathbf{y}_\mathcal{L}$, and the goal is to infer the labels of the  unlabeled set $\mathcal{U}:=\mathcal{V} \setminus \mathcal{L}$.  Given a measure of influence, a node most influenced by labeled nodes of a certain class is deemed to also belong to the same class. Thus, label-propagation on graphs boils down to quantifying the influence of $\mathcal{L}$ on $\mathcal{U}$, see, e.g. \cite{SSL,kveton2010semi,wu2012learning}.   
An intuitive yet simple measure of node-to-node influence relies on the notion of random walks on graphs. 
	
 {A simple random walk on a graph is a discrete-time Markov chain defined over the nodes, meaning with state space $\mathcal{V}$.} The transition probabilities are 
\begin{align*}
\Pr\{X_k=i|X_{k-1}=j\}&=W_{ij}/d_j = [{\bf W} {\bf D}^{-1}]_{ij} := [{\bf H}]_{ij}
\end{align*}
where $X_k\in\mathcal{V}$ denotes the position of the random walker (state) at the $k^\mathit{th}$  step; $d_j:=\sum_{k\in \mathcal{N}_j}W_{kj}$ is the degree of  node $j$; and, $\mathcal{N}_j$ its neighborhood. Since we consider undirected graphs the limiting distribution of $\{X_k\}$ always exists and it is unique if it is connected and non-bipartite. It is given by the dominant right eigenvector of the column-stochastic transition probability matrix $\mathbf{H}:=\mathbf{W}\mathbf{D}^{-1}$, where $\mathbf{D}:=\mathrm{diag}\left(d_1,d_2,\ldots,d_N\right)$ \cite{levin2017markov}. The steady-state distribution $\boldsymbol{\pi}$ can be shown to have entries 
		$$\pi_i:=\lim_{k\rightarrow\infty}\sum_{j\in\mathcal{V}}\Pr\{X_k=i|X_{0}=j\}\Pr\{X_{0}=j\}=\frac{d_i}{2|\mathcal{E}|}$$
		that are clearly not dependent on the initial ``seeding'' distribution $\Pr\{X_{0}\}$; and $\boldsymbol{\pi}$ is thus unsuitable for measuring influence among nodes. Instead, for graph-based SSL, we will utilize the $k-$step \emph{landing probability} per node $i$ given by
		\begin{align}\label{land_prob}
p^{(k)}_i:=\sum_{j\in\mathcal{V}}\Pr\{X_k=i|X_{0}=j\}\Pr\{X_{0}=j\}		
		\end{align}
        that in vector form $\mathbf{p}^{(k)}:=[p^{(k)}_1~\ldots~ p^{(k)}_N]^\mathsf{T}$ satisfies $\mathbf{p}^{(k)}=\mathbf{H}^k\mathbf{p}^{(0)}$, where $p_i^{(0)}:=\Pr\{X_0=i \}$. In words, $p_i^{(k)}$ is the probability that a random walker with initial distribution $\mathbf{p}^{(0)}$ is located at node $i$ after $k$ steps. Therefore, $p_i^{(k)}$ is a valid measure of the influence that $\mathbf{p}^{(0)}$ has on any node in $\mathcal{V}$. 
        
The landing probabilities per class $c\in\mathcal{Y}$ are (cf. \eqref{land_prob})
	\begin{align}  \label{class_land_prob}
		\mathbf{p}^{(k)}_c =  \mathbf{H}^k\mathbf{v}_{c}
	\end{align}				
		where for $\mathcal{L}_c := \{i~ \in \mathcal{L} : y_i=c \}$, we select as $\mathbf{ v}_c$
			 the normalized class-indicator vector with $i-$th entry 
		\begin{align}		
		[\mathbf{v}_{c}]_i=\left\{ \begin{array}{cc}
		1/|\mathcal{L}_c|,~&~i\in\mathcal{L}_c\\
		0,~&~\mathrm{else}
		\end{array} \right. 
		\end{align}
		   acts as initial distribution. Using \eqref{class_land_prob}, we model diffusions per class $c$ over the graph driven by  $\{\mathbf{p}_c^{(k)} \}_{k=0}^K$ as  
			\begin{align}\label{general1}  
				\mathbf{f}_c(\boldsymbol{\theta}) =	\sum_{k=0}^K\theta_k\mathbf{p}^{(k)}_c 
			\end{align}				
			where $\theta_k$ denotes the importance assigned to the $k^\mathit{th}$  hop neighborhood. By setting $\theta_0=0$ (since it is not useful for classification purposes) and constraining $\boldsymbol{\theta}\in{\mathcal{S}^K},$ where $\mathcal{S}^K :=\{ \mathbf{x}\in \mathbb{R}^K :~ \mathbf{x}\geq \mathbf{0},~\mathbf{1}^\mathsf{T}\mathbf{x} =1  \}$ is the $K-$dimensional probability simplex, $\mathbf{f}_c(\boldsymbol{\theta})$ can be compactly expressed as 
				\begin{align}\label{general}  
			\mathbf{f}_c(\boldsymbol{\theta}) =	\sum_{k=1}^K\theta_k\mathbf{p}^{(k)}_c =\mathbf{P}^{(K)}_c\boldsymbol{\theta}
			\end{align}	
			where $\mathbf{P}^{(K)}_c := \begin{bmatrix}
			\mathbf{p}^{(1)}_c& \cdots & \mathbf{p}^{(K)}_c 
			\end{bmatrix}$. Note that $\mathbf{f}_c(\boldsymbol{\theta})$ denotes a valid nodal probability mass function (pmf) for class $c$.	

		Given $\boldsymbol{\theta}$ and upon obtaining $\{\mathbf{f}_c(\boldsymbol{\theta})\}_{c\in\mathcal{Y}}$, our diffusion-based classifiers will predict labels over $\mathcal{U}$ as	
	\begin{align}\label{pred}  
	\hat{y}_i(\boldsymbol{\theta}):=\arg\max_{c\in \mathcal{Y}}\left[\mathbf{f}_c(\boldsymbol{\theta})\right]_i  
	\end{align}	
	where $\left[\mathbf{f}_c(\boldsymbol{\theta})\right]_i$ is the $i^\mathit{th}$  entry of $\mathbf{f}_c(\boldsymbol{\theta})$. 
	
	The upshot of \eqref{general1} is a \emph{unifying form} of superimposed
	diffusions allowing tunable simplex weights, taking up to $K$ steps per class to come up
	with an influence metric for the semi-supervised classifier
	\eqref{pred} over graphs. Next, we outline two notable members of the family of diffusion-based classifiers that can be viewed as special cases of \eqref{general1}.
	
	\subsection{Personalized PageRank Classifier}
	 Inspired by its celebrated network centrality metric \cite{brin2012reprint}, the Personalized PageRank (PPR) algorithm has well-documented merits for label propagation; see, e.g. \cite{lin2010semi}. PPR is a special case of \eqref{general1} corresponding to
	$\boldsymbol{\theta}_{\mathrm{PPR}}= (1-\alpha) \begin{bmatrix}
	\alpha^0 & \alpha^1 & \cdots & \alpha^K 
	\end{bmatrix}^\mathsf{T}$, where $0<\alpha<1$, and $1-\alpha$ can be interpreted  as the ``restart'' probability of random walks with restarts.
	
	 The PPR-based classifier relies on (cf. \eqref{general}) {
	\begin{align}
	\mathbf{f}_c(\boldsymbol{\theta}_{\mathrm{PPR}})= (1-\alpha) \sum_{k=0}^{K}\alpha^k \mathbf{p}^{(k)}_c
	\end{align}}
	satisfying asymptotically in the number of random walk steps 
	\begin{align*}
	\lim_{K\rightarrow \infty} \mathbf{f}_c(\boldsymbol{\theta}_{\mathrm{PPR}})= (1-\alpha)(\mathbf{I} - \alpha \mathbf{H})^{-1}\mathbf{v}_c
	\end{align*}
    which implies that $\mathbf{f}_c(\boldsymbol{\theta}_{\mathrm{PPR}})$ approximates the solution of a linear system. Indeed, as shown in \cite{avrachenkov2012generalized}, PPR amounts to solving a weighted regularized least-squares problem over $\mathcal{V}$; see also \cite{kloumann2017block} for a PPR interpretation as an approximate geometric discriminant function defined in the space of landing probabilities.
	
	\subsection{Heat Kernel Classifier}
	The heat kernel (HK) is another popular diffusion that has recently been employed for SSL \cite{merkurjev2016semi} and community detection on graphs \cite{kloster2014heat}. HK is also a special case of \eqref{general1} with 
	$\boldsymbol{\theta}_{\mathrm{HK}}= e^{-t} 
	\begin{bmatrix}
	1 & t & \tfrac{t^2}{2} & \cdots & \tfrac{t^K}{K!} 
	\end{bmatrix}^\mathsf{T}$, yielding class distributions (cf. \eqref{general1}) {
	\begin{align}
	\mathbf{f}_c(\boldsymbol{\theta}_{\mathrm{HK}})= e^{-t} \sum_{k=0}^{K}\frac{t^k}{k!} \mathbf{p}^{(k)}_c. 
	\end{align}}
	Furthermore, it can be readily shown that 
	\begin{align*}
	\lim_{K\rightarrow \infty} \mathbf{f}_c(\boldsymbol{\theta}_{\mathrm{HK}})= e^{-t(\mathbf{I} -  \mathbf{H})}\mathbf{v}_c
	\end{align*}
	allowing HK to be interpreted as an approximation of a heat diffusion process, where heat is flowing from $\mathcal{L}_c$ to the rest of the graph; and $\mathbf{f}_c(\boldsymbol{\theta}_{\mathrm{HK}})$ is a snapshot of the temperature after time $t$ has elapsed. HK provably yields low conductance communities, while also converging faster to its asymptotic closed-form expression than PPR ({depending on the value of $t$})  \cite{chung2007heat}.
	
	\section{ Adaptive Diffusions}\label{sec:main}		

\begin{figure}[t]
				\centering
				\includegraphics[angle=-90, width=\linewidth]{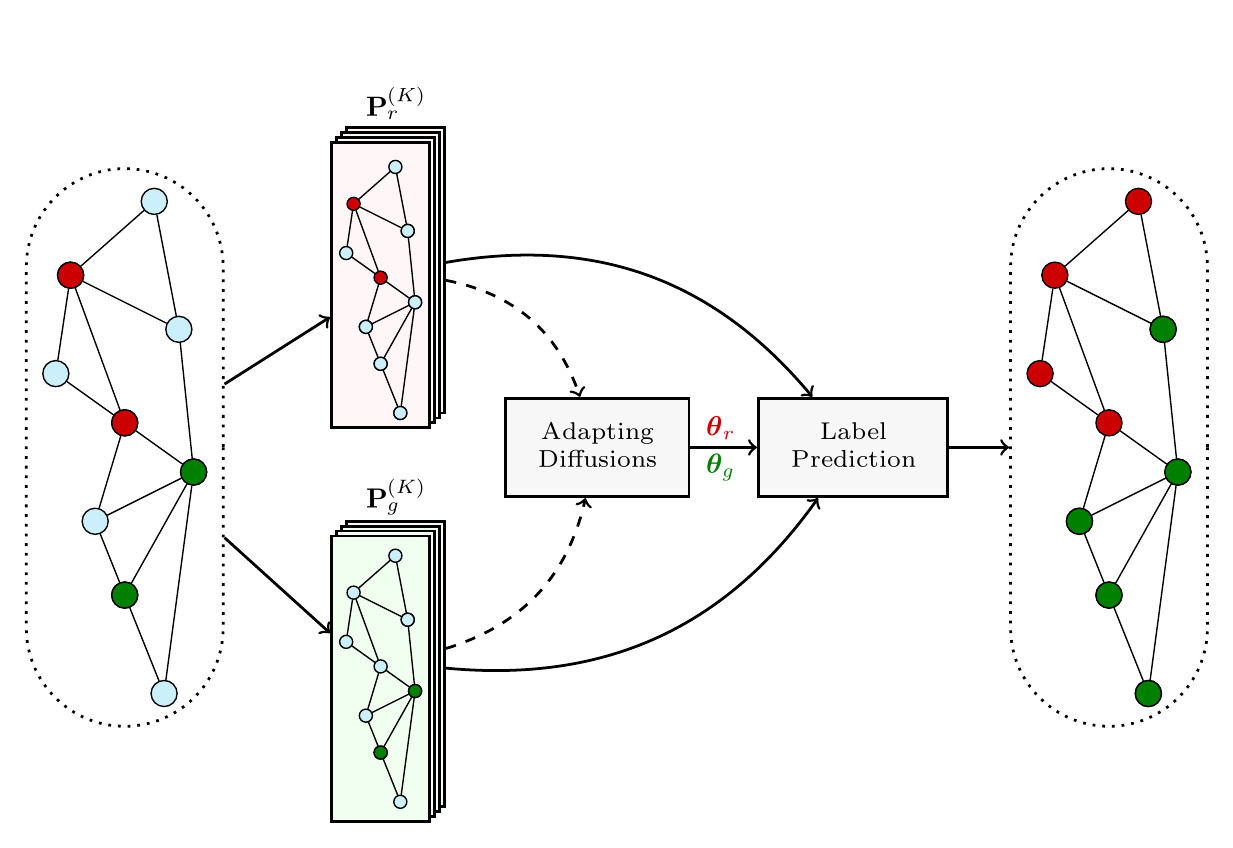}
	\caption{High-level illustration of adaptive diffusions. The nodes belong to two classes (\textcolor{red!80!black}{red} and \textcolor{green!50!black}{green}). The per-class diffusions are learned by exploiting the landing probability spaces produced by random walks rooted at the sample nodes  (second layer: \textcolor{red!80!black}{up} for red; \textcolor{green!50!black}{down} for green).}
				\label{description}
			\end{figure}

	Besides the unifying view of \eqref{general1}, the main contribution here is on efficiently  designing $\mathbf{f}_c(\boldsymbol{\theta}_c)$ in \eqref{general}, by learning the corresponding $\boldsymbol{\theta}_c$ per class. Thus, unlike PPR and HK, the methods introduced here can afford class-specific label propagation that is \emph{adaptive} to the graph structure, and also \emph{adaptive} to  the labeled nodes.  { Figure \ref{description} gives a high-level illustration of the proposed adaptive diffusion framework, where two classes (red and green) are to be diffused over the graph (cf. \eqref{class_land_prob}), with class-specific diffusion coefficients adapted as will be described next. Diffusions are then built (cf. \eqref{general}), and employed for class prediction (cf. \eqref{pred}). }
		
	Consider for generality a goodness-of-fit loss $\ell(\cdot)$, and a regularizer $R(\cdot)$ promoting e.g., smoothness over the graph. Using these, the sought class distribution will be 
	\begin{align}\label{highest_level}
	\hat{\mathbf{f}}_c= \arg \min _{\mathbf{f} \in \mathbb{R}^N} \ell (\mathbf{y}_{\mathcal{L}_c},\mathbf{f}) + \lambda R(\mathbf{f})
	\end{align}
	where $\lambda$ tunes the degree of regularization, and 
	\begin{align}\nonumber
	[\mathbf{y}_{\mathcal{L}_c}]_i=\left\{ \begin{array}{cc}
	1,~&~i\in\mathcal{L}_c\\
	0,~&~\mathrm{else}
	\end{array} \right. 
	\end{align}
	is the indicator vector of the nodes belonging to class $c$.	
	 Using our diffusion model in \eqref{general}, the $N-$dimensional optimization problem \eqref{highest_level} reduces to solving for the $K-$dimensional vector ($K\ll N$)  
	\begin{align}\label{QC_prob}
	\hat{\boldsymbol{\theta}}_c= \arg \min _{\boldsymbol{\theta}\in{\mathcal{S}^K}} \ell (\mathbf{y}_{\mathcal{L}_c},\mathbf{f}_c(\boldsymbol{\theta})) + \lambda R(\mathbf{f}_c(\boldsymbol{\theta})).
	\end{align}
	 Although many choices of $\ell(\cdot)$ may be of interest, we will focus for simplicity on the quadratic loss, namely   
	\begin{align}\nonumber
	\ell (\mathbf{y}_{\mathcal{L}_c},\mathbf{f})& := 
	\sum_{i\in\mathcal{L}}\frac{1}{d_{i}}([\bar{\mathbf{y}}_{\mathcal{L}_c}]_i - f_i )^2\\\label{fit}
	&= (\bar{\mathbf{y}}_{\mathcal{L}_c}-\mathbf{f})^\mathsf{T}\mathbf{D}^{\dagger}_{\mathcal{L}}(\bar{\mathbf{y}}_{\mathcal{L}_c}-\mathbf{f})
	\end{align}
	where $\bar{\mathbf{y}}_{\mathcal{L}_c}:=\left(1/|\mathcal{L}|\right)\mathbf{y}_{\mathcal{L}_c}$ is the class indicator vector after normalization  {to bring target variables (entries of $\bar{\mathbf{y}}_{\mathcal{L}_c}$) and entries of $\mathbf{f}$ to the same scale}, and  $\mathbf{D}^{\dagger}_{\mathcal{L}}=\mathrm{diag}(\mathbf{d}^{(-1)}_\mathcal{L})$ with entries
	\begin{align}\nonumber
	[\mathbf{d}^{(-1)}_\mathcal{L}]_i=\left\{ \begin{array}{cc}
	{1}/{d_i},~&~i\in\mathcal{L}\\
	0,~&~\mathrm{else}
	\end{array} \right. .
	\end{align}
	For a smoothness-promoting regularization, we will employ the following (normalized) Laplacian-based metric
	\begin{align}\nonumber
	R(\mathbf{f}) &= \frac{1}{2}\sum_{i\in\mathcal{V}}\sum_{j\in\mathcal{N}_i}\left(\frac{f_i}{d_{i}}-\frac{f_j}{d_{j}}\right)^2  \\\label{smooth}
	&= \mathbf{f}^\mathsf{T}\mathbf{D}^{-1}\mathbf{L}\mathbf{D}^{-1}\mathbf{f}.
	\end{align}
	 {where $\mathbf{L}:= \mathbf{D} - \mathbf{W}$ is the Laplacian matrix of the graph.} Intuitively speaking, \eqref{fit} favors vectors $\mathbf{f}$ having non-zero ($|1/|\mathcal{L}|$) values on nodes that are known to belong to class $c$, and zero values on nodes that are known to belong to other classes ($\mathcal{L}\setminus \mathcal{L}_c$), while \eqref{smooth} promotes similarity of the entries of $\mathbf{f}$ that correspond to neighboring nodes. In \eqref{fit} and \eqref{smooth}, each entry $f_i$ is normalized by $d_i^{-\frac{1}{2}}$ and $d_i^{-1}$ respectively. This normalization counterbalances the tendency of random walks to concentrate on high-degree nodes, thus placing equal importance to all nodes.
	
Substituting \eqref{fit} and \eqref{smooth} into \eqref{QC_prob}, and recalling from \eqref{general} that $\mathbf{f}_c (\boldsymbol{\theta}) = \mathbf{P}_c^{(K)} \boldsymbol{\theta}$,
	 yields the convex quadratic program  
	\begin{align} \label{high_level_prob}
	\hat{\boldsymbol{\theta}}_c= \arg \min   _{\boldsymbol{\theta}\in{\mathcal{S}^K}} \boldsymbol{\theta}^\mathsf{T} \mathbf{A}_c \boldsymbol{\theta} + \boldsymbol{\theta}^\mathsf{T} \mathbf{b}_c
	\end{align}
	with  $\mathbf{b}_c$ and $\mathbf{A}_c$ given by
	\begin{align} \label{b1}
	\mathbf{b}_c&= -\frac{2}{|\mathcal{L}|} (\mathbf{P}^{(K)}_c)^\mathsf{T} \mathbf{D}^{\dagger}_{\mathcal{L}}\mathbf{y}_{\mathcal{L}_c} \\ \label{A1_a}
	\mathbf{A}_c&= (\mathbf{P}^{(K)}_c)^\mathsf{T} \mathbf{D}^{\dagger}_{\mathcal{L}}\mathbf{P}^{(K)}_c + \lambda (\mathbf{P}^{(K)}_c)^\mathsf{T} \mathbf{D}^{-1} 
	\mathbf{L} \mathbf{D}^{-1}\mathbf{P}^{(K)}_c
	\\\nonumber
	&= (\mathbf{P}^{(K)}_c)^\mathsf{T} \left[ \left( \mathbf{D}^{\dagger}_{\mathcal{L}} + \lambda \mathbf{D}^{-1} \right)\mathbf{P}^{(K)}_c  - \lambda \mathbf{D}^{-1}  \mathbf{H}\mathbf{P}^{(K)}_c\right] \\\label{A1}
	&= (\mathbf{P}^{(K)}_c)^\mathsf{T} \left( \mathbf{D}^{\dagger}_{\mathcal{L}}\mathbf{P}^{(K)}_c + \lambda \mathbf{D}^{-1} \tilde{\mathbf{P}}^{(K)}_c  \right)
	\end{align}	
	where 
	\begin{align*}
	\mathbf{H} \mathbf{P}^{(K)}_c& = \begin{bmatrix}
	\mathbf{H} \mathbf{p}^{(1)}_c &  \mathbf{H} \mathbf{p}^{(2)}_c & \cdots & \mathbf{H} \mathbf{p}^{(K)}_c
	\end{bmatrix} \\
	& = \begin{bmatrix} \mathbf{p}^{(2)}_c &  \mathbf{p}^{(3)}_c & \cdots &  \mathbf{p}^{(K+1)}_c 	\end{bmatrix}
	\end{align*}
	is a ``shifted'' version of $\mathbf{P}^{(K)}_c$, where each $\mathbf{p}^{(k)}_c$ is advanced by one step, and
	\begin{align*}
	\tilde{\mathbf{P}}^{(K)}_c &: = \begin{bmatrix} \tilde{\mathbf{p}}^{(1)}_c & \tilde{\mathbf{p}}^{(2)}_c & \cdots &  \tilde{\mathbf{p}}^{(K)}_c \end{bmatrix}
	\end{align*}
   with $\tilde{\mathbf{p}}^{(i)}_c:= \mathbf{p}^{(i)}_c - \mathbf{p}^{(i+1)}_c$ containing the ``differential'' landing probabilities. 
The complexity of `naively' finding the $K\times K$ matrix $\mathbf{A}_c$ (and thus also $\mathbf{b}_c$) is $\mathcal{O}(K^2N)$ for computing the first summand, and $\mathcal{O}(|\mathcal{E}|K)$ for the second summand in \eqref{A1_a}, after leveraging 
the sparsity of $\mathbf{L}$, which means $|\mathcal{E}| \ll N^2$. But since columns of $\tilde{\mathbf{P}}^{(K)}_c$ are obtained as differences of consecutive columns of $\mathbf{P}^{(K)}_c$, this load of $\mathcal{O}(|\mathcal{E}|K)$ is saved. 

In a nutshell, the solver in \eqref{high_level_prob}-\eqref{A1} that we term  adaptive-diffusion (AdaDIF), incurs complexity of order $\mathcal{O}(K^2N)$. 

\noindent {\textbf{Remark 1}. The problem in \eqref{high_level_prob} is a \emph{quadratic program (QP)} of dimension $K$ (or the dictionary size $D$ to be discussed in Section III-C when in dictionary mode). In general, solving a QP with $K$ variables to a given precision requires a  $\mathcal{O}(K^3)$ worst-case complexity. Although this may appear heavy, $K$ in our setting is $10$ -- $30$ and thus negligibly small compared to the quantities that depend on the graph dimensions. For instance, the graphs that we tested have $\mathcal{O}(10^4)$ nodes ($N$) and $\mathcal{O}(10^5)$ edges ($|\mathcal{E}|$). Therefore, since $K\ll N$ and $K\ll |\mathcal{E}|$ by many orders of magnitude, the  complexity for QP is dominated by the $\mathcal{O}(|\mathcal{E}|K)$ (same as PPR and HK) performing the random walks and $\mathcal{O}(NK^2)$ for computing $\mathbf{A}_c$.}  	   
  	   
   \subsection{ Limiting behavior and computational complexity }

In this section, we offer further insights on the model \eqref{general}, along with complexity analysis of the parametric solution in \eqref{high_level_prob}. To start, the next proposition  establishes the limiting behavior of AdaDIF as the regularization parameter grows. 
\newtheorem{proposition}{Proposition}
\begin{proposition}
	If the second largest eigenvalue of $\mathbf{H}$  has multiplicity 1, then for $K$ sufficiently large but finite, the solution to \eqref{high_level_prob} as $\lambda\rightarrow\infty$ satisfies
	\begin{align}\label{prop_eq}
	\hat{\boldsymbol{\theta}}_c = \mathbf{e}_K, ~~~\forall~\mathcal{L}_c\subseteq\mathcal{V}.
	\end{align}
	\label{prop}
\end{proposition} 
Our experience with solving \eqref{high_level_prob} reveal that the 
sufficiently large $K$ required for \eqref{prop_eq} to hold, can be as small as $10^2$.  

As $\lambda \rightarrow \infty$, the effect of the loss in \eqref{QC_prob} vanishes. According to Proposition 1, this causes AdaDIF to boost smoothness by concentrating the simplex weights (entries of $\hat{\boldsymbol{\theta}}_c$) on landing probabilities of the late steps ($k$ close to $K$). If on the other extreme, smoothness-over-the-graph is not promoted (cf. $\lambda=0$), the sole objective of AdaDIF is to construct diffusions that best fit the available labeled data. Since short-length random walks from a given node typically lead to nodes of the same class, while longer walks to other classes, AdaDIF with $\lambda=0$ tends to leverage only a few landing probabilities of early steps ($k$ close to $1$). For moderate values of $\lambda$, AdaDIF effectively adapts per-class diffusions by balancing the emphasis on initial versus final landing probabilities. 

Fig. \ref{thetas} depicts an example of how AdaDIF places weights $\{\theta_k\}_{k=1}^K$ on landing probabilities after a maximum of $K=20$ steps, generated from few samples belonging to one of $7$ classes of the \texttt{Cora} citation network. Note that the learnt coefficients may follow radically different patterns than those dictated by standard \emph{non-adaptive} diffusions such as PPR or HK. It is worth noting that the simplex constraint induces sparsity of the solution in \eqref{high_level_prob}, thus `pushing' $\{\theta_k\}$ entries to zero.  

\setlength\figW{0.85\columnwidth}
\setlength\figH{0.55\columnwidth}
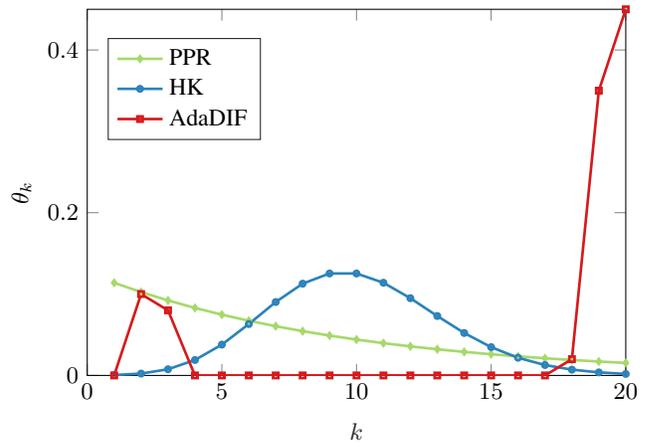
\begin{figure}[t] 
	\small
%
%

\definecolor{mycolor1}{RGB}{215,25,28}%
\definecolor{mycolor2}{RGB}{43,131,186}%
\definecolor{mycolor3}{RGB}{166,217,106}
\definecolor{mycolor4}{RGB}{253,174,97}

\begin{tikzpicture}

\begin{axis}[%
width=0.951\figW,
height=\figH,
at={(0\figW,0\figH)},
scale only axis,
xmin=0,
xmax=20,
xlabel style={font=\color{white!15!black}},
xlabel={$k$},
ymin=-0,
ymax=0.45,
ylabel style={font=\color{white!15!black}},
ylabel={$\theta_k$},
ylabel near ticks,
axis background/.style={fill=white},
legend style={at={(0.038,0.643)}, anchor=south west, legend cell align=left, align=left, draw=white!15!black}
]
\addplot [color=mycolor3, line width=1.0pt, mark size=1pt, mark=diamond, mark options={solid, mycolor3}]
  table[row sep=crcr]{%
1	0.113840325991553\\
2	0.102456293392398\\
3	0.0922106640531578\\
4	0.082989597647842\\
5	0.0746906378830578\\
6	0.067221574094752\\
7	0.0604994166852768\\
8	0.0544494750167492\\
9	0.0490045275150742\\
10	0.0441040747635668\\
11	0.0396936672872101\\
12	0.0357243005584891\\
13	0.0321518705026402\\
14	0.0289366834523762\\
15	0.0260430151071386\\
16	0.0234387135964247\\
17	0.0210948422367822\\
18	0.018985358013104\\
19	0.0170868222117936\\
20	0.0153781399906143\\
};
\addlegendentry{PPR}

\addplot [color=mycolor2, line width=1.0pt, mark size=1.0pt, mark=o, mark options={solid, mycolor2}]
  table[row sep=crcr]{%
1	0.000454742192023304\\
2	0.00227371096011652\\
3	0.00757903653372174\\
4	0.0189475913343043\\
5	0.0378951826686087\\
6	0.0631586377810145\\
7	0.0902266254014492\\
8	0.112783281751812\\
9	0.125314757502013\\
10	0.125314757502013\\
11	0.113922506820012\\
12	0.0949354223500097\\
13	0.0730272479615459\\
14	0.0521623199725328\\
15	0.0347748799816885\\
16	0.0217342999885553\\
17	0.012784882346209\\
18	0.00710271241456057\\
19	0.00373826969187398\\
20	0.00186913484593699\\
};
\addlegendentry{HK}

\addplot [color=mycolor1, line width=1.0pt, mark size=1pt, mark=square, mark options={solid, mycolor1}]
  table[row sep=crcr]{%
1	0\\
2	0.1\\
3	0.08\\
4	0\\
5	0\\
6	0\\
7	0\\
8	0\\
9	0\\
10	0\\
11	0\\
12	0\\
13	0\\
14	0\\
15	0\\
16	0\\
17	0\\
18	0.02\\
19	0.35\\
20	0.45\\
};
\addlegendentry{AdaDIF}

\end{axis}
\end{tikzpicture}%
	\caption{Illustration of $K=20$ landing probability coefficients for PPR with $\alpha=0.9$, HK with $t=10$, and AdaDIF ($\lambda = 15$).} 
	\label{thetas} 
\end{figure}

The computational core of the proposed method is to build the landing probability matrix $\mathbf{P}_c^{(K)}$, whose columns are computed fast using power iterations leveraging the sparsity of $\mathbf{H}$ (cf. \eqref{class_land_prob}). This endows AdaDIF with high  computational efficiency, especially for small $K$. Specifically, since for solving \eqref{high_level_prob} adaDIF incurs complexity $\mathcal{O}(K^2N)$ per class, 
if $K< |\mathcal{E}|/N$, this becomes $\mathcal{O}(|\mathcal{E}|K)$; and for $|\mathcal{Y}|$ classes, the overall complexity of AdaDIF is $\mathcal{O}(|\mathcal{Y}||\mathcal{E}|K)$, which is in the same order as that of non-adaptive diffusions such as PPR and HK. For larger $K$ however, an additional $\mathcal{O}(K^2N)$ is required per class, mainly to obtain $\mathbf{A}_c$ in \eqref{A1}. 

Overall, if $\mathcal{O}(KN)$ memory requirements are met, the runtime of AdaDIF scales \emph{linearly} with $|\mathcal{E}|$, provided that $K$ remains small.
Thankfully, small values of $K$ are usually sufficient to achieve high learning performance. 
As will be shown in the next section, this observation is in par with the analytical properties of diffusion based classifiers, where it turns out that $K$ large does not improve classification accuracy.    
    
 \subsection{On the choice of K}
Here we elaborate on how the selection of $K$ influences the classification 
task at hand. As expected, the effect of $K$ is intimately linked to the topology of the underlying graph, the labeled nodes, and their properties. 
For simplicity, we will focus on binary classification with the two classes denoted by $``+"$ and $``-."$ Central to our subsequent analysis is a concrete measure of the effect 
an extra landing probability vector $\mathbf{p}_c^{(k)}$ can have on the outcome of a diffusion-based classifier. Intuitively, this effect is diminishing as the number of steps $K$ grows, as both random walks eventually converge to the same stationary distribution. Motivated by this, we introduce next the $\gamma$-distinguishability threshold. 
\begin{definition}[$\gamma$-distinguishability threshold] Let $\mathbf{p}_+$ 
and $\mathbf{p}_-$ denote respectively the seed vectors for nodes of class $``+"$ and $``-,"$ 
initializing the landing probability vectors in matrices $\mathbf{X}_c := \mathbf{P}_c^{(K)}$, and $\check{\mathbf{X}}_c := \left[ \mathbf{p}_c^{(1)}  \cdots  \mathbf{p}_c^{(K-1)}  \mathbf{p}_c^{(K+1)} \right]$, where $c \in \{+,-\}$. With 
$\mathbf{y} := \mathbf{X}_+\boldsymbol{\theta} - \mathbf{X}_-\boldsymbol{\theta}$ and 
$\check{\mathbf{y}} := \check{\mathbf{X}}_+\boldsymbol{\theta} - \check{\mathbf{X}}_-\boldsymbol{\theta}$, the $\boldsymbol\gamma$-distinguishability threshold of the diffusion-based classifier is the smallest integer $K_\gamma$ satisfying  
	\begin{displaymath}
	\| \mathbf{y} - \check{\mathbf{y}}  \| \leq \gamma \;.
	\end{displaymath}
\end{definition}

The following theorem establishes an upper bound on $K_\gamma$ expressed in terms of fundamental quantities of the graph, as well as basic properties of the labeled nodes per class; see the Appendix B for a proof. 
\begin{theorem} 
For any diffusion-based classifier with coefficients $\boldsymbol{\theta}$ constrained to  a probability simplex of appropriate dimensions, the $\gamma$-distinguishability threshold is upper-bounded as 
	\begin{displaymath}
	K_\gamma \leq \frac{1}{\mu'} \log \left[ \tfrac{2\sqrt{d_{\max}}}{\gamma}  \left(\sqrt{\tfrac{1}{d_{\min_-}|\mathcal{L}_-|}}  +  \sqrt{\tfrac{1}{d_{\min_+}|\mathcal{L}_+|}} \right)   \right] 
	\end{displaymath}  
	where 
	\begin{displaymath}
	d_{\min+} := \min_{i \in \mathcal{L}_+}d_i, \quad d_{\min-} := \min_{j \in \mathcal{L}_-}d_j, \quad d_{\max} := \max_{i \in \mathcal{V}} d_i
	\end{displaymath}
	and
	\begin{displaymath}
	\mu' := \min\{\mu_2,2-\mu_N\}
	\end{displaymath} where $\{\mu_n\}_{n=1}^N$  denote the eigenvalues of the normalized graph Laplacian in ascending order. 
	\label{thm1}
\end{theorem}	
The $\gamma$-distinguishability threshold can guide the choice of the dimension $K$ of the landing probability space. Indeed, using class-specific landing probability steps $K\geq K_\gamma$, does not help distinguishing between the corresponding classes, in the sense that  the classifier output is not perturbed by more than $\gamma$. Intuitively, the information contained in the landing probabilities $ K_\gamma+1, K_\gamma+2, \dots$ is essentially the same for both classes and thus, using them as features unnecessarily increases the overall complexity of the classifier, and also ``opens the door" to curse of dimensionality related concerns.  {Note also that in settings where one can freely choose the nodes to sample, this result could be used to guide such choice in a disciplined way.}

Theorem \ref{thm1} makes no assumptions on the diffusion coefficients, so long they belong to a probability simplex. Of course, specifying the diffusion function can specialize and further tighten the corresponding $\gamma$-distinguishability threshold. In Appendix~\ref{ap:PRHR} we give a tighter threshold for PPR. 

\begin{figure}[!tbp]
  \centering
 \includegraphics[scale=1.1]{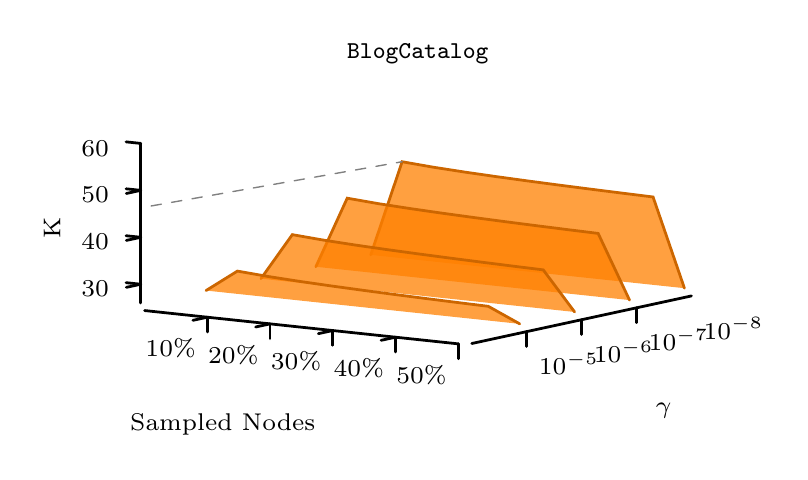}
  \caption{Experimental evaluation $K_\gamma$ for different values of $\gamma$-distinguishability threshold, and proportions of sampled nodes on \texttt{BlogCatalog} graph.}
  \label{blog_eval}
\end{figure}

Conveniently, our experiments suggest that $K\in[10,20]$ is usually sufficient to achieve high performance for most real graphs  {; see also Fig. \ref{blog_eval} where $K_\gamma$ is found numerically  for different values of $\gamma$-distinguishability threshold, and proportions of sampled nodes on the \texttt{BlogCatalog} graph}. Nevertheless, longer random walks may be necessary in e.g., graphs with small $\mu'$, especially when the number of labeled nodes is scarce.
To deal with such challenges, the ensuing modification of AdaDIF that scales linearly with $K$ is nicely motivated.

{\noindent  \textbf{Remark 2}. While PPR and HK in theory rely on infinitely long random walks, the coefficients decay rapidly ($\theta_k = \alpha^k$ for PPR and $\theta_k = t^k/k!$ for HK). This means that not only $\theta_k\rightarrow 0$ as   $k\rightarrow \infty$ in both cases, but the convergence rate is also very fast (especially for HK).  This agrees with our intuition on random walks, as well as our result in Theorem 1 suggesting that, to guarantee a  level of distinguishability (which is necessary for accuracy) between classes, classifiers should rely on relatively short-length random walks. Moreover, when operating in an adaptive framework such as the one proposed here, using finite-step (preferably short-length) landing probabilities is much more practical, since it restricts the number of free variables ($\theta_k$'s) which mitigates overfitting and results in optimization problems that scale well with the network size.} 

\subsection{ Dictionary of diffusions }
The present section deals with a modified version of AdaDIF, where the number of parameters (dimension of $\boldsymbol{\theta}$) is restricted to $D<K$, meaning the ``degrees of freedom'' of each class-specific distribution are fewer than the number of landing probabilities. Specifically, consider (cf. \eqref{general})
	\begin{align*}  
	\mathbf{f}_c(\boldsymbol{\theta}) =	\sum_{k=1}^K a_k(\boldsymbol{\theta})\mathbf{p}^{(k)}_c=\mathbf{P}^{(K)}_c\mathbf{a}(\boldsymbol{\theta})
	\end{align*}	
where $a_k(\boldsymbol{\theta}):=\sum_{d=1}^D\theta_d {C}_{kd}$, and 
$\mathbf{C}:=\begin{bmatrix} \mathbf{c}_1  \cdots\  \mathbf{c}_D
\end{bmatrix}\in\mathbb{R}^{K\times D}$ is a \emph{dictionary} of $D$ coefficient vectors, the $i^\mathit{th}$  forming the column $\mathbf{c}_i\in\mathcal{S}^K$. Since $\mathbf{a}(\boldsymbol{\theta})=\mathbf{C}\boldsymbol{\theta}$, it follows that
	\begin{align*}  
	\mathbf{f}_c(\boldsymbol{\theta}) = \mathbf{P}^{(K)}_c \mathbf{C}\boldsymbol{\theta}= \sum_{d=1}^D\theta_d \mathbf{f}_c^{(d)} 
	\end{align*}	
	where $\mathbf{f}_c^{(d)}:=\sum_{k=1}^KC_{kd}\mathbf{p}^{(k)}_c$  is the $d^\mathit{th}$  diffusion.
	
	 To find the optimal $\boldsymbol{\theta}$, the optimization problem in \eqref{high_level_prob} is solved with 
	\begin{align}\label{b2}
	\mathbf{b}_c&= -\frac{2}{|\mathcal{L}|} (\mathbf{F}^{\Delta}_c)^\mathsf{T} \mathbf{D}^{\dagger}_{\mathcal{L}}\mathbf{y}_{\mathcal{L}^c} \\\label{A2} 
	\mathbf{A}_c&= (\mathbf{F}^{\Delta}_c)^\mathsf{T} \mathbf{D}^{\dagger}_{\mathcal{L}}\mathbf{F}^{\Delta}_c + \lambda (\mathbf{F}^{\Delta}_c)^\mathsf{T} \mathbf{D}^{-1} \mathbf{L} \mathbf{D}^{-1}\mathbf{F}^{\Delta}_c
	\end{align}	
  where $\mathbf{F}^{\Delta}_c := [\mathbf{f}_c^{(1)} \cdots \ \mathbf{f}_c^{(D)} ]$ effectively replaces $\mathbf{P}^{(K)}_c$ as the basis of the space on which each $\mathbf{f}_c$ is constructed. The description of AdaDIF in \emph{dictionary mode} is given as a special case of Algorithm 1, together with the subroutine in Algorithm 2 for memory-efficient generation of $\mathbf{F}^{\Delta}_c$.

The motivation behind this dictionary-based variant of AdaDIF is two-fold. First, it leverages the properties of a judiciously selected basis of known diffusions, e.g. by constructing $\mathbf{C}=\begin{bmatrix}
\boldsymbol{\theta}_{\mathrm{PPR}} & \boldsymbol{\theta}_{\mathrm{HK}} & \cdots
\end{bmatrix}$. In that sense, our approach is related to multi-kernel methods, e.g. \cite{argyriou2006combining}, although significantly more scalable than the latter. Second, the complexity of AdaDIF in dictionary mode is $\mathcal{O}(|\mathcal{E}|(K+D))$, where $D$ can be arbitrarily smaller than $K$, leading to complexity that is \emph{linear} with respect to both $K$ and $|\mathcal{E}|$.
    	\begin{algorithm}[t]
	\caption{\textsc{Adaptive Diffusions}}
	\label{alg:AdaDIF}
	\begin{algorithmic}
		\State \textbf{Input:} Adjacency matrix: $\mathbf{W}$, Labeled nodes: $\{y_i \}_{i\in\mathcal{L}}$ \\{\bfseries \textit{parameters}:} Regularization parameter: $\lambda$, $\#$ of landing probabilities: $K$, Dictionary mode $\in\{\mathrm{True, False} \}  $, Unconstrained $\in\{\mathrm{True, False} \}$
		\State \textbf{Output:} Predictions: $\{\hat{y}_i \}_{i\in\mathcal{U}}$
		\State Extract $\mathcal{Y}=\{$  Set of unique labels in: $\{y_i \}_{i\in\mathcal{L}} \}$
		\For{$c \in  \mathcal{Y}$}
		\State $\mathcal{L}_c = \{i~ \in \mathcal{L} : y_i=c \}$
		\If{ Dictionary mode }
		\State $\mathbf{F}^{\Delta}_c = \textsc{Dictionary }(\mathbf{W},\mathcal{L}_c,K, \mathbf{C} )  $
		\State Obtain $\mathbf{b}_c$ and $\mathbf{A}_c$ as in \eqref{b2} and \eqref{A2}
		\State $\mathbf{F}_c = \mathbf{F}_c^{\Delta}$				
		\Else  
		\State $\{\mathbf{P}_c^{(K)}, \tilde{\mathbf{P}}_c^{(K)}\} = \textsc{LandProb}(\mathbf{W},\mathcal{L}_c,K )$
		\State Obtain $\mathbf{b}_c$ and $\mathbf{A}_c$ as in \eqref{b1} and \eqref{A1}
		\State $\mathbf{F}_c = \mathbf{P}_c^{(K)}$
		\EndIf 
		\If{Unconstrained }
		\State Obtain $\hat{\boldsymbol{\theta}}_c$ as in \eqref{unc1} and \eqref{unc2}
		\Else
		\State Obtain $\hat{\boldsymbol{\theta}}_c$ by solving \eqref{high_level_prob}
		\EndIf		    
		\State $\mathbf{f}_c(\hat{\boldsymbol{\theta}}_c)= \mathbf{F}_c \hat{\boldsymbol{\theta}}_c $
		\EndFor
		\State Obtain $\hat{y}_i=\arg\max_{c\in\mathcal{Y}}\left[\mathbf{f}_c(\hat{\boldsymbol{\theta}}_c)\right]_i,~~\forall i \in \mathcal{U}$	
	\end{algorithmic}
\end{algorithm}

\begin{algorithm}[t]
	\caption{\textsc{LandProb}}
	\label{alg:LPs}
	\begin{algorithmic}
		\State \textbf{Input:} $\mathbf{W},\mathcal{L}_c,K$
		\State \textbf{Output:} $\mathbf{P}^{(K)}_c,~~\tilde{\mathbf{P}}^{(K)}_c$
		\State $\mathbf{H}=\mathbf{WD}^{-1}$
		\State $\mathbf{p}_c^{(0)}=\mathbf{v}_c$
		\For { $k=1:K+1$ }				
		\State $\mathbf{p}_c^{(k)}=\mathbf{H}\mathbf{p}_c^{(k-1)}$
		\State $\tilde{\mathbf{p}}^{(k)}_c= \mathbf{p}^{(k-1)}_c - \mathbf{p}^{(k)}_c$
		\EndFor				
	\end{algorithmic}
\end{algorithm}

\begin{algorithm}[h!]
	\caption{\textsc{Dictionary}}
	\label{alg:DIC}
	\begin{algorithmic}
		\State \textbf{Input:} $\mathbf{W},\mathcal{L}_c,K,\mathbf{C}$
		\State \textbf{Output:} $\mathbf{F}_c^{\Delta}$
		\State $\mathbf{H}=\mathbf{WD}^{-1}$
		\State $\mathbf{p}_c^{(0)}=\mathbf{v}_c$
		\State $\{\mathbf{f}_c^{(d)}\}_{d=1}^D=\mathbf{0} $
		\For { $k=1:K$ }
		\State $\mathbf{p}_c^{(k)}=\mathbf{H}\mathbf{p}_c^{(k-1)}$
		\For { $d=1:D$ }
		\State  $\mathbf{f}_c^{(d)}=\mathbf{f}_c^{(d)} + C_{kd}\mathbf{p}_c^{(k)} $						
		\EndFor						
		\EndFor
	\end{algorithmic}
\end{algorithm}

\subsection {Unconstrained diffusions}
Thus far, the diffusion coefficients $\boldsymbol{\theta}$ have been constrained on the $K-$dimensional probability simplex $\mathcal{S}^K$, resulting in sparse solutions $\hat{\boldsymbol{\theta}}_c$, as well as $\mathbf{f}_c(\hat{\boldsymbol{\theta}}_c) \in \mathcal{S}^N$. The latter also allows each $\mathbf{f}_c(\boldsymbol{\theta})$ to be interpreted as a pmf over $\mathcal{V}$. Nevertheless, the simplex constraint imposes a limitation to the model: landing probabilities may only have \emph{non-negative} contribution on the resulting class distribution.  { Upon relaxing this non-negativity constraint, \eqref{high_level_prob} can afford 
a closed-form solution as 
	\begin{align}\label{unc1}
	\hat{\boldsymbol{\theta}}_c&= \mathbf{A}_c^{-1}(\mathbf{b}_c - \lambda^{\ast}\mathbf{1}) \\ \label{unc2}
	\lambda^{\ast}&=\frac{\mathbf{1}^\mathsf{T}\mathbf{A}_c^{-1}\mathbf{b}_c-1}{\mathbf{b}^\mathsf{T}\mathbf{A}_c^{-1}\mathbf{b}_c}. 
	\end{align}	
Retaining the hyperplane constraint $\mathbf{1}^\mathsf{T}\boldsymbol{\theta}=1$  forces at least one entry of $\boldsymbol{\theta}$ to be positive. Note that for $K>|\mathcal{L}|$, \eqref{unc1} may become ill-conditioned, and yield inaccurate solutions.} This can be mitigated by imposing $\ell_2-$norm regularization on $\boldsymbol{\theta}$, which is equivalent to adding $\epsilon\mathbf{I}$ to  $\mathbf{A}_c$, with $\epsilon>0$ sufficiently large to stabilize the linear system.

A step-by-step description of the proposed AdaDIF approach is given by Algorithm 1, along with the subroutine in Algorithm 2. Determining the limiting behavior of unconstrained AdaDIF, as well as exploring the effectiveness of different regularizers (e.g., sparsity inducing $\ell_1-$norm) is part of our ongoing research. Towards the goal of developing more robust methods to design diffusions, the ensuing section presents our proposed approach that relies on minimizing the leave-one-out loss of the resulting classifier.   

\section{Adaptive Diffusions Robust to Anomalies}\label{sec:robust}

Although the loss function in \eqref{fit} is simple and easy to implement, it may lack robustness against nodes with labels that do not comply with a diffusion-based information propagation model. In real-world graphs, such `difficult' nodes may arise due to model limitations, observation noise, or even deliberate mislabeling by adversaries. For such setups, this section introduces a novel adaptive diffusion classifier with: i) robustness in finding 
$\boldsymbol{\theta}$ by ignoring errors that arise due to outlying/anomalous nodes; as well as, ii) capability to identify and remove such `difficult' nodes.  

Let us begin by defining the following per-class $c\in\mathcal{Y}$ loss 
\begin{align}\label{rob-loss}
\ell_\mathrm{rob}^c (\mathbf{y}_{\mathcal{L}_c},\boldsymbol{\theta}):= 
\sum_{i\in\mathcal{L}}\frac{1}{d_{i}}\left([\bar{\mathbf{y}}_{\mathcal{L}_c}]_i - \left[\mathbf{f}_c({\boldsymbol{\theta}};\mathcal{L}\setminus i)\right]_i \right)^2
\end{align}
where $\mathbf{f}_c({\boldsymbol{\theta}};\mathcal{L}\setminus i)$ is the class-$c$ diffusion after removing the $i^\mathit{th}$ node from the set of all labels. Intuitively, \eqref{rob-loss} evaluates the ability of a propagation mechanism effected by $\boldsymbol{\theta}$ to predict the presence of class $c$ label on each node $i\in\mathcal{L}$, using the remaining labeled nodes $\mathcal{L}\setminus i$. Since each class-specific distribution $\mathbf{f}_c({\boldsymbol{\theta}})$ is constructed by random walks that are rooted in $\mathcal{L}_c\subseteq \mathcal{L}$, it follows that  
\begin{align}		
\mathbf{f}_c({\boldsymbol{\theta}};\mathcal{L} \setminus i)=\left\{ \begin{array}{cc}
\mathbf{f}_c({\boldsymbol{\theta}}),~&~i\notin\mathcal{L}_c\\
\mathbf{f}_c({\boldsymbol{\theta}};\mathcal{L}_c \setminus i),~&~i\in\mathcal{L}_c
\end{array} \right. 
\end{align}
since $\mathbf{f}_c({\boldsymbol{\theta}})$ is not directly affected by the removal of a label that belongs to other classes, and it is not used as a class-$c$ seed. The class-$c$ diffusion upon removing the $i^\mathit{th}$ node from the seeds $\mathcal{L}_c$ is given as (cf. \eqref{general})
\begin{align}\nonumber
 \mathbf{f}_c({\boldsymbol{\theta}};\mathcal{L}_c \setminus i) = \sum_{k=1}^K\theta_k\mathbf{p}^{(k)}_{\mathcal{L}_{c}\setminus i}
\end{align}
where $\mathbf{p}^{(k)}_{\mathcal{L}_{c}\setminus i} := \mathbf{H}^k\mathbf{v}_{\mathcal{L}_{c}\setminus i}$, and
\begin{align}		
[\mathbf{v}_{\mathcal{L}_{c}\setminus i}]_j=\left\{ \begin{array}{cc}
1/|\mathcal{L}_c\setminus i|,~&~j\in\mathcal{L}_c\setminus i\\
0,~&~\mathrm{else}
\end{array} \right. .
\end{align}
The robust loss in \eqref{rob-loss} can be expressed more compactly as
\begin{align}\label{l_rob}
\ell_\mathrm{rob}^c (\mathbf{y}_{\mathcal{L}_c},\boldsymbol{\theta}):= \|\mathbf{D}_{\mathcal{L}}^{-\frac{1}{2}}\left( \bar{\mathbf{y}}_{\mathcal{L}_c} - \mathbf{R}_c^{(K)}\boldsymbol{\theta}\right)\|_2^2
\end{align}
where  $\mathbf{D}_{\mathcal{L}}^{-\frac{1}{2}}:=\left({\mathbf{D}_{\mathcal{L}}^{\dagger}}\right)^{-\frac{1}{2}}$, and
\begin{align}		\label{R}
\left[\mathbf{R}_c^{(K)}\right]_{ik} := \left\{ \begin{array}{cc}
\left[\mathbf{p}_{\mathcal{L}_c\setminus i}^{(k)}\right]_i,~&~i\in\mathcal{L}_c\\[1em]
\left[\mathbf{p}_{c}^{(k)}\right]_{i},~&~\mathrm{else}
\end{array} \right. .
\end{align}
Since  $\mathbf{p}_{c}^{(k)}=|\mathcal{L}_c|^{-1}\sum_{i\in\mathcal{L}_c}\mathbf{p}_{\mathcal{L}_c\setminus i}^{(k)}$, evaluating \eqref{l_rob} only requires the rows of $\mathbf{R}_c^{(K)}$ and entries of $\mathbf{y}_{\mathcal{L}_c}$ that correspond to $\mathcal{L}$, since the rest of the diagonal entries of $\mathbf{D}_{\mathcal{L}}^\dagger$ are $0$. Having defined $\ell_{\mathrm{rob}}^c(\cdot)$, per-class diffusion coefficients $\hat{\boldsymbol{\theta}}_c$ can be obtained by solving 
\begin{align}\label{QC_prob2}
\hat{\boldsymbol{\theta}}_c= \arg \min _{\boldsymbol{\theta}\in{\mathcal{S}^K}} \ell_\mathrm{rob}^c (\mathbf{y}_{\mathcal{L}_c},\boldsymbol{\theta}) + \lambda_{\theta} \|\boldsymbol{\theta}\|_2^2
\end{align}
where $\ell_2$ regularization with parameter $\lambda_\theta$ is introduced in order to prevent overfitting and numerical instabilities.  {Note that smoothness regularization in \eqref{smooth} is less appropriate in the context of robustness, since it promotes ``spreading'' of the random walks (cf. Prop. 1), thus making class-diffusions more similar and increasing the difficulty of detecting outliers.} Similar to \eqref{high_level_prob}, quadratic programming can be adopted to solve \eqref{QC_prob2}. 

Towards mitigating the effects of outliers, and inspired by the robust estimators introduced in \cite{keka2011sparse}, we further enhance $\ell_{\mathrm{rob}}^c(\cdot)$ by explicitly modeling the effect of outliers with a sparse vector $\mathbf{o}\in\mathbb{R}^N$, leading to the modified cost
\begin{align}\label{l_rob2}
\ell_\mathrm{rob}^c (\mathbf{y}_{\mathcal{L}_c},\mathbf{o},\boldsymbol{\theta}):= \|\mathbf{D}_{\mathcal{L}}^{-\frac{1}{2}}\left( \mathbf{o} + \bar{\mathbf{y}}_{\mathcal{L}_c} - \mathbf{R}_c^{(K)}\boldsymbol{\theta}\right)\|_2^2.
\end{align}
The non-zero entries of ${\mathbf{o}}$ can capture large residuals (prediction errors $|[\bar{\mathbf{y}}_{\mathcal{L}_c}]_i - \left[\mathbf{f}_c({\boldsymbol{\theta}};\mathcal{L}\setminus i)\right]_i|$) which may be the result of outlying, anomalous or mislabeled nodes.
Thus, when operating in the presence of anomalies, the robust classifier aims at identifying both diffusion parameters $\{\hat{\boldsymbol{\theta}}_c\}_{c\in\mathcal{Y}}$ as well as per class outlier vectors $\{\hat{\mathbf{o}}_c\}_{c\in\mathcal{Y}}$. The two tasks can be performed jointly by solving the following optimization problem 
\begin{align}\nonumber
\{\hat{\boldsymbol{\theta}}_c,\hat{\mathbf{o}}_c\}_{c\in\mathcal{Y}}= \arg \min _{\overset{\boldsymbol{\theta}_c\in{\mathcal{S}^K}}{\mathbf{o}_c\in\mathbb{R}^{N}}} \sum_{{c\in\mathcal{Y}}}&\left[\ell_\mathrm{rob}^c (\mathbf{y}_{\mathcal{L}_c},\mathbf{o}_c,\boldsymbol{\theta}_c) + \lambda_{\theta} \|\boldsymbol{\theta}_c\|_2^2\right]\\\label{QC_prob3}
& + \lambda_o\|\mathbf{D}_{\mathcal{L}}^{-\frac{1}{2}}\mathbf{O}\|_{2,1}
\end{align}
where $\mathbf{O} :=\begin{bmatrix}
\mathbf{o}_1 & \cdots & \mathbf{o}_{|\mathcal{Y}|}
\end{bmatrix}$  concatenates the outlier vectors $\mathbf{o}_c$,  and $\|\mathbf{X}\|_{2,1}:=\sum_{i=1}^I\sqrt{\sum_{j=1}^J X_{i,j}^2}$ for any $\mathbf{X}\in\mathbb{R}^{I\times J}$. The term  $\lambda_o\|\mathbf{D}_{\mathcal{L}}^{-\frac{1}{2}}\mathbf{O}\|_{2,1}$ in \eqref{QC_prob3} acts as a regularizer that promotes sparsity over the rows of $\mathbf{O}$; it can also be interpreted as an $\ell_1$-norm regularizer over a vector that contains the $\ell_2$ norms of the rows of $\mathbf{O}$. The reason for using such block-sparse regularization is to force outlier vectors $\mathbf{o}_c$ of different classes to have the same support (pattern of non-zero entries). In other words, the $|\mathcal{Y}|$ different diffusion/outlier detectors are \emph{forced} to consent on which nodes are outliers. 

Since \eqref{QC_prob3} is non-convex, convergence of gradient-descent-type methods to the global optimum is not guaranteed. Nevertheless, since \eqref{QC_prob3} is biconvex (i.e., convex with respect to each variable) the following alternating minimization scheme 
\begin{align}\nonumber
\hspace*{-0.2cm}
\hat{\mathbf{O}}^{(t)} 
\hspace*{-0.2cm}
=& \arg \min _{\mathbf{O}} \sum_{{c\in\mathcal{Y}}}\left[\ell_\mathrm{rob}^c (\mathbf{y}_{\mathcal{L}_c},\mathbf{o}_c,\hat{\boldsymbol{\theta}}_c^{(t-1)}) + \lambda_{\theta} \|\hat{\boldsymbol{\theta}}_c^{(t-1)}\|_2^2\right]\\\label{AM_1}
& \qquad \qquad \qquad + \lambda_o\|\mathbf{D}_{\mathcal{L}}^{-\frac{1}{2}}\mathbf{O}\|_{2,1}\\ \label{AM_2}
\hat{\boldsymbol{\theta}}_c^{(t)}=& \arg \min _{\boldsymbol{\theta}\in{\mathcal{S}^K}} \ell_\mathrm{rob}^c (\mathbf{y}_{\mathcal{L}_c},\hat{\mathbf{o}}_c^{(t)},\boldsymbol{\theta}) 
\hspace*{-0.1cm} + \hspace*{-0.1cm} \lambda_{\theta} \|\boldsymbol{\theta}\|_2^2 
\hspace*{-0.1cm} + \hspace*{-0.1cm}  \lambda_o\|\mathbf{D}_{\mathcal{L}}^{-\frac{1}{2}}\hat{\mathbf{O}}^{(t)}\|_{2,1}
\end{align}
with $\hat{\mathbf{O}}^{(0)} := \left[\mathbf{0} \ldots \mathbf{0}\right]$ converges to a partial optimum  \cite{gorski2007biconvex}. 

By further simplifying \eqref{AM_2} and solving \eqref{AM_1} in closed form, we obtain
\begin{align}\label{AM_3}
\hat{\boldsymbol{\theta}}_c^{(t)}=& \arg \min _{\boldsymbol{\theta}\in{\mathcal{S}^K}} \ell_\mathrm{rob}^c (\bar{\mathbf{y}}_{\mathcal{L}_c}+\hat{\mathbf{o}}_c^{(t-1)},\boldsymbol{\theta}) + \lambda_{\theta} \|\boldsymbol{\theta}\|_2^2\\\label{AM_4}
\hat{\mathbf{O}}^{(t)}=&\operatorname{SoftThres}_{\lambda_o}\left( \tilde{\mathbf{Y}}^{(t)} \right)
\end{align}
where 
$$\tilde{\mathbf{Y}}^{(t)}:=\left[ \tilde{\mathbf{y}_1}^{(t)},\ldots,\mathbf{y}_{|\mathcal{Y}|}^{(t)}\right]$$ is the matrix that concatenates the per class residual vectors $\tilde{\mathbf{y}}_c^{(t)}:=\bar{\mathbf{y}}_{\mathcal{L}_c} -\mathbf{R}_c^{(K)}\hat{\boldsymbol{\theta}}_c^{(t)}$, and   $\mathbf{Z}=\mathrm{SoftThres}_{\lambda_o}(\mathbf{X})$ is a row-wise soft-thresholding operator such that 
$$\mathbf{z}_i=\|\mathbf{x}_i\|_2[1-\lambda_o/(2\|\mathbf{x}_i\|_2)]_+$$ where $\mathbf{z}_i$ and $\mathbf{x}_i$ are the $i^\mathit{th}$  rows of $\mathbf{Z}$ and $\mathbf{X}$ respectively, see e.g. \cite{puig2011multidimensional}. Intuitively, the soft-thresholding operation in \eqref{AM_4} extracts the outliers by scaling down residuals and ``trimming'' them wherever their across-classes $\ell_2$ norm is below a certain threshold.  

The alternating minimization between \eqref{AM_3} and \eqref{AM_4} terminates when 
\begin{displaymath}
\|\hat{\boldsymbol{\theta}}_c^{(t)}-\hat{\boldsymbol{\theta}}_c^{(t-1)}\|_{\infty}\leq \epsilon,~\forall c\in\mathcal{Y}
\end{displaymath}
 where $\epsilon\geq 0$ is a prescribed tolerance. Having obtained the tuples $\{\hat{\boldsymbol{\theta}}_c,\hat{\mathbf{o}}_c\}_{c\in\mathcal{Y}}$, one may remove the anomalous samples that correspond to non-zero rows of $\hat{\mathbf{O}}$ and perform the diffusion with the remaining samples. The robust (r) AdaDIF is summarized as Algorithm \ref{alg:AdaDIF-rob}, and has $\mathcal{O}(K|\mathcal{L}||\mathcal{E}|)$ computational complexity. 
	  	\begin{algorithm}[t]
	\caption{\textsc{Robust Adaptive Diffusions}}
	\label{alg:AdaDIF-rob}
	\begin{algorithmic}
		\State \textbf{Input:} Adjacency matrix: $\mathbf{W}$, Labeled nodes: $\{y_i \}_{i\in\mathcal{L}}$ \\{\bfseries \textit{parameters}:} Regularization parameters: $\lambda_\theta, \lambda_o$, $\#$ of landing probabilities: $K$
		\State \textbf{Output:} Predictions: $\{\hat{y}_i \}_{i\in\mathcal{U}}$
		\State ~~~~~~~~~~ Outliers: $\underset{{c\in\mathcal{Y}}}{\cup}\mathcal{L}_c^o$
		\State Extract $\mathcal{Y}=\{$  Set of unique labels in: $\{y_i \}_{i\in\mathcal{L}} \}$
		\For{$c \in  \mathcal{Y}$}
		\State $\mathcal{L}_c = \{i~ \in \mathcal{L} : y_i=c \}$
		\For{$i\in\mathcal{L}_c$} 
		\State $\{\mathbf{p}_{\mathcal{L}_c\setminus i}^{(k)}\}_{k=1}^{K} = \textsc{LandProb}(\mathbf{W},\mathcal{L}_c\setminus i,K )$
		\EndFor
		\State Obtain $\mathbf{R}_c^{(K)}$ as in \eqref{R}
		\EndFor
		\State $\hat{\mathbf{O}}^{(0)}=\left[\mathbf{0},\ldots,\mathbf{0}\right],t=0$
		\While{$\|\hat{\boldsymbol{\theta}}_c^{(t)}-\hat{\boldsymbol{\theta}}_c^{(t-1)}\|_{\infty}\leq \epsilon$}
		\State $t\leftarrow t+1$
		\State Obtain $\{\hat{\boldsymbol{\theta}}_c^{(t)}\}_{c\in\mathcal{Y}}$ as in \eqref{AM_3}
		\State Obtain $\hat{\mathbf{O}}^{(t)}$ as in \eqref{AM_4}
		\EndWhile
        \State Set of outliers: $\mathcal{S}:=\{i\in\mathcal{L}:\|[\hat{\mathbf{O}}]_{i,:}\|_2 >0\}$
        \For{$c \in  \mathcal{Y}$} 
		\State $\mathcal{L}_c^o = \mathcal{L}_c\cap\mathcal{S} $		
		\State $\mathcal{L}_c\leftarrow \mathcal{L}_c\setminus \mathcal{L}_c^o$
		\EndFor
		\State Obtain $\hat{y}_i=\arg\max_{c\in\mathcal{Y}}\left[\mathbf{f}_c(\hat{\boldsymbol{\theta}}_c)\right]_i,~~\forall i \in \mathcal{U}$	
	\end{algorithmic}
\end{algorithm}

\section{Contributions in Context of Prior Works}\label{sec:remarks}
Following the seminal contribution in~\cite{brin2012reprint} that introduced PageRank as a network centrality measure, there has been a vast body of works studying its theoretical properties, computational aspects, as well as applications beyond Web ranking \cite{LangvilleMeyer,GleichBeyond}. Most formal approaches to generalize PageRank focus either on the \textit{teleportation} component (see e.g.~\cite{Nikolakopoulos:2013:NNR:2433396.2433415,7840666} as well as~\cite{berberidis2018random} for an application to semi-supervised classification), 
or, on the so-termed \textit{damping} mechanism~\cite{constantine2009,FunctionalRankings}. Perhaps the most general setting can be found in~\cite{FunctionalRankings}, where a family of functional rankings was introduced by the choice of a parametric damping function that assigns weights to successive steps of a walk initialized according to the teleportation distribution. The per class distributions produced by AdaDIF are in fact members of this family of functional rankings. However, instead of choosing a fixed damping function as in the aforementioned approaches, AdaDIF learns a class-specific and graph-aware damping mechanism. In this sense, AdaDIF undertakes  statistical learning in the space of functional rankings, tailored to the underlying semi-supervised classification task.   { A related method termed AptRank was recently proposed in \cite{aptrank} specifically for protein function prediction. {Differently from AdaDIF the method exploits meta-information regarding the hierarchical organization of functional roles of proteins and it performs random walks on the heterogeneous protein-function network.  AptRank splits the data into training and validation sets of predetermined proportions and adopt as cross-validation approach for obtaining diffusion coefficients. Furthermore 
	a1) 
	AptRank trains a single diffusion for all classes whereas AdaDIF identifies different diffusions, and a2) 
	the proposed robust leave-one-out method (r-AdaDIF) gathers residuals from all leave-one-out splits into one cost function (cf. \eqref{rob-loss}) and then optimizes the (per class) diffusion.}}

{ Recently, community detection (CD) methods were proposed in \cite{lemon} and \cite{losp}, that search the Krylov subspace of landing probabilities of a given community's seeds, to identify a diffusion that satisfies \emph{locality} of non-zero entries over the nodes of the graph. In CD, the problem  definition is: ``given certain members of a community, identify the remaining (latent) members.'' There is a subtle but important distinction between CD and semi-supervised classification (SSC): CD focuses on the retrieval of \emph{communities} (that is nodes of a given class), whereas SSC focuses on the predicting the labels/attributes of every \emph{node}. 
While CD treats the detection of various overlapping communities of the graph as independent tasks, SSC classifies nodes by taking all information from labeled nodes into account. More specifically, the proposed AdaDIF trains the diffusion of each class 
by actively avoiding the assignment of large diffusion values to nodes that are known (they have been labeled) to belong to a different class. Another important difference of AdaDIF with \cite{lemon} and \cite{losp}---which again arises from the different contexts---is the length of the walk compared to the size of the graph. Since \cite{lemon} and \cite{losp} aim at identifying small and local communities, they perform local walks of length smaller than the diameter of the graph. In contrast, SSC typically demands a certain degree of globality in information exchange, achieved by longer random walks that surpass the graph diameter.
}

AdaDIF also shares links with SSL methods based on graph signal processing proposed in \cite{sandryhaila2013discrete}, and further pursued in \cite{Kovac2014semi} for bridge monitoring; see also \cite{segarra2017filter} and \cite{leus2017filter} for recent advances on graph filters. Similar to our approach, these graph filter based techniques are parametrized via assigning different weights to a number of consecutive powers of a matrix related to the structure of the graph. Our contribution however, introduces different loss and regularization functions for adapting the diffusions, including a novel approach for training the model in an anomaly/outlier-resilient manner. Furthermore, while \cite{sandryhaila2013discrete} focuses on binary classification and \cite{Kovac2014semi} identifies a single model for all classes, our  approach allows for different classes to have different propagation mechanisms. This feature can accommodate  differences in the label distribution of each class over the nodes, while also making AdaDIF readily applicable to multi-label graphs. Moreover, while in \cite{sandryhaila2013discrete} the weighting parameters remain  unconstrained and in \cite{Kovac2014semi} belong to a hyperplane, AdaDIF constrains the diffusion parameters on the probability simplex, which allows the random-walk-based diffusion vectors to denote valid probability mass functions over the nodes  of the network. This certainly enhances interpretability of the method, improves the numerical stability of the involved computations, and also reduces the search-space of the model is beneficial under data scarcity. 
Finally, imposing the simplex constraint makes the model amenable to a rigorous analysis that relates the dimensionality of the feature space to basic graph properties, as well as to a disciplined exploration of its limiting behavior. 

           \begin{table}[t]
           	\centering
	\caption{ Network Characteristics }
	\rowcolors{2}{}{gray!7}
	\begin{tabular} {lcccc}         			
		\toprule
		Graph & $|\mathcal{V}|$ & $|\mathcal{E}|$ & $|\mathcal{Y}|$ & Multilabel \\
		\midrule 
		
		\texttt{Citeseer} &  3,233  & 9,464 & 6 & No \\
		
		\texttt{Cora}  & 2,708  &  10,858 & 7 & No\\
		
		\texttt{PubMed} &  19,717 & 88,676 & 3 & No \\
		
		\texttt{PPI (H. Sapiens)} & 3,890 & 76,584 & 50 & Yes \\
		
		\texttt{Wikipedia} & 4,733 & 184,182 & 40 & Yes \\

		\texttt{BlogCatalog} & 10,312 & 333,983 & 39 & Yes \\
		\bottomrule
	\end{tabular}\label{tab:graphs}
\end{table}
\section{Experimental Evaluation}
\label{sec:numerical}
Our experiments compare the classification accuracy of the novel AdaDIF approach with  state-of-the-art alternatives. For the comparisons, we use 6 benchmark labeled graphs whose dimensions and basic attributes are summarized in Table \ref{tab:graphs}. All 6 graphs have nodes that belong to multiple classes, while the last 3 are \emph{multilabeled} (each node has \emph{one or more} labels). We evaluate performance of AdaDIF and the following: i) PPR and HK, which are special cases of AdaDIF as discussed in Section \ref{sec:problem};  {ii) Label propagation (LP) \cite{zhu2003semi};} iii) Node2vec \cite{grover2016node2vec};  iv) Deepwalk \cite{perozzi2014deepwalk}; v) Planetoid-G \cite{yang2016revisiting}; and, vi) graph convolutional networks (GCNs) \cite{kipf2016semi}. {We note here that AptRank~\cite{aptrank} was not considered in our experiments since it relies on meta-information that is not available for the benchmark datasets used here.  }

\begin{table*}[th!]
	\caption{ Micro F1 and Macro F1 Scores on Multiclass Networks (class-balanced sampling) }
	\resizebox{1\textwidth}{!}{		\begin{tabular} {clccccccccc}
			\toprule
			& Graph & \multicolumn{3}{c}{\texttt{Cora}} & \multicolumn{3}{c}{\texttt{Citeseer}} & \multicolumn{3}{c}{\texttt{PubMed}}\\
			\cmidrule(lr){3-5}\cmidrule(lr){6-8}\cmidrule(lr){9-11}
			& $|\mathcal{L}_c|$ & $5$ & $10$  & $20$ & $5$ & $10$  & $20$ & $5$ & $10$  & $20$ \\
			\midrule
			\parbox[t]{2mm}{\multirow{6}{*}{\rotatebox[origin=c]{90}{Micro-F1}}} &  
			AdaDIF   & $67.5\pm2.2$  & $\mathbf{71.0\pm2.0}$  & $\mathbf{73.2\pm1.2}$   &  $\mathbf{42.3\pm4.4}$  & $\mathbf{49.5\pm3.0}$ & $\mathbf{53.5\pm1.2}$ & $62.0\pm6.0$  & $68.5\pm4.5$  & $\mathbf{74.1\pm1.7}$\\
			& \CC{7}PPR & \CC{7}$67.1\pm2.3$  & \CC{7}$70.2\pm2.1$ & \CC{7}$72.8\pm1.5$  & \CC{7}$41.1\pm5.2$  & \CC{7}$48.7\pm2.5$  & \CC{7}$52.5\pm0.9$     &  \CC{7}$\mathbf{63.1\pm1.1}$ & \CC{7}$\mathbf{69.5\pm3.8}$ & \CC{7}$\mathbf{74.1\pm1.8}$ \\			
			&
			HK  & $67.0\pm2.5$  & $70.5\pm2.5$  & $72.9\pm1.2$   & $40.0\pm5.6$  & $48.0\pm2.4$  & $51.8\pm 1.1$     & $62.0\pm0.6$  & $68.3\pm4.7$ & $\mathbf{74.0\pm1.8}$ \\
			&
			\CC{7}LP  & \CC{7}$61.8\pm3.5$  & \CC{7}$66.3\pm4.2$  & \CC{7}$71.0\pm2.7$   & \CC{7}$40.7\pm2.5$  & \CC{7}$48.0\pm3.7$  & \CC{7}$51.9\pm1.3$     & \CC{7}$56.2\pm11.0$  & \CC{7}$68.0\pm6.1$ & \CC{7}$69.3\pm2.4$ \\
			& Node2vec  & $\mathbf{68.9\pm1.9}$  & $70.2\pm1.6$  & $72.4\pm1.2$  & $39.2\pm3.7$  & $46.5\pm2.4$  &  $51.0\pm1.4$  & $61.7\pm13.0$  & $66.4\pm4.6$ & $71.1\pm2.4$ \\
			& 
			\CC{7}Deepwalk & \CC{7}$68.4\pm2.0$  & \CC{7}$70.0\pm 1.6$  & \CC{7}$72.0\pm1.4$   & \CC{7}$38.4\pm 3.9$  & \CC{7}$45.5\pm2.0$  & \CC{7}$50.4\pm1.5$   & \CC{7}$61.5\pm1.3$  & \CC{7}$65.8\pm5.0$ & \CC{7}$70.5\pm2.2$ \\
			& Planetoid-G & $63.5\pm4.7$  & $65.6\pm2.7$  &  $69.0\pm1.5$   &  $37.8\pm4.0$  & $44.9\pm3.3$  &  $49.8\pm1.4$  & $60.7\pm2.0$  & $63.4\pm2.3$ & $68.0\pm1.5$ \\
			& 
			\CC{7}GCN & \CC{7}$60.1\pm3.7$  & \CC{7}$65.5\pm2.5$  & \CC{7}$68.6\pm1.9$  & \CC{7}$38.3\pm3.2$  & \CC{7}$44.2\pm2.2$  & \CC{7}$48.0\pm1.8$    & \CC{7}$60.0\pm1.9$  & \CC{7}$63.6\pm2.5$ & \CC{7}$70.5\pm1.5$\\
			\midrule
			\parbox[t]{2mm}{\multirow{7}{*}{\rotatebox[origin=c]{90}{Macro-F1}}}
			 &\CC{7}AdaDIF   & \CC{7}$\mathbf{65.5\pm2.5}$  & \CC{7}$\mathbf{70.6\pm2.2}$ & \CC{7}$\mathbf{72.0\pm1.1}$  & \CC{7}$\mathbf{36.1\pm3.9}$ & \CC{7}$\mathbf{44.0\pm2.8}$ & \CC{7}$\mathbf{48.1\pm1.2}$  & \CC{7}$60.4\pm0.6$  & \CC{7}$67.0\pm4.4$  & \CC{7}$\mathbf{72.6\pm1.8}$\\
			& 
			PPR & $65.0\pm2.3$  & $70.0\pm2.3$ & $71.9\pm1.5$  & $34.7\pm5.0$ & $43.5\pm2.3$ & $47.6\pm0.6$  & $\mathbf{61.7\pm0.6}$ & $\mathbf{68.1\pm3.6}$  & $\mathbf{72.6\pm1.8}$ \\			
			& \CC{7}HK  & \CC{7}$65.0\pm2.5$  & \CC{7}$70.0\pm2.6$ & \CC{7}$\mathbf{72.0\pm1.1}$  & \CC{7}$33.9\pm5.4$ & \CC{7}$42.8\pm2.2$ & \CC{7}$47.0\pm0.6$ & \CC{7}$60.5\pm0.6$  & \CC{7}$66.8\pm4.7$ & \CC{7}$\mathbf{72.7\pm1.8}$ \\
			& LP  & $60.1\pm3.2$  & $66.5\pm4.1$ & $70.6\pm2.3$  & $34.8\pm4.6$ & $41.8\pm3.9$ & $51.5\pm1.2$ & $51.5\pm12.3$  & $66.2\pm6.6$ & $67.8\pm2.0$ \\
			& 
			\CC{7}Node2vec  & \CC{7}$62.4\pm2.0$  & \CC{7}$64.7\pm1.7$  & \CC{7}$69.2\pm1.2$   & \CC{7}$34.6\pm2.7$  & \CC{7}$41.6\pm1.9$ & \CC{7}$45.3\pm1.5$ & \CC{7}$59.5\pm1.2$ & \CC{7}$64.0\pm3.8$ & \CC{7}$72.3\pm1.4$ \\
			& Deepwalk & $61.8\pm2.2$  & $64.5\pm2.0$ & $68.5\pm1.4$   & $34.0\pm2.5$  & $41.0\pm2.0$ & $44.7\pm1.8$   & $59.3\pm1.2$  & $63.8\pm4.0$ & $72.1\pm1.3$ \\
			& 
			\CC{7}Planetoid-G & \CC{7}$59.9\pm4.5$  & \CC{7}$63.0\pm3.0$  & \CC{7}$68.7\pm1.9$   & \CC{7}$33.3\pm2.5$  & \CC{7}$40.2\pm2.2$  & \CC{7}$43.6\pm2.0$  & \CC{7}$57.7\pm1.5$  & \CC{7}$61.9\pm3.5$ & \CC{7}$66.1\pm1.8$ \\
			& 
			GCN & $53.8\pm6.6$  & $61.9\pm2.6$ &  $63.8\pm1.5$  &  $32.8\pm2.0$ &  $39.1\pm1.8$ &  $43.0\pm1.7$  & $54.4\pm4.1$  & $57.2\pm5.2$ & $60.5\pm2.4$ \\
			\bottomrule	        		       				           			
	\end{tabular}}\label{tab:Micro_mclass_balanced}
\vspace*{0.37cm}
\end{table*}

\begin{table*}[th!]
	\caption{ Micro F1 and Macro F1 Scores of Various Algorithms on Multilabel Networks }
	\resizebox{1\textwidth}{!}{	\begin{tabular} {clccccccccc}
			\toprule
			& Graph & \multicolumn{3}{c}{\texttt{PPI}} & \multicolumn{3}{c}{\texttt{BlogCatalog}} & \multicolumn{3}{c}{\texttt{Wikipedia}}\\
			\cmidrule(lr){3-5}\cmidrule(lr){6-8}\cmidrule(lr){9-11}
			& $|\mathcal{L}|/|\mathcal{V}|$ & $10\%$ & $20\%$  & $30\%$ & $10\%$ & $20\%$  & $30\%$ & $10\%$ & $20\%$  & $30\%$ \\
			\midrule 
			\parbox[t]{2mm}{\multirow{5}{*}{\rotatebox[origin=c]{90}{Micro-F1}}} 
			&\CC{7}AdaDIF   & \CC{7}$15.4\pm0.5$ & \CC{7}$17.9\pm0.7$ & \CC{7}$\mathbf{19.2\pm0.6}$ & \CC{7}$31.5\pm0.6$   & \CC{7}$34.4\pm0.5$ & \CC{7}$36.3\pm0.4$ & \CC{7}$28.2\pm0.9$ & \CC{7}$30.0\pm0.5$  & \CC{7}$31.2\pm0.7$ \\
			& PPR  & $13.8\pm0.5$   & $15.8\pm0.6$ & $17.0\pm0.4$ & $21.1\pm0.8$  & $23.6\pm0.6$ & $25.2\pm0.6$ & $10.5\pm1.5$ & $8.1\pm0.7$ & $7.2\pm0.5$ \\			
			& \CC{7}
			\CC{7}HK  & \CC{7}$14.5\pm0.5$   & \CC{7}$16.7\pm0.6$ & \CC{7}$18.1\pm0.5$ & \CC{7}$22.2\pm1.0$  & \CC{7}$24.7\pm0.7$ & \CC{7}$26.6\pm0.7$ & \CC{7}$9.3\pm1.4$ & \CC{7}$7.3\pm0.7$ & \CC{7}$6.0\pm0.7$\\
			& Node2vec   & $\mathbf{16.5\pm0.6}$  & $\mathbf{18.2\pm0.3}$ & $19.1\pm0.3$  &  $\mathbf{35.0\pm0.3}$  & $\mathbf{36.3\pm0.3}$ & $\mathbf{37.2\pm0.2}$  &    $\mathbf{42.3\pm0.9}$ & $\mathbf{44.0\pm0.6}$ & $\mathbf{45.1\pm0.4}$ \\
			&
			\CC{7}Deepwalk  & \CC{7}$16.0\pm0.6$  & \CC{7}$17.9\pm0.5$ & \CC{7}$18.8\pm0.4$ & \CC{7}$34.2\pm0.4$ & \CC{7}$35.7\pm0.3$ & \CC{7}$36.4\pm0.4$  & \CC{7}$41.0\pm0.8$  & \CC{7}$43.5\pm0.5$ & \CC{7}$44.1\pm0.5$\\
			\midrule
			\parbox[t]{2mm}{\multirow{5}{*}{\rotatebox[origin=c]{90}{Macro-F1}}} 
			& AdaDIF    & $\mathbf{13.4\pm0.6}$   & $\mathbf{15.4\pm0.7}$ & $\mathbf{16.5\pm0.7}$ & $\mathbf{23.0\pm0.6}$  & $\mathbf{25.3\pm0.4}$ & $\mathbf{27.0\pm0.4}$ & $\mathbf{7.7\pm0.3}$  & $\mathbf{8.3\pm0.3}$ &  $\mathbf{9.0\pm0.2}$ \\
			& 
			\CC{7}PPR  & \CC{7}$12.9\pm0.4$ & \CC{7}$14.7\pm0.5$ & \CC{7}$15.8\pm0.4$ & \CC{7}$17.3\pm0.5$  & \CC{7}$19.5\pm0.4$  & \CC{7}$20.8\pm0.3$ & \CC{7}$4.4\pm0.3$ & \CC{7}$3.8\pm0.6$ & \CC{7}$3.6\pm0.2$ \\			
			& HK   & $\mathbf{13.4\pm0.6}$  & $\mathbf{15.4\pm0.5}$ & $\mathbf{16.5\pm0.4}$ & $18.4\pm0.6$  & $20.7\pm0.4$  & $22.3\pm0.4$ & $4.2\pm0.4$ & $3.7\pm0.5$ & $3.5\pm0.2$\\
			& 
			\CC{7}Node2vec   & \CC{7}$13.1\pm0.6$  & \CC{7}$15.2\pm0.5$ & \CC{7}$16.0\pm0.5$ & \CC{7}$16.8\pm0.5$ & \CC{7}$19.0\pm0.3$ & \CC{7}$20.1\pm0.4$ & \CC{7}$7.6\pm0.3$ & \CC{7}$8.2\pm0.3$ & \CC{7}$8.5\pm0.3$ \\
			& Deepwalk & $12.7\pm0.7$  & $15.1\pm0.6$ & $16.0\pm0.5$ & $16.6\pm0.5$  & $18.7\pm0.5$ &  $19.6\pm0.4$  & $7.3\pm0.3$ & $8.1\pm0.2$ & $8.2\pm0.2$ \\
			\bottomrule      					           			
		\end{tabular}\label{tab:Micro_mlabel}
	}
\vspace*{0.35cm}
\end{table*}

We performed $10$-fold cross-validation to select parameters needed by i) - v). For HK, we performed grid search over $t\in[1.0,5.0,10.0,15.0]$. For PPR, we fixed $\alpha=0.98$ since it is well documented that $\alpha$ close to $1$ yields reliable performance; see e.g., \cite{lin2010semi}. Both HK and PPR were run for $50$ steps for convergence to be in effect; see Fig \ref{wrt_K}; LP was also run for 50 steps. For Node2vec, we fixed most parameters to the values suggested in \cite{grover2016node2vec}, and performed grid search for $p,q\in[0.25,1.0,2.0,4.0]$. Since Deepwalk can be seen as Node2vec with $p=q=1.0$, we used the Node2vec Python implementation for both. As in \cite{grover2016node2vec,perozzi2014deepwalk}, we used the embeded node-features to train a supervised logistic regression classifier with $\ell_2$ regularization. For AdaDIF, we fixed $\lambda=15.0$, while $K=15$ was sufficient to attain desirable accuracy (cf. Fig. \ref{wrt_K}); only the values of Boolean variables \emph{Unconstained} and \emph{Dictionary Mode} (see Algorithm 1) were tuned by validation. For the multilabel graphs, we found $\lambda=5.0$ and even shorter walks of $K=10$ to perform well. For the dictionary mode of AdaDIF, we preselected $D=10$, with the first five collumns of $\mathbf{C}$ being HK coefficients with parameters $t\in[5,8,12,15, 20]$, and the other five polynomial coefficients $c_i=k^\beta$ with $\beta\in[2,4,6,8,10]$.

\setlength\figW{0.85\columnwidth}
\setlength\figH{0.55\columnwidth}
\begin{figure}[t!]  
	\small
%
%

\definecolor{mycolor1}{RGB}{215,25,28}%
\definecolor{mycolor2}{RGB}{43,131,186}%
\definecolor{mycolor3}{RGB}{166,217,106}

\begin{tikzpicture}

\begin{axis}[%
width=0.951\figW,
height=\figH,
at={(0\figW,0\figH)},
scale only axis,
xmin=2,
xmax=31,
xlabel style={font=\color{white!15!black}},
xlabel={ \# of landing probabilities (K)},
ymin=58,
ymax=73.5,
ylabel style={font=\color{white!15!black}},
ylabel near ticks,
ylabel={Micro-F1 Score (\%)},
axis background/.style={fill=white},
legend style={at={(0.668,0.038)}, anchor=south west, legend cell align=left, align=left, draw=none}
]
\addplot [color=mycolor3, line width=1.0pt, mark size=2pt, mark=diamond, mark options={solid, mycolor3}]
 plot [error bars/.cd, y dir = both, y explicit]
 table[row sep=crcr, y error plus index=2, y error minus index=3]{%
3	59.7	1.5	1.5\\
5	65.1	3.5	3.5\\
7	67.8	2.5	2.5\\
10	68	2.7	2.7\\
15	68.2	2	2\\
20	69.8	2.4	2.4\\
30	70.1	2.1	2.1\\
};
\addlegendentry{PPR}

\addplot [color=mycolor2, line width=1.0pt, mark size=2pt, mark=o, mark options={solid, mycolor2}]
 plot [error bars/.cd, y dir = both, y explicit]
 table[row sep=crcr, y error plus index=2, y error minus index=3]{%
3	59.5	1.5	1.5\\
5	64.9	3.6	3.6\\
7	67.4	2.6	2.6\\
10	67.5	2.8	2.8\\
15	68.2	1.9	1.9\\
20	69.75	2.4	2.4\\
30	70.5	2.2	2.2\\
};
\addlegendentry{HK}

\addplot [color=mycolor1, line width=1.0pt, mark size=2pt, mark=square, mark options={solid, mycolor1}]
 plot [error bars/.cd, y dir = both, y explicit]
 table[row sep=crcr, y error plus index=2, y error minus index=3]{%
3	59.8	1.4	1.4\\
5	65.5	3.4	3.4\\
7	68.6	2.4	2.4\\
10	69.2	2.7	2.7\\
15	69.3	2.1	2.1\\
20	70.4	2.5	2.5\\
30	70.5	2.5	2.5\\
};
\addlegendentry{AdaDIF}

\end{axis}
\end{tikzpicture}%
	\caption{Micro-F1 score for AdaDIF and non-adaptive diffusions on $5\%$ labeled \texttt{Cora} graph as a function of the length of underline random walks.} \label{wrt_K}
\end{figure}
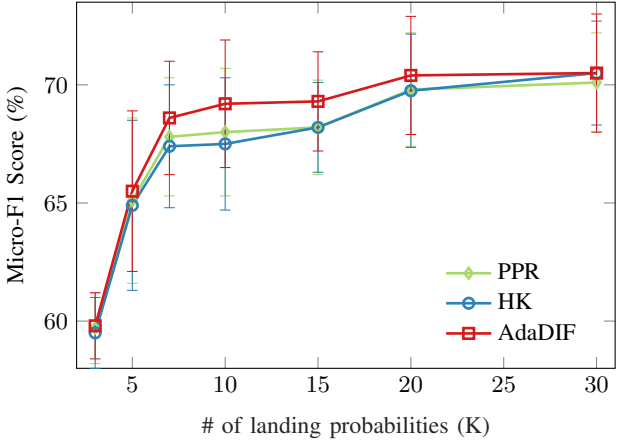

For multiclass experiments, we evaluated the performance of all algorithms on the three benchmark citation networks, namely \texttt{Cora}, \texttt{Citeseer}, and \texttt{PubMed}.   {We obtained the labels of an increasing number of nodes via uniform, class-balanced sampling, and predicted the labels of the remaining nodes.   
Thus, instead of sampling nodes over the graph uniformly at random, we randomly sample a given number of nodes per class. For each graph, we performed 20 experiments, each time sampling $5,10,$ and $20$ nodes per class.} For each experiment, classification accuracy was measured on the unlabeled nodes in terms of Micro-F1 and Macro-F1 scores; see e.g.,~\cite{manning2008ir}. The results were averaged over 20 experiments, with mean and standard deviation reported in Table \ref{tab:Micro_mclass_balanced}. Evidently, AdaDIF achieves state of the art performance for all graphs. For \texttt{Cora} and \texttt{PubMed}, AdaDIF was switched to dictionary mode, while for \texttt{Citeseer}, where the gain in accuracy is more significant, unconstrained diffusions were employed. In the multiclass setting, diffusion-based classifiers (AdaDIF, PPR, and HK) outperformed the embedding-based methods by a small margin, and GCNs by a larger margin. It should be noted however that GCNs were mainly designed to combine the graph with node features. In our ``featureless'' setting, we used the identity matrix columns as input features, as suggested in \cite[Appendix]{kipf2016semi}. 
\begin{figure*}[!tbp]
	\definecolor{color1}{RGB}{215,25,28}%
	\definecolor{color0}{RGB}{166,206,227}%
	\definecolor{color2}{RGB}{43,131,186}%
	\definecolor{color3}{RGB}{166,217,106}
				\scalebox{.7}{
\begin{tikzpicture}


\definecolor{color1}{RGB}{215,25,28}%
\definecolor{color0}{RGB}{166,206,227}%
\definecolor{color2}{RGB}{43,131,186}%
\definecolor{color3}{RGB}{166,217,106}

\begin{axis}[
ylabel={\large Micro F1-score},
xmin=0, xmax=100,
ymin=0.65, ymax=0.74,
tick align=outside,
tick pos=left,
x grid style={lightgray!92.026143790849673!black},
y grid style={lightgray!92.026143790849673!black},
legend columns = 4,
legend style={draw=none ,fill=white,legend cell align=center},
legend entries={{$k$-step landing probabilities},{AdaDIF},{PPR},{HK}},
legend cell align={left},
legend to name=named2
]
\addplot [semithick, color0]
table {%
0 0.412893203883495
1 0.563106796116505
2 0.650796116504854
3 0.674563106796117
4 0.689941747572816
5 0.693980582524272
6 0.70221359223301
7 0.704932038834951
8 0.709592233009709
9 0.711844660194175
10 0.714485436893204
11 0.716038834951456
12 0.716427184466019
13 0.71852427184466
14 0.719300970873786
15 0.720621359223301
16 0.722097087378641
17 0.72147572815534
18 0.72295145631068
19 0.721242718446602
20 0.721553398058252
21 0.720077669902913
22 0.720543689320388
23 0.718601941747573
24 0.718601941747573
25 0.716582524271845
26 0.716427184466019
27 0.715495145631068
28 0.716271844660194
29 0.713864077669903
30 0.715029126213592
31 0.71231067961165
32 0.712543689320388
33 0.710058252427184
34 0.709825242718447
35 0.707339805825243
36 0.707029126213592
37 0.704
38 0.702679611650485
39 0.70221359223301
40 0.70104854368932
41 0.698485436893204
42 0.697941747572816
43 0.696388349514563
44 0.696
45 0.693902912621359
46 0.693126213592233
47 0.690563106796117
48 0.690330097087379
49 0.688155339805825
50 0.686679611650485
51 0.683572815533981
52 0.682252427184466
53 0.679922330097087
54 0.678757281553398
55 0.678291262135922
56 0.676504854368932
57 0.674796116504854
58 0.673320388349514
59 0.670135922330097
60 0.669126213592233
61 0.665398058252427
62 0.663223300970874
63 0.660194174757282
64 0.659417475728155
65 0.655456310679612
66 0.654601941747573
67 0.651029126213592
68 0.64947572815534
69 0.645980582524272
70 0.643805825242718
71 0.640233009708738
72 0.63821359223301
73 0.63495145631068
74 0.632233009708738
75 0.627805825242718
76 0.625941747572816
77 0.621902912621359
78 0.61895145631068
79 0.613825242718447
80 0.611883495145631
81 0.607766990291262
82 0.606058252427184
83 0.602097087378641
84 0.598834951456311
85 0.594873786407767
86 0.592310679611651
87 0.586873786407767
88 0.58547572815534
89 0.580582524271845
90 0.577864077669903
91 0.573359223300971
92 0.571339805825243
93 0.567378640776699
94 0.565203883495146
95 0.56178640776699
96 0.561165048543689
97 0.557514563106796
98 0.556582524271845
99 0.553553398058252
};
\addplot [semithick, color1, mark=diamond, mark size=1.5, mark options={solid}]
table {%
0 0.73004854368932
1 0.73004854368932
2 0.73004854368932
3 0.73004854368932
4 0.73004854368932
5 0.73004854368932
6 0.73004854368932
7 0.73004854368932
8 0.73004854368932
9 0.73004854368932
10 0.73004854368932
11 0.73004854368932
12 0.73004854368932
13 0.73004854368932
14 0.73004854368932
15 0.73004854368932
16 0.73004854368932
17 0.73004854368932
18 0.73004854368932
19 0.73004854368932
20 0.73004854368932
21 0.73004854368932
22 0.73004854368932
23 0.73004854368932
24 0.73004854368932
25 0.73004854368932
26 0.73004854368932
27 0.73004854368932
28 0.73004854368932
29 0.73004854368932
30 0.73004854368932
31 0.73004854368932
32 0.73004854368932
33 0.73004854368932
34 0.73004854368932
35 0.73004854368932
36 0.73004854368932
37 0.73004854368932
38 0.73004854368932
39 0.73004854368932
40 0.73004854368932
41 0.73004854368932
42 0.73004854368932
43 0.73004854368932
44 0.73004854368932
45 0.73004854368932
46 0.73004854368932
47 0.73004854368932
48 0.73004854368932
49 0.73004854368932
50 0.73004854368932
51 0.73004854368932
52 0.73004854368932
53 0.73004854368932
54 0.73004854368932
55 0.73004854368932
56 0.73004854368932
57 0.73004854368932
58 0.73004854368932
59 0.73004854368932
60 0.73004854368932
61 0.73004854368932
62 0.73004854368932
63 0.73004854368932
64 0.73004854368932
65 0.73004854368932
66 0.73004854368932
67 0.73004854368932
68 0.73004854368932
69 0.73004854368932
70 0.73004854368932
71 0.73004854368932
72 0.73004854368932
73 0.73004854368932
74 0.73004854368932
75 0.73004854368932
76 0.73004854368932
77 0.73004854368932
78 0.73004854368932
79 0.73004854368932
80 0.73004854368932
81 0.73004854368932
82 0.73004854368932
83 0.73004854368932
84 0.73004854368932
85 0.73004854368932
86 0.73004854368932
87 0.73004854368932
88 0.73004854368932
89 0.73004854368932
90 0.73004854368932
91 0.73004854368932
92 0.73004854368932
93 0.73004854368932
94 0.73004854368932
95 0.73004854368932
96 0.73004854368932
97 0.73004854368932
98 0.73004854368932
99 0.73004854368932
};
\addplot [semithick, color2, mark=o, mark size=1, mark options={solid}]
table {%
0 0.727067961165049
1 0.727067961165049
2 0.727067961165049
3 0.727067961165049
4 0.727067961165049
5 0.727067961165049
6 0.727067961165049
7 0.727067961165049
8 0.727067961165049
9 0.727067961165049
10 0.727067961165049
11 0.727067961165049
12 0.727067961165049
13 0.727067961165049
14 0.727067961165049
15 0.727067961165049
16 0.727067961165049
17 0.727067961165049
18 0.727067961165049
19 0.727067961165049
20 0.727067961165049
21 0.727067961165049
22 0.727067961165049
23 0.727067961165049
24 0.727067961165049
25 0.727067961165049
26 0.727067961165049
27 0.727067961165049
28 0.727067961165049
29 0.727067961165049
30 0.727067961165049
31 0.727067961165049
32 0.727067961165049
33 0.727067961165049
34 0.727067961165049
35 0.727067961165049
36 0.727067961165049
37 0.727067961165049
38 0.727067961165049
39 0.727067961165049
40 0.727067961165049
41 0.727067961165049
42 0.727067961165049
43 0.727067961165049
44 0.727067961165049
45 0.727067961165049
46 0.727067961165049
47 0.727067961165049
48 0.727067961165049
49 0.727067961165049
50 0.727067961165049
51 0.727067961165049
52 0.727067961165049
53 0.727067961165049
54 0.727067961165049
55 0.727067961165049
56 0.727067961165049
57 0.727067961165049
58 0.727067961165049
59 0.727067961165049
60 0.727067961165049
61 0.727067961165049
62 0.727067961165049
63 0.727067961165049
64 0.727067961165049
65 0.727067961165049
66 0.727067961165049
67 0.727067961165049
68 0.727067961165049
69 0.727067961165049
70 0.727067961165049
71 0.727067961165049
72 0.727067961165049
73 0.727067961165049
74 0.727067961165049
75 0.727067961165049
76 0.727067961165049
77 0.727067961165049
78 0.727067961165049
79 0.727067961165049
80 0.727067961165049
81 0.727067961165049
82 0.727067961165049
83 0.727067961165049
84 0.727067961165049
85 0.727067961165049
86 0.727067961165049
87 0.727067961165049
88 0.727067961165049
89 0.727067961165049
90 0.727067961165049
91 0.727067961165049
92 0.727067961165049
93 0.727067961165049
94 0.727067961165049
95 0.727067961165049
96 0.727067961165049
97 0.727067961165049
98 0.727067961165049
99 0.727067961165049
};
\addplot [semithick, color3, mark=o, mark size=1, mark options={solid}]
table {%
0 0.723572815533981
1 0.723572815533981
2 0.723572815533981
3 0.723572815533981
4 0.723572815533981
5 0.723572815533981
6 0.723572815533981
7 0.723572815533981
8 0.723572815533981
9 0.723572815533981
10 0.723572815533981
11 0.723572815533981
12 0.723572815533981
13 0.723572815533981
14 0.723572815533981
15 0.723572815533981
16 0.723572815533981
17 0.723572815533981
18 0.723572815533981
19 0.723572815533981
20 0.723572815533981
21 0.723572815533981
22 0.723572815533981
23 0.723572815533981
24 0.723572815533981
25 0.723572815533981
26 0.723572815533981
27 0.723572815533981
28 0.723572815533981
29 0.723572815533981
30 0.723572815533981
31 0.723572815533981
32 0.723572815533981
33 0.723572815533981
34 0.723572815533981
35 0.723572815533981
36 0.723572815533981
37 0.723572815533981
38 0.723572815533981
39 0.723572815533981
40 0.723572815533981
41 0.723572815533981
42 0.723572815533981
43 0.723572815533981
44 0.723572815533981
45 0.723572815533981
46 0.723572815533981
47 0.723572815533981
48 0.723572815533981
49 0.723572815533981
50 0.723572815533981
51 0.723572815533981
52 0.723572815533981
53 0.723572815533981
54 0.723572815533981
55 0.723572815533981
56 0.723572815533981
57 0.723572815533981
58 0.723572815533981
59 0.723572815533981
60 0.723572815533981
61 0.723572815533981
62 0.723572815533981
63 0.723572815533981
64 0.723572815533981
65 0.723572815533981
66 0.723572815533981
67 0.723572815533981
68 0.723572815533981
69 0.723572815533981
70 0.723572815533981
71 0.723572815533981
72 0.723572815533981
73 0.723572815533981
74 0.723572815533981
75 0.723572815533981
76 0.723572815533981
77 0.723572815533981
78 0.723572815533981
79 0.723572815533981
80 0.723572815533981
81 0.723572815533981
82 0.723572815533981
83 0.723572815533981
84 0.723572815533981
85 0.723572815533981
86 0.723572815533981
87 0.723572815533981
88 0.723572815533981
89 0.723572815533981
90 0.723572815533981
91 0.723572815533981
92 0.723572815533981
93 0.723572815533981
94 0.723572815533981
95 0.723572815533981
96 0.723572815533981
97 0.723572815533981
98 0.723572815533981
99 0.723572815533981
};
\end{axis}

\end{tikzpicture}}
				\scalebox{.7}{
\begin{tikzpicture}

\definecolor{color1}{RGB}{215,25,28}%
\definecolor{color0}{RGB}{166,206,227}%
\definecolor{color2}{RGB}{43,131,186}%
\definecolor{color3}{RGB}{166,217,106}

\begin{axis}[
xmin=0, xmax=100,
ymin=0.48, ymax=0.54,
tick align=outside,
tick pos=left,
x grid style={lightgray!92.026143790849673!black},
y grid style={lightgray!92.026143790849673!black},
]
\addlegendimage{no markers, color0}
\addlegendimage{no markers, color1}
\addlegendimage{no markers, color2}
\addlegendimage{no markers, color3}
\addplot [semithick, color0]
table {%
0 0.257827575474634
1 0.337068160597572
2 0.417055711173358
3 0.456271397447868
4 0.478244631185808
5 0.483162153750389
6 0.494926859632742
7 0.493557422969188
8 0.500155617802677
9 0.497354497354497
10 0.50432617491441
11 0.50140056022409
12 0.507874260815437
13 0.506131341425459
14 0.511858076563959
15 0.508621226268285
16 0.514098972922503
17 0.510364145658263
18 0.515655150949269
19 0.511360099595394
20 0.516526610644258
21 0.513600995953937
22 0.518020541549953
23 0.514970432617491
24 0.518829754123872
25 0.516090880796763
26 0.520323685029567
27 0.516837846249611
28 0.52156862745098
29 0.517335823218176
30 0.522813569872393
31 0.518954248366013
32 0.523934018051665
33 0.519701213818861
34 0.524743230625584
35 0.520759414877062
36 0.525427948957361
37 0.521444133208839
38 0.52536570183629
39 0.522004357298475
40 0.52536570183629
41 0.522004357298475
42 0.525676937441643
43 0.521942110177404
44 0.525801431683785
45 0.522315592903828
46 0.525925925925926
47 0.521630874572051
48 0.525116713352007
49 0.520821661998133
50 0.524867724867725
51 0.521008403361345
52 0.524867724867725
53 0.520634920634921
54 0.52443199502023
55 0.520634920634921
56 0.524058512293806
57 0.519701213818861
58 0.523000311235605
59 0.518580765639589
60 0.522689075630252
61 0.518207282913165
62 0.522004357298475
63 0.5177093059446
64 0.521817615935263
65 0.517460317460318
66 0.52150638032991
67 0.517273576097105
68 0.520821661998133
69 0.516588857765328
70 0.519887955182073
71 0.515655150949269
72 0.519203236850296
73 0.514908185496421
74 0.51864301276066
75 0.514410208527856
76 0.518331777155307
77 0.514036725801432
78 0.517833800186741
79 0.51384998443822
80 0.517211328976035
81 0.513289760348584
82 0.51677559912854
83 0.513040771864301
84 0.51671335200747
85 0.512854030501089
86 0.516028633675693
87 0.512418300653595
88 0.515904139433551
89 0.5119825708061
90 0.515655150949269
91 0.511422346716464
92 0.515157173980703
93 0.511111111111111
94 0.514721444133209
95 0.510488639900405
96 0.514223467164644
97 0.51005291005291
98 0.513725490196078
99 0.509243697478992
};
\addplot [semithick, color1, mark=diamond, mark size=1.5, mark options={solid}]
table {%
0 0.530727046374105
1 0.530727046374105
2 0.530727046374105
3 0.530727046374105
4 0.530727046374105
5 0.530727046374105
6 0.530727046374105
7 0.530727046374105
8 0.530727046374105
9 0.530727046374105
10 0.530727046374105
11 0.530727046374105
12 0.530727046374105
13 0.530727046374105
14 0.530727046374105
15 0.530727046374105
16 0.530727046374105
17 0.530727046374105
18 0.530727046374105
19 0.530727046374105
20 0.530727046374105
21 0.530727046374105
22 0.530727046374105
23 0.530727046374105
24 0.530727046374105
25 0.530727046374105
26 0.530727046374105
27 0.530727046374105
28 0.530727046374105
29 0.530727046374105
30 0.530727046374105
31 0.530727046374105
32 0.530727046374105
33 0.530727046374105
34 0.530727046374105
35 0.530727046374105
36 0.530727046374105
37 0.530727046374105
38 0.530727046374105
39 0.530727046374105
40 0.530727046374105
41 0.530727046374105
42 0.530727046374105
43 0.530727046374105
44 0.530727046374105
45 0.530727046374105
46 0.530727046374105
47 0.530727046374105
48 0.530727046374105
49 0.530727046374105
50 0.530727046374105
51 0.530727046374105
52 0.530727046374105
53 0.530727046374105
54 0.530727046374105
55 0.530727046374105
56 0.530727046374105
57 0.530727046374105
58 0.530727046374105
59 0.530727046374105
60 0.530727046374105
61 0.530727046374105
62 0.530727046374105
63 0.530727046374105
64 0.530727046374105
65 0.530727046374105
66 0.530727046374105
67 0.530727046374105
68 0.530727046374105
69 0.530727046374105
70 0.530727046374105
71 0.530727046374105
72 0.530727046374105
73 0.530727046374105
74 0.530727046374105
75 0.530727046374105
76 0.530727046374105
77 0.530727046374105
78 0.530727046374105
79 0.530727046374105
80 0.530727046374105
81 0.530727046374105
82 0.530727046374105
83 0.530727046374105
84 0.530727046374105
85 0.530727046374105
86 0.530727046374105
87 0.530727046374105
88 0.530727046374105
89 0.530727046374105
90 0.530727046374105
91 0.530727046374105
92 0.530727046374105
93 0.530727046374105
94 0.530727046374105
95 0.530727046374105
96 0.530727046374105
97 0.530727046374105
98 0.530727046374105
99 0.530727046374105
};
\addplot [semithick, color2, mark=o, mark size=1, mark options={solid}]
table {%
0 0.52150638032991
1 0.52150638032991
2 0.52150638032991
3 0.52150638032991
4 0.52150638032991
5 0.52150638032991
6 0.52150638032991
7 0.52150638032991
8 0.52150638032991
9 0.52150638032991
10 0.52150638032991
11 0.52150638032991
12 0.52150638032991
13 0.52150638032991
14 0.52150638032991
15 0.52150638032991
16 0.52150638032991
17 0.52150638032991
18 0.52150638032991
19 0.52150638032991
20 0.52150638032991
21 0.52150638032991
22 0.52150638032991
23 0.52150638032991
24 0.52150638032991
25 0.52150638032991
26 0.52150638032991
27 0.52150638032991
28 0.52150638032991
29 0.52150638032991
30 0.52150638032991
31 0.52150638032991
32 0.52150638032991
33 0.52150638032991
34 0.52150638032991
35 0.52150638032991
36 0.52150638032991
37 0.52150638032991
38 0.52150638032991
39 0.52150638032991
40 0.52150638032991
41 0.52150638032991
42 0.52150638032991
43 0.52150638032991
44 0.52150638032991
45 0.52150638032991
46 0.52150638032991
47 0.52150638032991
48 0.52150638032991
49 0.52150638032991
50 0.52150638032991
51 0.52150638032991
52 0.52150638032991
53 0.52150638032991
54 0.52150638032991
55 0.52150638032991
56 0.52150638032991
57 0.52150638032991
58 0.52150638032991
59 0.52150638032991
60 0.52150638032991
61 0.52150638032991
62 0.52150638032991
63 0.52150638032991
64 0.52150638032991
65 0.52150638032991
66 0.52150638032991
67 0.52150638032991
68 0.52150638032991
69 0.52150638032991
70 0.52150638032991
71 0.52150638032991
72 0.52150638032991
73 0.52150638032991
74 0.52150638032991
75 0.52150638032991
76 0.52150638032991
77 0.52150638032991
78 0.52150638032991
79 0.52150638032991
80 0.52150638032991
81 0.52150638032991
82 0.52150638032991
83 0.52150638032991
84 0.52150638032991
85 0.52150638032991
86 0.52150638032991
87 0.52150638032991
88 0.52150638032991
89 0.52150638032991
90 0.52150638032991
91 0.52150638032991
92 0.52150638032991
93 0.52150638032991
94 0.52150638032991
95 0.52150638032991
96 0.52150638032991
97 0.52150638032991
98 0.52150638032991
99 0.52150638032991
};
\addplot [semithick, color3, mark=o, mark size=1, mark options={solid}]
table {%
0 0.518269530034236
1 0.518269530034236
2 0.518269530034236
3 0.518269530034236
4 0.518269530034236
5 0.518269530034236
6 0.518269530034236
7 0.518269530034236
8 0.518269530034236
9 0.518269530034236
10 0.518269530034236
11 0.518269530034236
12 0.518269530034236
13 0.518269530034236
14 0.518269530034236
15 0.518269530034236
16 0.518269530034236
17 0.518269530034236
18 0.518269530034236
19 0.518269530034236
20 0.518269530034236
21 0.518269530034236
22 0.518269530034236
23 0.518269530034236
24 0.518269530034236
25 0.518269530034236
26 0.518269530034236
27 0.518269530034236
28 0.518269530034236
29 0.518269530034236
30 0.518269530034236
31 0.518269530034236
32 0.518269530034236
33 0.518269530034236
34 0.518269530034236
35 0.518269530034236
36 0.518269530034236
37 0.518269530034236
38 0.518269530034236
39 0.518269530034236
40 0.518269530034236
41 0.518269530034236
42 0.518269530034236
43 0.518269530034236
44 0.518269530034236
45 0.518269530034236
46 0.518269530034236
47 0.518269530034236
48 0.518269530034236
49 0.518269530034236
50 0.518269530034236
51 0.518269530034236
52 0.518269530034236
53 0.518269530034236
54 0.518269530034236
55 0.518269530034236
56 0.518269530034236
57 0.518269530034236
58 0.518269530034236
59 0.518269530034236
60 0.518269530034236
61 0.518269530034236
62 0.518269530034236
63 0.518269530034236
64 0.518269530034236
65 0.518269530034236
66 0.518269530034236
67 0.518269530034236
68 0.518269530034236
69 0.518269530034236
70 0.518269530034236
71 0.518269530034236
72 0.518269530034236
73 0.518269530034236
74 0.518269530034236
75 0.518269530034236
76 0.518269530034236
77 0.518269530034236
78 0.518269530034236
79 0.518269530034236
80 0.518269530034236
81 0.518269530034236
82 0.518269530034236
83 0.518269530034236
84 0.518269530034236
85 0.518269530034236
86 0.518269530034236
87 0.518269530034236
88 0.518269530034236
89 0.518269530034236
90 0.518269530034236
91 0.518269530034236
92 0.518269530034236
93 0.518269530034236
94 0.518269530034236
95 0.518269530034236
96 0.518269530034236
97 0.518269530034236
98 0.518269530034236
99 0.518269530034236
};
\end{axis}

\end{tikzpicture}}		
                \scalebox{.7}{
\begin{tikzpicture}

\definecolor{color1}{RGB}{215,25,28}%
\definecolor{color0}{RGB}{166,206,227}%
\definecolor{color2}{RGB}{43,131,186}%
\definecolor{color3}{RGB}{166,217,106}

\begin{axis}[
xmin=0,xmax=100,
ymin=0.68, ymax=0.76,
tick align=outside,
tick pos=left,
x grid style={lightgray!92.026143790849673!black},
y grid style={lightgray!92.026143790849673!black},
]
\addplot [semithick, color0]
table {%
0 0.397110885045778
1 0.431892166836216
2 0.512370295015259
3 0.626785350966429
4 0.667202441505595
5 0.69206510681587
6 0.702146490335707
7 0.710925737538149
8 0.717924720244151
9 0.72175991861648
10 0.72766022380468
11 0.729959308240081
12 0.733987792472025
13 0.736052899287894
14 0.740213631739573
15 0.739654120040692
16 0.743550356052899
17 0.744231943031536
18 0.746113936927772
19 0.746622583926755
20 0.745869786368261
21 0.748056968463886
22 0.745991861648016
23 0.749043743641913
24 0.746897253306205
25 0.74942014242116
26 0.747263479145473
27 0.749633774160733
28 0.748026449643947
29 0.749938962360122
30 0.748026449643947
31 0.750091556459817
32 0.748189216683622
33 0.750111902339776
34 0.747497456765005
35 0.749206510681587
36 0.746815869786368
37 0.748504577822991
38 0.745707019328586
39 0.748046795523906
40 0.744852492370295
41 0.747151576805697
42 0.744516785350967
43 0.746734486266531
44 0.743397761953204
45 0.744842319430316
46 0.742533062054934
47 0.744221770091556
48 0.741597151576806
49 0.743173957273652
50 0.740793489318413
51 0.74206510681587
52 0.739196337741607
53 0.740732451678535
54 0.738158697863682
55 0.739786368260427
56 0.736642929806714
57 0.739196337741607
58 0.735554425228891
59 0.738321464903357
60 0.73441505595117
61 0.736246185147508
62 0.733102746693794
63 0.734496439471007
64 0.731261444557477
65 0.732746693794507
66 0.729369277721261
67 0.730508646998983
68 0.727314343845371
69 0.728748728382503
70 0.725534079348932
71 0.72646998982706
72 0.723886063072228
73 0.725055951169888
74 0.722390640895219
75 0.723316378433367
76 0.720671414038657
77 0.721230925737538
78 0.71912512716175
79 0.720142421159715
80 0.718128179043744
81 0.71880976602238
82 0.716856561546287
83 0.71706002034588
84 0.715717192268566
85 0.715849440488301
86 0.714669379450661
87 0.7146998982706
88 0.713774160732452
89 0.713804679552391
90 0.712919633774161
91 0.7129501525941
92 0.711851475076297
93 0.711973550356053
94 0.711139369277721
95 0.711159715157681
96 0.710325534079349
97 0.710406917599186
98 0.709613428280773
99 0.709593082400814
};
\addplot [semithick, color1, mark=diamond, mark size=1.5, mark options={solid}]
table {%
0 0.743583926754832
1 0.743583926754832
2 0.743583926754832
3 0.743583926754832
4 0.743583926754832
5 0.743583926754832
6 0.743583926754832
7 0.743583926754832
8 0.743583926754832
9 0.743583926754832
10 0.743583926754832
11 0.743583926754832
12 0.743583926754832
13 0.743583926754832
14 0.743583926754832
15 0.743583926754832
16 0.743583926754832
17 0.743583926754832
18 0.743583926754832
19 0.743583926754832
20 0.743583926754832
21 0.743583926754832
22 0.743583926754832
23 0.743583926754832
24 0.743583926754832
25 0.743583926754832
26 0.743583926754832
27 0.743583926754832
28 0.743583926754832
29 0.743583926754832
30 0.743583926754832
31 0.743583926754832
32 0.743583926754832
33 0.743583926754832
34 0.743583926754832
35 0.743583926754832
36 0.743583926754832
37 0.743583926754832
38 0.743583926754832
39 0.743583926754832
40 0.743583926754832
41 0.743583926754832
42 0.743583926754832
43 0.743583926754832
44 0.743583926754832
45 0.743583926754832
46 0.743583926754832
47 0.743583926754832
48 0.743583926754832
49 0.743583926754832
50 0.743583926754832
51 0.743583926754832
52 0.743583926754832
53 0.743583926754832
54 0.743583926754832
55 0.743583926754832
56 0.743583926754832
57 0.743583926754832
58 0.743583926754832
59 0.743583926754832
60 0.743583926754832
61 0.743583926754832
62 0.743583926754832
63 0.743583926754832
64 0.743583926754832
65 0.743583926754832
66 0.743583926754832
67 0.743583926754832
68 0.743583926754832
69 0.743583926754832
70 0.743583926754832
71 0.743583926754832
72 0.743583926754832
73 0.743583926754832
74 0.743583926754832
75 0.743583926754832
76 0.743583926754832
77 0.743583926754832
78 0.743583926754832
79 0.743583926754832
80 0.743583926754832
81 0.743583926754832
82 0.743583926754832
83 0.743583926754832
84 0.743583926754832
85 0.743583926754832
86 0.743583926754832
87 0.743583926754832
88 0.743583926754832
89 0.743583926754832
90 0.743583926754832
91 0.743583926754832
92 0.743583926754832
93 0.743583926754832
94 0.743583926754832
95 0.743583926754832
96 0.743583926754832
97 0.743583926754832
98 0.743583926754832
99 0.743583926754832
};
\addplot [semithick, color2, mark=o, mark size=1, mark options={solid}]
table {%
0 0.742706002034588
1 0.742706002034588
2 0.742706002034588
3 0.742706002034588
4 0.742706002034588
5 0.742706002034588
6 0.742706002034588
7 0.742706002034588
8 0.742706002034588
9 0.742706002034588
10 0.742706002034588
11 0.742706002034588
12 0.742706002034588
13 0.742706002034588
14 0.742706002034588
15 0.742706002034588
16 0.742706002034588
17 0.742706002034588
18 0.742706002034588
19 0.742706002034588
20 0.742706002034588
21 0.742706002034588
22 0.742706002034588
23 0.742706002034588
24 0.742706002034588
25 0.742706002034588
26 0.742706002034588
27 0.742706002034588
28 0.742706002034588
29 0.742706002034588
30 0.742706002034588
31 0.742706002034588
32 0.742706002034588
33 0.742706002034588
34 0.742706002034588
35 0.742706002034588
36 0.742706002034588
37 0.742706002034588
38 0.742706002034588
39 0.742706002034588
40 0.742706002034588
41 0.742706002034588
42 0.742706002034588
43 0.742706002034588
44 0.742706002034588
45 0.742706002034588
46 0.742706002034588
47 0.742706002034588
48 0.742706002034588
49 0.742706002034588
50 0.742706002034588
51 0.742706002034588
52 0.742706002034588
53 0.742706002034588
54 0.742706002034588
55 0.742706002034588
56 0.742706002034588
57 0.742706002034588
58 0.742706002034588
59 0.742706002034588
60 0.742706002034588
61 0.742706002034588
62 0.742706002034588
63 0.742706002034588
64 0.742706002034588
65 0.742706002034588
66 0.742706002034588
67 0.742706002034588
68 0.742706002034588
69 0.742706002034588
70 0.742706002034588
71 0.742706002034588
72 0.742706002034588
73 0.742706002034588
74 0.742706002034588
75 0.742706002034588
76 0.742706002034588
77 0.742706002034588
78 0.742706002034588
79 0.742706002034588
80 0.742706002034588
81 0.742706002034588
82 0.742706002034588
83 0.742706002034588
84 0.742706002034588
85 0.742706002034588
86 0.742706002034588
87 0.742706002034588
88 0.742706002034588
89 0.742706002034588
90 0.742706002034588
91 0.742706002034588
92 0.742706002034588
93 0.742706002034588
94 0.742706002034588
95 0.742706002034588
96 0.742706002034588
97 0.742706002034588
98 0.742706002034588
99 0.742706002034588
};
\addplot [semithick, color3, mark=o, mark size=1, mark options={solid}]
table {%
0 0.742166836215666
1 0.742166836215666
2 0.742166836215666
3 0.742166836215666
4 0.742166836215666
5 0.742166836215666
6 0.742166836215666
7 0.742166836215666
8 0.742166836215666
9 0.742166836215666
10 0.742166836215666
11 0.742166836215666
12 0.742166836215666
13 0.742166836215666
14 0.742166836215666
15 0.742166836215666
16 0.742166836215666
17 0.742166836215666
18 0.742166836215666
19 0.742166836215666
20 0.742166836215666
21 0.742166836215666
22 0.742166836215666
23 0.742166836215666
24 0.742166836215666
25 0.742166836215666
26 0.742166836215666
27 0.742166836215666
28 0.742166836215666
29 0.742166836215666
30 0.742166836215666
31 0.742166836215666
32 0.742166836215666
33 0.742166836215666
34 0.742166836215666
35 0.742166836215666
36 0.742166836215666
37 0.742166836215666
38 0.742166836215666
39 0.742166836215666
40 0.742166836215666
41 0.742166836215666
42 0.742166836215666
43 0.742166836215666
44 0.742166836215666
45 0.742166836215666
46 0.742166836215666
47 0.742166836215666
48 0.742166836215666
49 0.742166836215666
50 0.742166836215666
51 0.742166836215666
52 0.742166836215666
53 0.742166836215666
54 0.742166836215666
55 0.742166836215666
56 0.742166836215666
57 0.742166836215666
58 0.742166836215666
59 0.742166836215666
60 0.742166836215666
61 0.742166836215666
62 0.742166836215666
63 0.742166836215666
64 0.742166836215666
65 0.742166836215666
66 0.742166836215666
67 0.742166836215666
68 0.742166836215666
69 0.742166836215666
70 0.742166836215666
71 0.742166836215666
72 0.742166836215666
73 0.742166836215666
74 0.742166836215666
75 0.742166836215666
76 0.742166836215666
77 0.742166836215666
78 0.742166836215666
79 0.742166836215666
80 0.742166836215666
81 0.742166836215666
82 0.742166836215666
83 0.742166836215666
84 0.742166836215666
85 0.742166836215666
86 0.742166836215666
87 0.742166836215666
88 0.742166836215666
89 0.742166836215666
90 0.742166836215666
91 0.742166836215666
92 0.742166836215666
93 0.742166836215666
94 0.742166836215666
95 0.742166836215666
96 0.742166836215666
97 0.742166836215666
98 0.742166836215666
99 0.742166836215666
};
\end{axis}

\end{tikzpicture}}\\[0.3cm]					
                \scalebox{.7}{\input{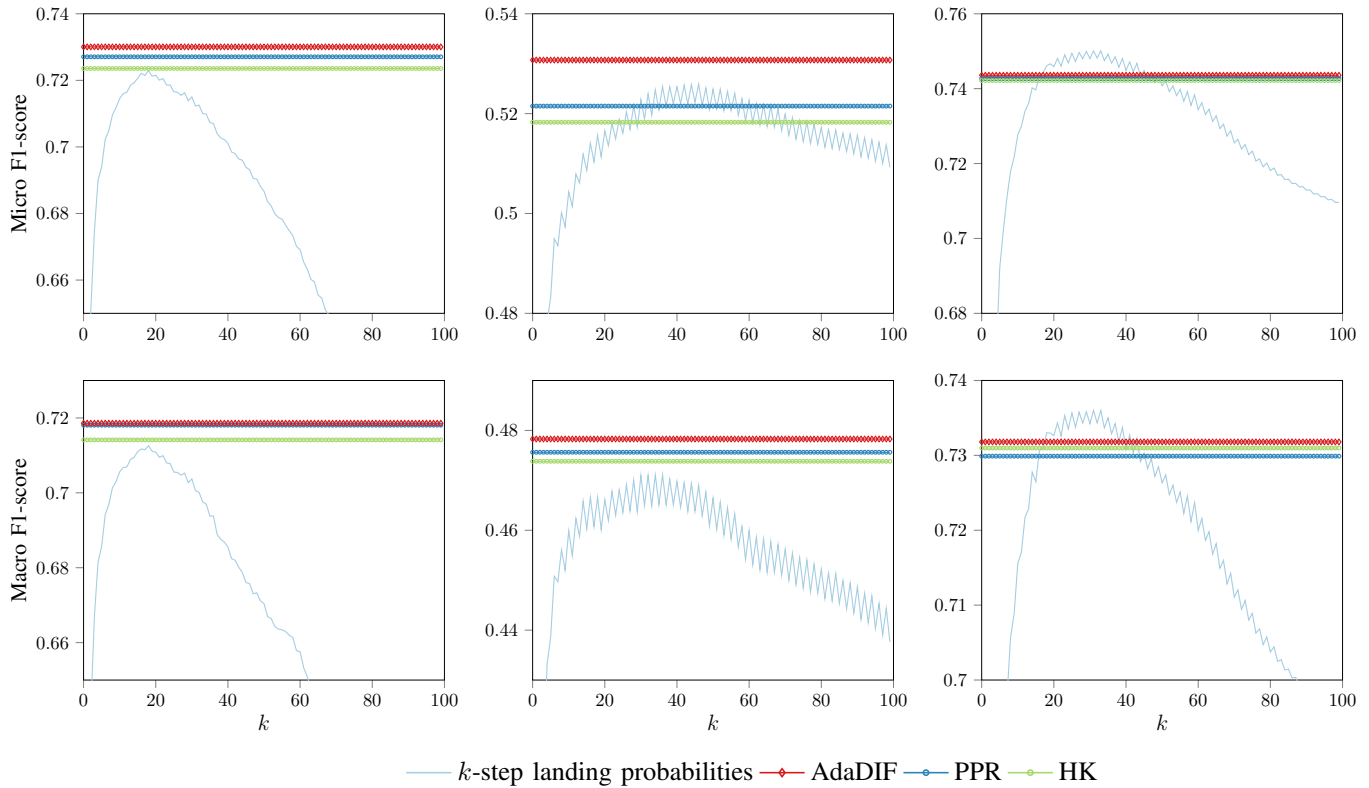}}
				\scalebox{.7}{
\begin{tikzpicture}

\definecolor{color1}{RGB}{215,25,28}%
\definecolor{color0}{RGB}{166,206,227}%
\definecolor{color2}{RGB}{43,131,186}%
\definecolor{color3}{RGB}{166,217,106}

\begin{axis}[
xlabel={\large $k$},
xmin=0,xmax=100,
ymin=0.43, ymax=0.49,
tick align=outside,
tick pos=left,
x grid style={lightgray!92.026143790849673!black},
y grid style={lightgray!92.026143790849673!black},
]
\addlegendimage{no markers, color0}
\addlegendimage{no markers, color1}
\addlegendimage{no markers, color2}
\addlegendimage{no markers, color3}
\addplot [semithick, color0]
table {%
0 0.153301340987163
1 0.272203082978034
2 0.369377791031998
3 0.409604191000797
4 0.433139154697513
5 0.43866062676227
6 0.450783009507912
7 0.449701266745906
8 0.45606587276608
9 0.452042656432347
10 0.459885859257201
11 0.454913397449054
12 0.462310831347078
13 0.458797638997607
14 0.465776742826365
15 0.46031039520556
16 0.466601533402863
17 0.460754046706113
18 0.466624787874547
19 0.460144732412666
20 0.466103848460369
21 0.461945178845025
22 0.467763609436915
23 0.463011486387746
24 0.468067688021137
25 0.463583970856807
26 0.469094487598496
27 0.46389378162572
28 0.470028480451189
29 0.463677006809892
30 0.470971613163886
31 0.464944470960379
32 0.470996992069379
33 0.464906009163864
34 0.471070299445016
35 0.465132047913158
36 0.471082003444346
37 0.465133960023236
38 0.470239810666156
39 0.465126367014667
40 0.469731521063917
41 0.464620629449273
42 0.469482146588758
43 0.463813745200838
44 0.469342817837144
45 0.463841404415377
46 0.468978766383869
47 0.462918647679482
48 0.467605598582262
49 0.460886416902608
50 0.466390942448679
51 0.460198179761891
52 0.465599888648387
53 0.459056414965722
54 0.464464015275861
55 0.458270326591406
56 0.463271052152978
57 0.456651995970276
58 0.460753737862515
59 0.454158249895283
60 0.459938159934431
61 0.453117773238707
62 0.458581760190891
63 0.452152401272101
64 0.458260892033595
65 0.451591555267089
66 0.457682153401661
67 0.451582031524633
68 0.457102536474365
69 0.451014779729176
70 0.456187260642752
71 0.450016213110382
72 0.455428465902595
73 0.449156666268937
74 0.454492185024678
75 0.448443961604049
76 0.453966861990558
77 0.447696035561309
78 0.45315337590797
79 0.447439068588709
80 0.452190848671896
81 0.446340500136008
82 0.45143530633121
83 0.445914557704026
84 0.451238096541081
85 0.445389268668889
86 0.449869332960308
87 0.444265890850716
88 0.44940121044978
89 0.443428922907009
90 0.448649538931667
91 0.442214608486417
92 0.447370658138828
93 0.441498699545111
94 0.446808021470433
95 0.440342098231908
96 0.445303971955953
97 0.439029244979995
98 0.444359061398745
99 0.437652226960393
};
\addplot [semithick, color1, mark=diamond, mark size=1.5, mark options={solid}]
table {%
0 0.478276809288431
1 0.478276809288431
2 0.478276809288431
3 0.478276809288431
4 0.478276809288431
5 0.478276809288431
6 0.478276809288431
7 0.478276809288431
8 0.478276809288431
9 0.478276809288431
10 0.478276809288431
11 0.478276809288431
12 0.478276809288431
13 0.478276809288431
14 0.478276809288431
15 0.478276809288431
16 0.478276809288431
17 0.478276809288431
18 0.478276809288431
19 0.478276809288431
20 0.478276809288431
21 0.478276809288431
22 0.478276809288431
23 0.478276809288431
24 0.478276809288431
25 0.478276809288431
26 0.478276809288431
27 0.478276809288431
28 0.478276809288431
29 0.478276809288431
30 0.478276809288431
31 0.478276809288431
32 0.478276809288431
33 0.478276809288431
34 0.478276809288431
35 0.478276809288431
36 0.478276809288431
37 0.478276809288431
38 0.478276809288431
39 0.478276809288431
40 0.478276809288431
41 0.478276809288431
42 0.478276809288431
43 0.478276809288431
44 0.478276809288431
45 0.478276809288431
46 0.478276809288431
47 0.478276809288431
48 0.478276809288431
49 0.478276809288431
50 0.478276809288431
51 0.478276809288431
52 0.478276809288431
53 0.478276809288431
54 0.478276809288431
55 0.478276809288431
56 0.478276809288431
57 0.478276809288431
58 0.478276809288431
59 0.478276809288431
60 0.478276809288431
61 0.478276809288431
62 0.478276809288431
63 0.478276809288431
64 0.478276809288431
65 0.478276809288431
66 0.478276809288431
67 0.478276809288431
68 0.478276809288431
69 0.478276809288431
70 0.478276809288431
71 0.478276809288431
72 0.478276809288431
73 0.478276809288431
74 0.478276809288431
75 0.478276809288431
76 0.478276809288431
77 0.478276809288431
78 0.478276809288431
79 0.478276809288431
80 0.478276809288431
81 0.478276809288431
82 0.478276809288431
83 0.478276809288431
84 0.478276809288431
85 0.478276809288431
86 0.478276809288431
87 0.478276809288431
88 0.478276809288431
89 0.478276809288431
90 0.478276809288431
91 0.478276809288431
92 0.478276809288431
93 0.478276809288431
94 0.478276809288431
95 0.478276809288431
96 0.478276809288431
97 0.478276809288431
98 0.478276809288431
99 0.478276809288431
};
\addplot [semithick, color2, mark=o, mark size=1, mark options={solid}]
table {%
0 0.475624515615459
1 0.475624515615459
2 0.475624515615459
3 0.475624515615459
4 0.475624515615459
5 0.475624515615459
6 0.475624515615459
7 0.475624515615459
8 0.475624515615459
9 0.475624515615459
10 0.475624515615459
11 0.475624515615459
12 0.475624515615459
13 0.475624515615459
14 0.475624515615459
15 0.475624515615459
16 0.475624515615459
17 0.475624515615459
18 0.475624515615459
19 0.475624515615459
20 0.475624515615459
21 0.475624515615459
22 0.475624515615459
23 0.475624515615459
24 0.475624515615459
25 0.475624515615459
26 0.475624515615459
27 0.475624515615459
28 0.475624515615459
29 0.475624515615459
30 0.475624515615459
31 0.475624515615459
32 0.475624515615459
33 0.475624515615459
34 0.475624515615459
35 0.475624515615459
36 0.475624515615459
37 0.475624515615459
38 0.475624515615459
39 0.475624515615459
40 0.475624515615459
41 0.475624515615459
42 0.475624515615459
43 0.475624515615459
44 0.475624515615459
45 0.475624515615459
46 0.475624515615459
47 0.475624515615459
48 0.475624515615459
49 0.475624515615459
50 0.475624515615459
51 0.475624515615459
52 0.475624515615459
53 0.475624515615459
54 0.475624515615459
55 0.475624515615459
56 0.475624515615459
57 0.475624515615459
58 0.475624515615459
59 0.475624515615459
60 0.475624515615459
61 0.475624515615459
62 0.475624515615459
63 0.475624515615459
64 0.475624515615459
65 0.475624515615459
66 0.475624515615459
67 0.475624515615459
68 0.475624515615459
69 0.475624515615459
70 0.475624515615459
71 0.475624515615459
72 0.475624515615459
73 0.475624515615459
74 0.475624515615459
75 0.475624515615459
76 0.475624515615459
77 0.475624515615459
78 0.475624515615459
79 0.475624515615459
80 0.475624515615459
81 0.475624515615459
82 0.475624515615459
83 0.475624515615459
84 0.475624515615459
85 0.475624515615459
86 0.475624515615459
87 0.475624515615459
88 0.475624515615459
89 0.475624515615459
90 0.475624515615459
91 0.475624515615459
92 0.475624515615459
93 0.475624515615459
94 0.475624515615459
95 0.475624515615459
96 0.475624515615459
97 0.475624515615459
98 0.475624515615459
99 0.475624515615459
};
\addplot [semithick, color3, mark=o, mark size=1, mark options={solid}]
table {%
0 0.473815774145439
1 0.473815774145439
2 0.473815774145439
3 0.473815774145439
4 0.473815774145439
5 0.473815774145439
6 0.473815774145439
7 0.473815774145439
8 0.473815774145439
9 0.473815774145439
10 0.473815774145439
11 0.473815774145439
12 0.473815774145439
13 0.473815774145439
14 0.473815774145439
15 0.473815774145439
16 0.473815774145439
17 0.473815774145439
18 0.473815774145439
19 0.473815774145439
20 0.473815774145439
21 0.473815774145439
22 0.473815774145439
23 0.473815774145439
24 0.473815774145439
25 0.473815774145439
26 0.473815774145439
27 0.473815774145439
28 0.473815774145439
29 0.473815774145439
30 0.473815774145439
31 0.473815774145439
32 0.473815774145439
33 0.473815774145439
34 0.473815774145439
35 0.473815774145439
36 0.473815774145439
37 0.473815774145439
38 0.473815774145439
39 0.473815774145439
40 0.473815774145439
41 0.473815774145439
42 0.473815774145439
43 0.473815774145439
44 0.473815774145439
45 0.473815774145439
46 0.473815774145439
47 0.473815774145439
48 0.473815774145439
49 0.473815774145439
50 0.473815774145439
51 0.473815774145439
52 0.473815774145439
53 0.473815774145439
54 0.473815774145439
55 0.473815774145439
56 0.473815774145439
57 0.473815774145439
58 0.473815774145439
59 0.473815774145439
60 0.473815774145439
61 0.473815774145439
62 0.473815774145439
63 0.473815774145439
64 0.473815774145439
65 0.473815774145439
66 0.473815774145439
67 0.473815774145439
68 0.473815774145439
69 0.473815774145439
70 0.473815774145439
71 0.473815774145439
72 0.473815774145439
73 0.473815774145439
74 0.473815774145439
75 0.473815774145439
76 0.473815774145439
77 0.473815774145439
78 0.473815774145439
79 0.473815774145439
80 0.473815774145439
81 0.473815774145439
82 0.473815774145439
83 0.473815774145439
84 0.473815774145439
85 0.473815774145439
86 0.473815774145439
87 0.473815774145439
88 0.473815774145439
89 0.473815774145439
90 0.473815774145439
91 0.473815774145439
92 0.473815774145439
93 0.473815774145439
94 0.473815774145439
95 0.473815774145439
96 0.473815774145439
97 0.473815774145439
98 0.473815774145439
99 0.473815774145439
};
\end{axis}

\end{tikzpicture}}		
                \scalebox{.7}{
\begin{tikzpicture}

\definecolor{color1}{RGB}{215,25,28}%
\definecolor{color0}{RGB}{166,206,227}%
\definecolor{color2}{RGB}{43,131,186}%
\definecolor{color3}{RGB}{166,217,106}

\begin{axis}[
xlabel={\large $k$},
xmin=0,xmax=100,
ymin=0.70, ymax=0.74,
tick align=outside,
tick pos=left,
x grid style={lightgray!92.026143790849673!black},
y grid style={lightgray!92.026143790849673!black},
]
\addplot [semithick, color0]
table {%
0 0.200518002824947
1 0.285718117712408
2 0.440723097317838
3 0.60150730158398
4 0.650776762705169
5 0.677883300803084
6 0.689594812286107
7 0.697556412830996
8 0.705708026849921
9 0.708865350680046
10 0.71557406932307
11 0.717063430019054
12 0.721710668939292
13 0.722866374228136
14 0.727749756424734
15 0.726486920920302
16 0.730658062469462
17 0.730864225548804
18 0.733019895020821
19 0.732972304492462
20 0.732634220047007
21 0.734328997148399
22 0.732605632251985
23 0.735295953736512
24 0.733507256752448
25 0.73564646065309
26 0.733751472515656
27 0.735771187667185
28 0.734310308079135
29 0.735821848219191
30 0.73423751329875
31 0.735998300822676
32 0.734337541753623
33 0.735981270972545
34 0.733672517884631
35 0.7350095769065
36 0.732887541398444
37 0.734225179752073
38 0.731768450490577
39 0.73370218198517
40 0.730801170865909
41 0.732715097269709
42 0.73037603624724
43 0.732225589708908
44 0.729141010846278
45 0.730275194702761
46 0.728215957657299
47 0.729551233794605
48 0.727199022072702
49 0.728474966082512
50 0.726334232160646
51 0.727394446957278
52 0.724745042854257
53 0.725990276927827
54 0.723737370813701
55 0.725095243505123
56 0.722255800755671
57 0.724545915993433
58 0.721158313084121
59 0.72367999794911
60 0.719990683671822
61 0.72155050387118
62 0.718653237206251
63 0.719795513328849
64 0.716805904179554
65 0.718197167946051
66 0.714927895687107
67 0.715974803297374
68 0.712845372165069
69 0.714230264066798
70 0.711072732314492
71 0.712037326966497
72 0.709438948861399
73 0.710617574984577
74 0.708012352026774
75 0.708892301344322
76 0.70625603249348
77 0.706846598545948
78 0.704727544031606
79 0.705744524461424
80 0.70374364845795
81 0.704438807881297
82 0.702497329451136
83 0.702717244512824
84 0.701347150634884
85 0.701458349799558
86 0.700302068193995
87 0.700328297692557
88 0.699413925361847
89 0.699478735444232
90 0.698610810633981
91 0.698681009689709
92 0.697584963343289
93 0.697761928634181
94 0.696905371321139
95 0.696969503831793
96 0.696095838395147
97 0.696271932526229
98 0.695418941907461
99 0.69546652618679
};
\addplot [semithick, color1, mark=diamond, mark size=1.5, mark options={solid}]
table {%
0 0.731792036076864
1 0.731792036076864
2 0.731792036076864
3 0.731792036076864
4 0.731792036076864
5 0.731792036076864
6 0.731792036076864
7 0.731792036076864
8 0.731792036076864
9 0.731792036076864
10 0.731792036076864
11 0.731792036076864
12 0.731792036076864
13 0.731792036076864
14 0.731792036076864
15 0.731792036076864
16 0.731792036076864
17 0.731792036076864
18 0.731792036076864
19 0.731792036076864
20 0.731792036076864
21 0.731792036076864
22 0.731792036076864
23 0.731792036076864
24 0.731792036076864
25 0.731792036076864
26 0.731792036076864
27 0.731792036076864
28 0.731792036076864
29 0.731792036076864
30 0.731792036076864
31 0.731792036076864
32 0.731792036076864
33 0.731792036076864
34 0.731792036076864
35 0.731792036076864
36 0.731792036076864
37 0.731792036076864
38 0.731792036076864
39 0.731792036076864
40 0.731792036076864
41 0.731792036076864
42 0.731792036076864
43 0.731792036076864
44 0.731792036076864
45 0.731792036076864
46 0.731792036076864
47 0.731792036076864
48 0.731792036076864
49 0.731792036076864
50 0.731792036076864
51 0.731792036076864
52 0.731792036076864
53 0.731792036076864
54 0.731792036076864
55 0.731792036076864
56 0.731792036076864
57 0.731792036076864
58 0.731792036076864
59 0.731792036076864
60 0.731792036076864
61 0.731792036076864
62 0.731792036076864
63 0.731792036076864
64 0.731792036076864
65 0.731792036076864
66 0.731792036076864
67 0.731792036076864
68 0.731792036076864
69 0.731792036076864
70 0.731792036076864
71 0.731792036076864
72 0.731792036076864
73 0.731792036076864
74 0.731792036076864
75 0.731792036076864
76 0.731792036076864
77 0.731792036076864
78 0.731792036076864
79 0.731792036076864
80 0.731792036076864
81 0.731792036076864
82 0.731792036076864
83 0.731792036076864
84 0.731792036076864
85 0.731792036076864
86 0.731792036076864
87 0.731792036076864
88 0.731792036076864
89 0.731792036076864
90 0.731792036076864
91 0.731792036076864
92 0.731792036076864
93 0.731792036076864
94 0.731792036076864
95 0.731792036076864
96 0.731792036076864
97 0.731792036076864
98 0.731792036076864
99 0.731792036076864
};
\addplot [semithick, color2, mark=o, mark size=1, mark options={solid}]
table {%
0 0.729889970505886
1 0.729889970505886
2 0.729889970505886
3 0.729889970505886
4 0.729889970505886
5 0.729889970505886
6 0.729889970505886
7 0.729889970505886
8 0.729889970505886
9 0.729889970505886
10 0.729889970505886
11 0.729889970505886
12 0.729889970505886
13 0.729889970505886
14 0.729889970505886
15 0.729889970505886
16 0.729889970505886
17 0.729889970505886
18 0.729889970505886
19 0.729889970505886
20 0.729889970505886
21 0.729889970505886
22 0.729889970505886
23 0.729889970505886
24 0.729889970505886
25 0.729889970505886
26 0.729889970505886
27 0.729889970505886
28 0.729889970505886
29 0.729889970505886
30 0.729889970505886
31 0.729889970505886
32 0.729889970505886
33 0.729889970505886
34 0.729889970505886
35 0.729889970505886
36 0.729889970505886
37 0.729889970505886
38 0.729889970505886
39 0.729889970505886
40 0.729889970505886
41 0.729889970505886
42 0.729889970505886
43 0.729889970505886
44 0.729889970505886
45 0.729889970505886
46 0.729889970505886
47 0.729889970505886
48 0.729889970505886
49 0.729889970505886
50 0.729889970505886
51 0.729889970505886
52 0.729889970505886
53 0.729889970505886
54 0.729889970505886
55 0.729889970505886
56 0.729889970505886
57 0.729889970505886
58 0.729889970505886
59 0.729889970505886
60 0.729889970505886
61 0.729889970505886
62 0.729889970505886
63 0.729889970505886
64 0.729889970505886
65 0.729889970505886
66 0.729889970505886
67 0.729889970505886
68 0.729889970505886
69 0.729889970505886
70 0.729889970505886
71 0.729889970505886
72 0.729889970505886
73 0.729889970505886
74 0.729889970505886
75 0.729889970505886
76 0.729889970505886
77 0.729889970505886
78 0.729889970505886
79 0.729889970505886
80 0.729889970505886
81 0.729889970505886
82 0.729889970505886
83 0.729889970505886
84 0.729889970505886
85 0.729889970505886
86 0.729889970505886
87 0.729889970505886
88 0.729889970505886
89 0.729889970505886
90 0.729889970505886
91 0.729889970505886
92 0.729889970505886
93 0.729889970505886
94 0.729889970505886
95 0.729889970505886
96 0.729889970505886
97 0.729889970505886
98 0.729889970505886
99 0.729889970505886
};
\addplot [semithick, color3, mark=o, mark size=1, mark options={solid}]
table {%
0 0.730978079102848
1 0.730978079102848
2 0.730978079102848
3 0.730978079102848
4 0.730978079102848
5 0.730978079102848
6 0.730978079102848
7 0.730978079102848
8 0.730978079102848
9 0.730978079102848
10 0.730978079102848
11 0.730978079102848
12 0.730978079102848
13 0.730978079102848
14 0.730978079102848
15 0.730978079102848
16 0.730978079102848
17 0.730978079102848
18 0.730978079102848
19 0.730978079102848
20 0.730978079102848
21 0.730978079102848
22 0.730978079102848
23 0.730978079102848
24 0.730978079102848
25 0.730978079102848
26 0.730978079102848
27 0.730978079102848
28 0.730978079102848
29 0.730978079102848
30 0.730978079102848
31 0.730978079102848
32 0.730978079102848
33 0.730978079102848
34 0.730978079102848
35 0.730978079102848
36 0.730978079102848
37 0.730978079102848
38 0.730978079102848
39 0.730978079102848
40 0.730978079102848
41 0.730978079102848
42 0.730978079102848
43 0.730978079102848
44 0.730978079102848
45 0.730978079102848
46 0.730978079102848
47 0.730978079102848
48 0.730978079102848
49 0.730978079102848
50 0.730978079102848
51 0.730978079102848
52 0.730978079102848
53 0.730978079102848
54 0.730978079102848
55 0.730978079102848
56 0.730978079102848
57 0.730978079102848
58 0.730978079102848
59 0.730978079102848
60 0.730978079102848
61 0.730978079102848
62 0.730978079102848
63 0.730978079102848
64 0.730978079102848
65 0.730978079102848
66 0.730978079102848
67 0.730978079102848
68 0.730978079102848
69 0.730978079102848
70 0.730978079102848
71 0.730978079102848
72 0.730978079102848
73 0.730978079102848
74 0.730978079102848
75 0.730978079102848
76 0.730978079102848
77 0.730978079102848
78 0.730978079102848
79 0.730978079102848
80 0.730978079102848
81 0.730978079102848
82 0.730978079102848
83 0.730978079102848
84 0.730978079102848
85 0.730978079102848
86 0.730978079102848
87 0.730978079102848
88 0.730978079102848
89 0.730978079102848
90 0.730978079102848
91 0.730978079102848
92 0.730978079102848
93 0.730978079102848
94 0.730978079102848
95 0.730978079102848
96 0.730978079102848
97 0.730978079102848
98 0.730978079102848
99 0.730978079102848
};
\end{axis}

\end{tikzpicture}}\\[0.2cm]
                \phantom{ppppppppppppppppppppppppppppp} \ref{named2}
  \caption{Classification accuracy of AdaDIF, PPR, and HK compared to the accuracy of $k-$step landing probability classifier. \textbf{Top Left)} \texttt{Cora}  Micro-F1 score; \textbf{Bottom Left)} \texttt{Cora}  Macro-F1 score; \textbf{Top Middle)} \texttt{Citeseer} Micro-F1 score; \textbf{Bottom Middle)} \texttt{Citeseer} Macro-F1 score; \textbf{Top Right)} \texttt{PubMed} Micro-F1 score; \textbf{Bottom Right)} \texttt{PubMed} Macro-F1 score   }
  \label{per_step}
\end{figure*}

\begin{figure}[!tbp]
  \centering
 \input{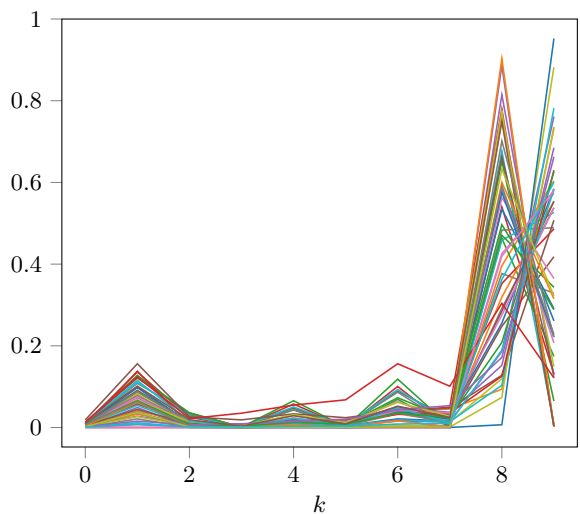}	
  \caption{AdaDIF diffusion coefficients for the $50$ different classes of PPI graph ($30\%$ sampled). Each line corresponds to a different $\boldsymbol{\theta}_c$. Diffusion is characterized by high diversity among classes.   }
  \label{per_class}
\end{figure}


The scalabilty of AdaDIF is reflected on the runtime comparisons listed in Fig. \ref{runtime}. All experiments were run on a machine with i5 @3.50 Mhz CPU, and 16GB of RAM. We used the Python implementations provided by the authors of the compared algorithms. The Python implementation of AdaDIF, uses only tools provided by scipy, {numpy}, and {CVX-OPT} libraries. We also developped an efficient implementation 
that exploits parallelism, which is straightforward since each class can be treated separately.  { While AdaDIF incurs (as expected) a relatively small computational overhead over fixed diffusions, it is faster than GCNs that use Tensorflow, and orders of magnitude faster than embedding-based approaches.}

Finally, Table \ref{tab:Micro_mlabel} presents the results on multilabel graphs, where we  compare with Deepwalk and Node2vec, since the rest of the methods are designed for multiclass problems. Since these graphs entail a large number of classes, we increased the number of training samples. Similar to \cite{grover2016node2vec} and \cite{perozzi2014deepwalk}, during evaluation of accuracy the number of labels per sampled node is known, and check how many of them are in the top predictions. First, we observe that AdaDIF markedly outperforms PPR and HK across graphs and metrics. Furthermore, for the \texttt{PPI} and \texttt{BlogCatalog} graphs  the Micro-F1 score of AdaDIF comes close to that of the much heavier state-of-the-art Node2vec. Finally, AdaDIF outperforms the competing alternatives in terms of Macro-F1 score.   { It is worth noting that for  multilabel graphs with many classes, the performance boost over fixed diffusions can  be largely attributed to AdaDif's flexibility to treat each class differently. To demonstrate that different classes are indeed diffused in a markedly different manner, Fig. \ref{per_class}  plots all $50$ diffusion coefficient vectors $\{\boldsymbol{\theta}_c\}_{c\in\mathcal{C}}$ yielded by AdaDIF on the \texttt{PPI} graph with $30\%$ of nodes labeled. Each line in the plot corresponds to the values of $\boldsymbol{\theta}_c$ for a different $c$;  evidently, while the overall ``form'' of the corresponding diffusion coefficients adheres to the general pattern observed in Fig.\ref{thetas}  there is indeed large diversity among classes.}

\setlength\figW{0.85\columnwidth}
\setlength\figH{0.53\columnwidth}
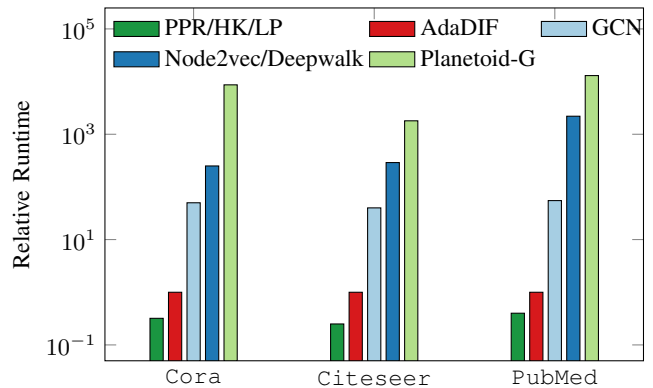
\begin{figure}[t!] 
	\small
%
%
\definecolor{mycolor1}{RGB}{215,25,28}%
\definecolor{mycolor2}{RGB}{166,206,227}%
\definecolor{mycolor3}{RGB}{31,120,180}%
\definecolor{mycolor4}{RGB}{178,223,138}
\definecolor{mycolor5}{RGB}{26,150,65}
\begin{tikzpicture}

\begin{axis}[%
width=0.951\figW,
height=\figH,
at={(0\figW,0\figH)},
scale only axis,
bar shift auto,
log origin=infty,
xmin=0.509090909090909,
xmax=3.49090909090909,
xtick={1,2,3},
xticklabels={{\texttt{Cora}},{\texttt{Citeseer}},{\texttt{PubMed}}},
ymode=log,
ymin=0.05,
ymax=250000,
yminorticks=true,
ylabel style={font=\color{white!15!black}},
ylabel={Relative Runtime},
ylabel near ticks,
axis background/.style={fill=white},
 legend columns=3,
legend style={at={(0,0.795)}, anchor=south west, legend cell align=left, align=left, draw=none, fill=none}
]

\addplot[ybar, bar width=5, fill=mycolor5, draw=black, area legend] table[row sep=crcr] {%
1	0.32\\
2	0.25\\
3	0.4\\
};
\addlegendentry{PPR/HK/LP}

\addplot[ybar, bar width=5, fill=mycolor1, draw=black, area legend] table[row sep=crcr] {%
	1	1\\
	2	1\\
	3	1\\
};
\addlegendentry{AdaDIF}

\addplot[ybar, bar width=5, fill=mycolor2, draw=black, area legend] table[row sep=crcr] {%
	1	50\\
	2	40\\
	3	55\\
};
\addlegendentry{GCN}

\addplot[ybar, bar width=5, fill=mycolor3, draw=black, area legend] table[row sep=crcr] {%
	1	250\\
	2	290\\
	3	2200\\
};
\addlegendentry{Node2vec/Deepwalk}

\addplot[ybar,  bar width=5, fill=mycolor4, draw=black, area legend] table[row sep=crcr] {%
	1	8672\\
	2	1800\\
	3	13000\\
};
\addlegendentry{Planetoid-G}

%
%
%

\end{axis}
\end{tikzpicture}%
	\caption{Relative runtime comparisons for multiclass graphs.} 
	\vspace{-0.5cm}
	\label{runtime}
\end{figure}

\setlength\figW{0.85\columnwidth}
\setlength\figH{0.53\columnwidth}
\begin{figure*}[t!] 
	\small
%
%
\definecolor{mycolor1}{RGB}{215,25,28}%
\definecolor{mycolor2}{RGB}{43,131,186}%
\definecolor{mycolor3}{RGB}{166,217,106}
\definecolor{mycolor4}{RGB}{253,174,97}
\definecolor{mycolor5}{RGB}{26,150,65}
\begin{tikzpicture}

\begin{axis}[%
width=0.951\figW,
height=\figH,
at={(0\figW,0\figH)},
scale only axis,
xmin=0,
xmax=0.355,
xtick={0, .05, 0.1, 0.15, 0.2, 0.25, 0.3, 0.35},
xticklabels={$0$, $0.05$, $0.1$, $0.15$, $0.2$, $0.25$, $0.3$, $0.35$ },
xlabel style={font=\color{white!15!black}},
xlabel={Label corruption rate $p_{\text{cor}}$},
ymin=59,
ymax=74,
ylabel style={font=\color{white!15!black}},
ylabel={Micro-F1 Score (\%)},
ylabel near ticks,
axis background/.style={fill=white},
legend style={at={(0.038,0.038)}, draw=none, fill=none, anchor=south west, legend cell align=left, align=left}
]
\addplot [color=mycolor4, line width=1.0pt, mark size=3.0pt, mark=+, mark options={solid, mycolor4}]
  table[row sep=crcr]{%
0	71.3\\
0.05	71.2\\
0.1	70\\
0.15	68.4\\
0.2	67.2\\
0.25	66.1\\
0.3	65.3\\
0.35	63.4\\
};
\addlegendentry{r-AdaDIF}

\addplot [color=mycolor1, line width=1.0pt, mark size=2pt, mark=square, mark options={solid, mycolor1}]
  table[row sep=crcr]{%
0	73\\
0.05	71\\
0.1	69.5\\
0.15	67.4\\
0.2	66\\
0.25	64.6\\
0.3	63\\
0.35	61.6\\
};
\addlegendentry{AdaDIF}

\addplot [color=mycolor3, line width=1.0pt, mark=diamond, mark options={solid, mycolor3}]
  table[row sep=crcr]{%
0	72.4\\
0.05	70.9\\
0.1	69.1\\
0.15	66.5\\
0.2	64.5\\
0.25	63.3\\
0.3	61.6\\
0.35	59.8\\
};
\addlegendentry{PPR}

\addplot [color=mycolor2, line width=1.0pt, mark size=2pt, mark=o, mark options={solid, mycolor2}]
  table[row sep=crcr]{%
0	72.6\\
0.05	70.8\\
0.1	69.3\\
0.15	67\\
0.2	65.3\\
0.25	64.4\\
0.3	62.7\\
0.35	60.9\\
};
\addlegendentry{HK}

\addplot [color=mycolor5, line width=1.0pt, mark size=2pt, mark=*, mark options={solid, mycolor5}]
  table[row sep=crcr]{%
0	69.5\\
0.05	68.6\\
0.1	67.5\\
0.15	65.5\\
0.2	63.5\\
0.25	62.2\\
0.3	58.5\\
0.35	56.2\\
};
\addlegendentry{PPR w. ranking}

\end{axis}
\end{tikzpicture}%
	\hspace{2em}
	\centering	
%
%
\definecolor{mycolor1}{RGB}{215,25,28}%
\definecolor{mycolor2}{RGB}{43,131,186}%
\definecolor{mycolor3}{RGB}{166,217,106}
\definecolor{mycolor4}{RGB}{253,174,97}
\definecolor{mycolor5}{RGB}{26,150,65}
\begin{tikzpicture}

\begin{axis}[%
width=0.951\figW,
height=\figH,
at={(0\figW,0\figH)},
scale only axis,
xmin=0,
xmax=0.355,
xtick={0, .05, 0.1, 0.15, 0.2, 0.25, 0.3, 0.35},
xticklabels={$0$, $0.05$, $0.1$, $0.15$, $0.2$, $0.25$, $0.3$, $0.35$ },
xlabel style={font=\color{white!15!black}},
xlabel={Label corruption rate $p_{\text{cor}}$},
ymin=58,
ymax=72,
ylabel style={font=\color{white!15!black}},
ylabel near ticks,
ylabel={Macro-F1 Score (\%)},
axis background/.style={fill=white},
legend style={at={(0.038,0.038)}, anchor=south west, draw=none, fill=none, legend cell align=left, align=left,}
]
\addplot [color=mycolor4, line width=1.0pt, mark size=4.0pt, mark=+, mark options={solid, mycolor4}]
  table[row sep=crcr]{%
0	70\\
0.05	69.5\\
0.1	68.7\\
0.15	66.8\\
0.2	65.5\\
0.25	64.4\\
0.3	63.6\\
0.35	61.5\\
};
\addlegendentry{r-AdaDIF}

\addplot [color=mycolor1, line width=1.0pt, mark size=2pt, mark=square, mark options={solid, mycolor1}]
  table[row sep=crcr]{%
0	72\\
0.05	69\\
0.1	67.7\\
0.15	65.2\\
0.2	64.1\\
0.25	62.5\\
0.3	60\\
0.35	59\\
};
\addlegendentry{AdaDIF}

\addplot [color=mycolor3, line width=1.0pt, mark size=2pt, mark=diamond, mark options={solid, mycolor3}]
  table[row sep=crcr]{%
0	71.5\\
0.05	69.3\\
0.1	68\\
0.15	65.1\\
0.2	63.3\\
0.25	61.7\\
0.3	59.8\\
0.35	58.1\\
};
\addlegendentry{PPR}

\addplot [color=mycolor2, line width=1.0pt, mark size=2pt, mark=o, mark options={solid, mycolor2}]
  table[row sep=crcr]{%
0	72\\
0.05	69.4\\
0.1	68.3\\
0.15	65.6\\
0.2	64.1\\
0.25	62.9\\
0.3	61\\
0.35	59.3\\
};
\addlegendentry{HK}

\addplot [color=mycolor5, line width=1.0pt, mark size=2pt, mark=*, mark options={solid, mycolor5}]
  table[row sep=crcr]{%
0	68.2\\
0.05	67.5\\
0.1	66.3\\
0.15	64.5\\
0.2	62.2\\
0.25	60.5\\
0.3	 57.5\\
0.35	54.7\\
};
\addlegendentry{PPR w. ranking}

\end{axis}
\end{tikzpicture}%
	\caption{Classification accuracy of various diffusion-based classifiers on Cora, as a function of the probability of label corruption.} 
	\label{fig:cora_rob_micro}
\end{figure*}

\begin{figure}[t!] 
	\small
%
%
\definecolor{mycolor1}{RGB}{186,228,188}%
\definecolor{mycolor2}{RGB}{123,204,196}%
\definecolor{mycolor3}{RGB}{43,140,190}
\begin{tikzpicture}

\begin{axis}[%
width=0.951\figW,
height=\figH,
at={(0\figW,0\figH)},
scale only axis,
xmin=0,
xmax=1,
xlabel style={font=\color{white!15!black}},
xlabel={$p_{\textsc{fa}}$},
ymin=0,
ymax=1,
ylabel style={font=\color{white!15!black}},
ylabel near ticks,
ylabel={$p_{\textsc{d}}$},
axis background/.style={fill=white},
xmajorgrids,
ymajorgrids,
legend style={at={(0.613,0.038)}, anchor=south west, legend cell align=left, align=left, draw=none}
]
\addplot [color=mycolor3, line width=1.0pt, mark size=2.0pt, mark=+, mark options={solid, mycolor3}]
  table[row sep=crcr]{%
0	0\\
0.01	0.054\\
0.016	0.063\\
0.027	0.153\\
0.108	0.242\\
0.126	0.426\\
0.164	0.55\\
0.227	0.663\\
0.27	0.72\\
0.35	0.848\\
0.46	0.91\\
0.725	0.977\\
0.9	0.995\\
1	1\\
};
\addlegendentry{$p_{\text{cor}} = 0.35$}

\addplot [color=mycolor2, line width=1.0pt, mark size=2.0pt, mark=o, mark options={solid, mycolor2}]
  table[row sep=crcr]{%
0	0\\
0.01	0.065\\
0.012	0.085\\
0.014	0.18\\
0.1	0.415\\
0.12	0.523\\
0.164	0.7\\
0.183	0.735\\
0.212	0.823\\
0.31	0.9\\
0.39	0.94\\
0.65	0.978\\
0.85	0.998\\
1	1\\
};
\addlegendentry{$p_{\text{cor}} = 0.15$}

\addplot [color=mycolor1, line width=1.0pt, mark size=2pt, mark=square, mark options={solid, mycolor1}]
  table[row sep=crcr]{%
0	0\\
0.008	0.09\\
0.009	0.1\\
0.012	0.156\\
0.0136	0.24\\
0.09	0.485\\
0.113	0.57\\
0.177	0.82\\
0.21	0.84\\
0.28	0.9\\
0.38	0.925\\
0.5	0.97\\
0.624	0.975\\
0.83	0.985\\
1	1\\
};
\addlegendentry{$p_{\text{cor}} = 0.05$}

\addplot [color=gray, dashed, forget plot]
  table[row sep=crcr]{%
0.01	0.01\\
0.02	0.02\\
0.03	0.03\\
0.04	0.04\\
0.05	0.05\\
0.06	0.06\\
0.07	0.07\\
0.08	0.08\\
0.09	0.09\\
0.1	0.1\\
0.11	0.11\\
0.12	0.12\\
0.13	0.13\\
0.14	0.14\\
0.15	0.15\\
0.16	0.16\\
0.17	0.17\\
0.18	0.18\\
0.19	0.19\\
0.2	0.2\\
0.21	0.21\\
0.22	0.22\\
0.23	0.23\\
0.24	0.24\\
0.25	0.25\\
0.26	0.26\\
0.27	0.27\\
0.28	0.28\\
0.29	0.29\\
0.3	0.3\\
0.31	0.31\\
0.32	0.32\\
0.33	0.33\\
0.34	0.34\\
0.35	0.35\\
0.36	0.36\\
0.37	0.37\\
0.38	0.38\\
0.39	0.39\\
0.4	0.4\\
0.41	0.41\\
0.42	0.42\\
0.43	0.43\\
0.44	0.44\\
0.45	0.45\\
0.46	0.46\\
0.47	0.47\\
0.48	0.48\\
0.49	0.49\\
0.5	0.5\\
0.51	0.51\\
0.52	0.52\\
0.53	0.53\\
0.54	0.54\\
0.55	0.55\\
0.56	0.56\\
0.57	0.57\\
0.58	0.58\\
0.59	0.59\\
0.6	0.6\\
0.61	0.61\\
0.62	0.62\\
0.63	0.63\\
0.64	0.64\\
0.65	0.65\\
0.66	0.66\\
0.67	0.67\\
0.68	0.68\\
0.69	0.69\\
0.7	0.7\\
0.71	0.71\\
0.72	0.72\\
0.73	0.73\\
0.74	0.74\\
0.75	0.75\\
0.76	0.76\\
0.77	0.77\\
0.78	0.78\\
0.79	0.79\\
0.8	0.8\\
0.81	0.81\\
0.82	0.82\\
0.83	0.83\\
0.84	0.84\\
0.85	0.85\\
0.86	0.86\\
0.87	0.87\\
0.88	0.88\\
0.89	0.89\\
0.9	0.9\\
0.91	0.91\\
0.92	0.92\\
0.93	0.93\\
0.94	0.94\\
0.95	0.95\\
0.96	0.96\\
0.97	0.97\\
0.98	0.98\\
0.99	0.99\\
1	1\\
};
\end{axis}
\end{tikzpicture}%
	\caption{Anomaly detection performance of r-AdaDIF for different label corruption probabilities. The horizontal axis corresponds to the frequency with which r-AdaDIF returns a true positive (probability of detection) and the vertical axis corresponds to the frequency of false positives (probability of false alarm).} 
	\label{fig:roc}
\end{figure}
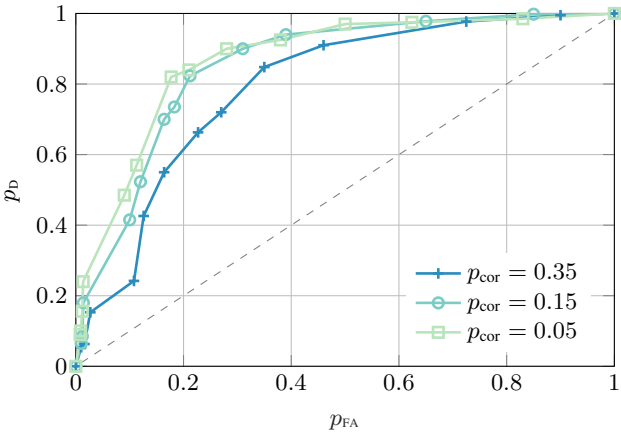

 {
\subsection{Analysis/interpretation of results}
Here we will follow an experimental approach that is aimed at understanding and interpreting our results. We will focus on diffusion-based classifiers, along with a simple benchmark for diffusion-based classification: the $k-$step landing probabilities. Specifically, we compare the classification accuracy on the three multiclass datasets of AdaDIF, PPR, and HK, with the accuracy of the classifier that uses only the $k-$th landing probability vectors $\{\mathbf{p}_c^{(k)}\}_{c\in\mathcal{Y},k\in[1,K]}$. The setting is similar to the one in the previous section, and with class-balanced sampling of $20$ nodes per class, while the $k-$step classifiers were examined for a wide range of steps $k\in[1,100]$. The $k-$step classifier reveals the predictive power of individual landing probabilities, resulting in curves (see Fig. \ref{per_step}) that appear to be different for each network, characterizing the graph-label distribution relationship of the latter. For the \texttt{Cora}  graph (left two plots), performance of the $k-$step classifier improves sharply after the first few steps, peaks for $ k\approx20$, and then quickly degrades, suggesting that using the landing probabilities of $k > 40$ or $50$ would most likely degrade the performance of a diffusion-based classifier. Interestingly, AdaDIF relying on combinations of the first 15 steps, and PPR and HK of the first 50, all achieve higher accuracy than that of the best single step.  On the other hand, the \texttt{Citeseer} graph (middle two plots) behaves in a significantly different manner, with the $k-$step classifier requiring longer walks to reach high accuracy that was retained for much longer. Furthermore, accumulating landing probabilities the way PPR or HK does yields lower Micro-F1 accuracy than that of the single best step. On the other hand, by smartly combining the first 15 steps that are of lower quality, AdaDIF surpasses the 
Micro-F1 scores of the longer walks. Interestingly, the Macro-F1 metric for the \texttt{Citeseer} behaves differently than the Micro-F1, and quickly decreases after $\sim 25$ steps. The disagreement between the two metrics can be explained as the diffusions of one or more of the larger classes gradually ``overwhelms'' those of one or more smaller classes,  thus lowering the Macro-F1 score, since the latter is a metric that averages \emph{per-class}. In contrast, the Micro-F1 metric averages \emph{per-node} and takes much less of an impact if a few nodes from the smaller 
classes are mislabeled. Finally, for the \texttt{PubMed} graph (right two plots), steps in the range $[20,40]$ yield \emph{consistently} high accuracy both in terms of Micro- as well as Macro-averaged F1-score. 
Since HK and mostly PPR largely accumulate steps in that range, it seems reasonable that both fixed diffusions are fairly accurate in the \texttt{PubMed} graph. }

\subsection{Tests on simulated label-corruption setup }

Here we outline experimental results performed to evaluate the performance of different diffusion-based classifiers in the presence of anomalous nodes. The main goal is to evaluate whether r-AdaDIF (Algorithm \ref{alg:AdaDIF-rob}) yields improved performance over AdaDIF, HK and PPR, as well as the ability of r-AdaDIF to detect anomalous nodes.  { We also tested a different type of rounding from class-diffusions to class labels that was shown in \cite{gleich_robust} to be robust in the presence of erroneous labels on a graph constructed by images of handwritten digits. The idea is to first normalize diffusions with node degrees, sort each diffusion vector, and assign to each node the class for which the corresponding rank is higher. We applied this type of rounding on PPR diffusions (denoted as PPR w. ranking).  }  Since a ground truth set of anomalous nodes is not available in real graphs, we chose to infuse the true labels with artificial anomalies generated by the following simulated label corruption process: Go through $\mathbf{y}_{\mathcal{L}}$ and for each entry $\left[\mathbf{y}_{\mathcal{L}}\right]_i=c$ draw with probability $p_{\text{cor}}$ a label $c'\sim \mathrm{Unif}\{\mathcal{Y}\setminus c\}$; and replace $\left[\mathbf{y}_{\mathcal{L}}\right]_i\leftarrow {c}^\prime$. In other words, anomalies are created by corrupting some of the true labels by randomly and uniformly ``flipping'' them to a different label. Increasing the corruption probability $p_{\text{cor}}$ of the training labels $\mathbf{y}_{\mathcal{L}}$ is expected to have increasingly negative impact on classification accuracy over $\mathbf{y}_{\mathcal{U}}$. Indeed, as depicted in Fig. \ref{fig:cora_rob_micro}, the accuracy of all diffusion-based classifiers on \texttt{Cora} graph degrades as    
$p_{\text{cor}}$ increases. All diffusions were run for $K=50$, while for r-AdaDIF we found  $\lambda_o=14.6\times 10^{-3}$ and $\lambda_\theta=67.5\times 10^{-5}$ to perform well for moderate values of $p_{\text{cor}}$. Results were averaged over $50$ Monte Carlo experiments, and for each experiment $5\%$ of the nodes were sampled uniformly at random. While tuning $\lambda_o$ for a specific $p_{\text{cor}}$ generally yields improved results, we use the same $\lambda_o$ across the range of $p_{\text{cor}}$ values, since the true value of the latter is generally not available in practice. In this setup, r-AdaDIF demonstrates higher accuracy compared to non-robust classifiers. Moreover, the performance gap increases as more labels become corrupted, until it reaches a ``break point'' at $p_{\rm cor}\approx 0.35$. Interestingly, r-AdaDIF performs worse in the absence of anomalies ($p_{\text{cor}}=0$) that can be attributed to the fact that it only removes useful samples and thus reduces the training set.  {Although PPR w. ranking displays  relative robustness as $p_{cor}$ increases, overall it performs worse than PPR with value based rounding, at least on the \texttt{Cora}  graph.}

 As mentioned earlier, the performance of r-AdaDIF in terms of outlier detection depends on parameter $\lambda_o$. Specifically, for $\lambda_o\rightarrow 0$ the regularizer in \eqref{QC_prob3} is effectively removed and all samples are characterized as outliers. On the other hand, for $\lambda_o\gg 1$ \eqref{QC_prob3} yields $\hat{\mathbf{O}}=\left[\mathbf{0},\ldots,\mathbf{0}\right]$, meaning that no outliers are unveiled. For intermediate values of $\lambda_o$, r-AdaDIF trades off falsely identifying nominal samples as outliers (false alarm) with correctly identifying anomalies (correct detection). This tradeoff of r-AdaDIF's anomaly detection behavior was experimentally evaluated over $50$ Monte Carlo runs by sweeping over a large range of values of $\lambda_o$, and for different values of $p_{\text{cor}}$; see the probability of detection ($p_{\textsc{d}}$) versus probability of false alarms ($p_{\textsc{fa}}$) depicted in Fig. \ref{fig:roc}. Evidently, r-AdaDIF performs much better than a random guess (``coin toss'') detector whose curve is given by the grey dotted line, while the detection performance improves as the corruption rate decreases.

\section{Conclusions}
\label{sec:conclusions}
		The present work, introduces a principled, data-efficient approach to learning class-specific diffusion  functions 
			tailored for the underlying network topology. 	Experiments on real networks confirm that adapting the diffusion function to the given graph and observed labels, significantly improves the performance over fixed diffusions; reaching -- and many times surpassing -- the classification accuracy of computationally heavier state-of-the-art competing methods.
				
	Emerging from this work are many exciting directions  to explore. First, one can investigate different cost functions with respect to which the diffusions are  adapted, e.g., by taking into account robustness of the resulting classifier in the presence of adversarial data. Furthermore, it is worth investigating the space of nonlinear functions of the landing probabilities to determine the degree to which accuracy can be boosted further. Last but not least, it will be interesting to develop adaptive diffusion methods, where learning and adaptation are performed \emph{on-the-fly}, without any memory and computational overhead.   

\appendix

\subsection{Proof of Proposition 1}
For $\lambda\rightarrow\infty$, the effect of  $\ell(\cdot)$ in \eqref{QC_prob} vanishes, and the optimization problem becomes equivalent to solving
\begin{align}\label{smooth_only}
\min_{\boldsymbol{\theta}\in\mathcal{S}^K}\boldsymbol{\theta}^\mathsf{T}\mathbf{A}\boldsymbol{\theta}
\end{align}
where $\mathbf{A}:=(\mathbf{P}^{(K)}_c)^\mathsf{T} \mathbf{D}^{-1} \mathbf{L} \mathbf{D}^{-1}\mathbf{P}^{(K)}_c$ has $(i,j)$ entry  given by $A_{ij}=(\mathbf{p}^{(i)}_c)^\mathsf{T} \mathbf{D}^{-1} \mathbf{L} \mathbf{D}^{-1}\mathbf{p}^{(j)}_c$; and $\mathbf{p}^{(K)}_c$ is the vector of $K$-step landing probabilities with initial distribution $\mathbf{v}_c$ and transition matrix $\mathbf{H}=\sum_{n=1}^N\lambda_{n}\mathbf{u}_n\mathbf{v}_n^\mathsf{T}$, where $\lambda_1> \lambda_2> \cdots >\lambda_N$ are its eigenvalues.  Since $\mathbf{H}$ is a column-stochastic  transition probability matrix, it holds that $\lambda_1=1$, $\mathbf{v}=\mathbf{1}$, and $\mathbf{u}_1=\boldsymbol{\pi}$, where $\boldsymbol{\pi}=\lim_{k\rightarrow\infty}\mathbf{p}^{(k)}_c$ is the steady-state distribution that can be also expressed as $\boldsymbol{\pi}=\mathbf{d}/(2|\mathcal{E}|)$~\cite{levin2017markov}. The landing probability vector for class $c$ is thus
\begin{align}\nonumber
\mathbf{p}^{(K)}_c &=\mathbf{H}^K\mathbf{v}_c= \left[\frac{1}{2|\mathcal{E}|}\mathbf{d}\mathbf{1}^\mathsf{T} + \sum_{n=2}^{N}\lambda_{n}^K\mathbf{u}_n\mathbf{v}_n^\mathsf{T}\right]\mathbf{v}_c\\\label{1}
& = \frac{1}{2|\mathcal{E}|} \mathbf{d} +  \sum_{n=2}^{N}\lambda_{n}^K\mathbf{u}_n\gamma_n \approx \frac{1}{2|\mathcal{E}|} \mathbf{d} + \lambda_{2}^K\mathbf{u}_2\gamma_2
\end{align}
where $\gamma_n:=\mathbf{v}_n^\mathsf{T}\mathbf{v}_c$, and the approximation in \eqref{1} holds because $\lambda_2^K \gg \lambda_n^K$, for $n\in[3,N]$, and $K$ large enough but finite.
Using \eqref{1}, $A_{ij}$ can be rewritten as 
\begin{align}\nonumber
A_{ij}  =& \left[\frac{1}{2|\mathcal{E}|} \mathbf{d}^\mathsf{T} + \lambda_{2}^i\mathbf{u}^\mathsf{T}_2\gamma_2\right]\mathbf{D}^{-1}\mathbf{L}\mathbf{D}^{-1}\left[\frac{1}{2|\mathcal{E}|} \mathbf{d} + \lambda_{2}^j\mathbf{u}_2\gamma_2\right]\\\nonumber
=&\left[\frac{1}{2|\mathcal{E}|} \mathbf{1}^\mathsf{T} + \lambda_{2}^i\mathbf{u}^\mathsf{T}_2\mathbf{D}^{-1}\gamma_2\right]\mathbf{L}\left[\frac{1}{2|\mathcal{E}|} \mathbf{1} + \lambda_{2}^j\mathbf{D}^{-1}\mathbf{u}_2\gamma_2\right]  \\\nonumber
=& \frac{1}{4|\mathcal{E}|^2}\mathbf{1}^\mathsf{T}\mathbf{L}\mathbf{1} + \frac{\lambda_2^{i}\gamma_2}{2|\mathcal{E}|}\mathbf{u}^\mathsf{T}_2\mathbf{D}^{-1}\mathbf{L}\mathbf{1} +\frac{\lambda_2^{j}\gamma_2}{2|\mathcal{E}|}\mathbf{1}^\mathsf{T}\mathbf{L}\mathbf{D}^{-1}\mathbf{u}_2\\\nonumber
&+ \gamma_2^2 \lambda_2^{i+j} \mathbf{u}^\mathsf{T}_2\mathbf{D}^{-1}\mathbf{L}\mathbf{D}^{-1}\mathbf{u}_2\\\label{A_ij}
=& C \lambda_2^{i+j}
\end{align}
where $C:=\gamma_2^2 \mathbf{u}^\mathsf{T}_2\mathbf{D}^{-1}\mathbf{L}\mathbf{D}^{-1}\mathbf{u}_2$, the second equality uses $\mathbf{D}^{-1}\mathbf{d}=\mathbf{1}$, and the last equality follows because $\mathbf{L}\mathbf{1}=\mathbf{0}$. Using  \eqref{A_ij}, one obtains $\mathbf{A}=C\boldsymbol{\lambda}_2\boldsymbol{\lambda}_2^\mathsf{T}$, where $\boldsymbol{\lambda}_2:=
\begin{bmatrix}
\lambda_2 & \lambda_2^2 & \cdots & \lambda_2^K
\end{bmatrix}
^\mathsf{T}$, while \eqref{smooth_only} reduces to 
\begin{align}\label{equiv}
\min_{\boldsymbol{\theta}\in\mathcal{S}^K}\left(\boldsymbol{\lambda}_2^\mathsf{T}\boldsymbol{\theta}\right)^2.
\end{align}
Since $\boldsymbol{\lambda}_2^\mathsf{T}\boldsymbol{\theta}>0~~ \forall \boldsymbol{\theta}\in\mathcal{S}^K$, it can be shown that the KKT optimality conditions for \eqref{equiv} are identical to those of 
\begin{align}\label{equiv2}
\min_{\boldsymbol{\theta}\in\mathcal{S}^K} \boldsymbol{\lambda}_2^\mathsf{T}\boldsymbol{\theta}.
\end{align}
Therefore, \eqref{equiv} admits minimizer(s) identical to \eqref{equiv2}. Finally, we will  show that the minimizer of \eqref{equiv2} is $\mathbf{e}_K$. Since the problem is convex, it suffices to show that $\nabla_{\boldsymbol{\theta}}^\mathsf{T}(\boldsymbol{\lambda}_2^\mathsf{T}\boldsymbol{\theta})_{\boldsymbol{\theta}=\mathbf{e}_K}\left( \boldsymbol{\theta} - \mathbf{e}_K \right)\geq 0 ~~\forall \boldsymbol{\theta}\in\mathcal{S}^K$, or, equivalently
\begin{align*}
\boldsymbol{\lambda}_2^\mathsf{T}\left( \boldsymbol{\theta} - \mathbf{e}_K \right)\geq 0 &\Leftrightarrow \sum_{k=1}^{K}\theta_k\lambda_2^k - \lambda_2^K \geq 0 \\
& \Leftrightarrow \sum_{k=1}^{K}\theta_k\lambda_2^{k-K} \geq 1\\
& \Leftrightarrow \sum_{k=1}^{K}\theta_k\lambda_2^{k-K} \geq \sum_{k=1}^{K}\theta_k \\
& \Leftrightarrow \sum_{k=1}^{K}\theta_k\left(\lambda_2^{k-K}-1\right) \geq 0
\end{align*}
which holds since $\boldsymbol{\theta}\geq \mathbf{0}$ and $\lambda_2^{k-K}\geq 1~~\forall k\in [1,K]$, and completes the proof of the proposition.

\subsection{Proof of Theorem 1}
We need to find the smallest integer $K$ such that 
$\max_{\boldsymbol{\theta}\in\mathcal{S}^K}	\lVert \mathbf{y} - \check{\mathbf{y}}  \rVert \leq \gamma.$
We have
\begin{align}
&\lVert \mathbf{y} - \check{\mathbf{y}}  \rVert  = \lVert  \mathbf{X}_+\boldsymbol{\theta} - \mathbf{X}_-\boldsymbol{\theta}  - \check{\mathbf{X}}_+\boldsymbol{\theta} + \check{\mathbf{X}}_-\boldsymbol{\theta} \rVert \leq \nonumber \\
&\quad  \leq \lVert \theta_K\mathbf{p}_+^{(K)} - \theta_K\mathbf{p}_-^{(K)}\rVert + \lVert \theta_K\mathbf{p}_+^{(K+1)} - \theta_K\mathbf{p}_-^{(K+1)}  \rVert \nonumber \\
&\quad \leq \lVert\mathbf{H}^K\mathbf{p}_+ - \mathbf{H}^K\mathbf{p}_-\rVert+\lVert\mathbf{H}^{K+1}\mathbf{p}_+ - \mathbf{H}^{K+1}\mathbf{p}_-\rVert
\label{eq:th0}
\end{align}
since $\boldsymbol{\theta}\in\mathcal{S}^K$. 
Therefore, to determine an upper bound for the $\gamma$-distinguishability threshold it suffices to find the smallest integer $K$ for which \eqref{eq:th0} is upper bounded by $\gamma$.

Let $\mathbf{q}_1, \dots, \mathbf{q}_N$ be the eigenvectors corresponding to the eigenvalues $0=\mu_1< \mu_2\leq \cdots \leq \mu_N<2$ of the normalized Laplacian $\tilde{\mathbf{L}} $.  The transition probability matrix is then 
\begin{equation}
\mathbf{H} =  \mathbf{D}^{\frac{1}{2}}(\mathbf{I}-\tilde{\mathbf{L}} ) \mathbf{D}^{-\frac{1}{2}}. 
\end{equation}

For the first term of the RHS of \eqref{eq:th0}, we have 
\begin{align}
&\|\mathbf{H}^K\mathbf{p}_+ - \mathbf{H}^K\mathbf{p}_-\|  \leq \|\mathbf{H}^K\mathbf{p}_+ - \boldsymbol{\pi} \| + \| \mathbf{H}^K\mathbf{p}_- - \boldsymbol{\pi} \|  \nonumber \\
& \qquad=   \| \mathbf{D}^{\frac{1}{2}}(\mathbf{I}-\tilde{\mathbf{L}} )^K\mathbf{D}^{-\frac{1}{2}}\mathbf{p}_+ - \frac{\mathbf{D}\boldsymbol{1}}{2|\mathcal{E}|} \|\nonumber \\ 
&\qquad + \| \mathbf{D}^{\frac{1}{2}}(\mathbf{I}-\tilde{\mathbf{L}} )^K\mathbf{D}^{-\frac{1}{2}}\mathbf{p}_- -  \frac{\mathbf{D}\boldsymbol{1}}{2|\mathcal{E}|}\| .  \label{eq:th1} 
\end{align}
Since  $\mathbf{q}_1 = \frac{\mathbf{D}^{\frac{1}{2}}\mathbf{1}}{\sqrt{2|\mathcal{E}|}}$~\cite{levin2017markov}, we 
have for $c\in\{+,-\}$ that 
\begin{align}
\mathbf{D}^{\frac{1}{2}}\mathbf{q}_1\langle\mathbf{q}_1, \mathbf{D}^{-\frac{1}{2}}\mathbf{p}_c\rangle & =
\mathbf{D}^{\frac{1}{2}}\frac{\mathbf{D}^{\frac{1}{2}}\mathbf{1}}{\sqrt{2|\mathcal{E}|}}\left\langle\frac{\mathbf{D}^{\frac{1}{2}}\mathbf{1}}{\sqrt{2|\mathcal{E}|}}, \mathbf{D}^{-\frac{1}{2}}\mathbf{p}_c\right\rangle \nonumber \\
& = \frac{\mathbf{D}\mathbf{1}}{\sqrt{2|\mathcal{E}|}}
\frac{\langle\mathbf{1},\mathbf{p}_c\rangle}{\sqrt{2|\mathcal{E}|}} = \frac{\mathbf{D}\mathbf{1}}{2|\mathcal{E}|} \;. \label{eq:th1_2}
\end{align}
Upon defining $\mathbf{M} := (\mathbf{I}-\tilde{\mathbf{L}} )^K- \mathbf{q}_1\mathbf{q}_1^\mathsf{T}$, and taking into account \eqref{eq:th1_2}, inequality  \eqref{eq:th1} can be written as 
\begin{align}
&\|\mathbf{H}^K\mathbf{p}_+ - \mathbf{H}^K\mathbf{p}_-\|  \nonumber \\
&\qquad \leq \| \mathbf{D}^{\frac{1}{2}} \|  \| \mathbf{M} \|  \left(\| \mathbf{D}^{-\frac{1}{2}}\mathbf{p}_+ \| + \| \mathbf{D}^{-\frac{1}{2}}\mathbf{p}_- \| \right).  \label{eq:step2}
\end{align} 	
The factors in \eqref{eq:step2} can be bounded as 
\begin{align}
\| \mathbf{D}^{-\frac{1}{2}}\mathbf{p}_+ \| & = \sqrt{\sum_{i \in \mathcal{L}_+} \left( \frac{1}{|\mathcal{L}_+|}d_i^{-\frac{1}{2}} \right)^2 } \nonumber \\
&= \sqrt{\sum_{i \in \mathcal{L}_+} \frac{1}{|\mathcal{L}_+|^2}d_i^{-1}}  \leq  \frac{1}{\sqrt{d_{\min_+}|\mathcal{L}_+|}},	  \\
\| \mathbf{D}^{-\frac{1}{2}}\mathbf{p}_- \| & = \sqrt{\sum_{i \in \mathcal{L}_-} \frac{1}{|\mathcal{L}_-|^2}d_i^{-1}} \leq  \frac{1}{\sqrt{d_{\min_-}|\mathcal{L}_-|}},  \\
\| \mathbf{M} \| &= \sup_{\mathbf{v}}\frac{\langle\mathbf{Mv},\mathbf{v}\rangle}{\mathbf{Mv}} = \max_{i\neq 1}|1-\mu_i|^K, \label{eq:spectralM} \\
\| \mathbf{D}^{\frac{1}{2}} \| &= \sqrt{d_{\max}}
\end{align}
where \eqref{eq:spectralM} follows from the properties of the normalized Laplacian.  Therefore, 	\eqref{eq:step2} becomes
\begin{align}
\| \mathbf{H}^K \mathbf{p}_+ - \mathbf{H}^K \mathbf{p}_-\|  \leq & \left(  \frac{1}{\sqrt{d_{\min_-}|\mathcal{L}_-|}}  +  \frac{1}{\sqrt{d_{\min_+}|\mathcal{L}_+|}}   \right) \nonumber \\ 
& \cdot \max_{i\neq 1}|1-\mu_i|^K \cdot \sqrt{d_{\max}}  .
\end{align} 
Letting $\mu' := \min\{\mu_2,2-\mu_N\}$, and using the fact that 
\begin{align}
(1-\mu')^K\leq e^{-K\mu'}
\end{align} we obtain 
\begin{align}
&\| \mathbf{H}^K \mathbf{p}_+ - \mathbf{H}^K \mathbf{p}_-\|  \nonumber \\
&\ \ \ \	\leq \left(  \sqrt{\frac{d_{\max}}{d_{\min_-}|\mathcal{L}_-|}}  +  \sqrt{\frac{d_{\max}}{d_{\min_+}|\mathcal{L}_+|}}   \right)  e^{-K\mu'}.
\label{eq:Aterm}
\end{align} 
Likewise, we can bound the second term in \eqref{eq:th0} as  
\begin{align}
&\| \mathbf{H}^{K+1} \mathbf{p}_+ - \mathbf{H}^{K+1} \mathbf{p}_-\|   \nonumber \\
&\ \ \ \ \ \leq \left(  \sqrt{\frac{d_{\max}}{d_{\min_-}|\mathcal{L}_-|}}  +  \sqrt{\frac{d_{\max}}{d_{\min_+}|\mathcal{L}_+|}}   \right)  e^{-(K+1)\mu'}.
\label{eq:Bterm}
\end{align}
In addition, we note that for all $\mu'>0, K\in\mathbb{Z}$ it holds that
\begin{align}
e^{-K\mu'} + e^{-(K+1)\mu'} < 2e^{-K\mu'}.
\label{eq:fact}
\end{align}
Upon substituting \eqref{eq:Aterm} and \eqref{eq:Bterm} into \eqref{eq:th0}, and also using \eqref{eq:fact}, we arrive at
\begin{equation}
\lVert \mathbf{y} - \check{\mathbf{y}}  \rVert  \leq 2\left(  \sqrt{\frac{d_{\max}}{d_{\min_-}|\mathcal{L}_-|}}  +  \sqrt{\frac{d_{\max}}{d_{\min_+}|\mathcal{L}_+|}}   \right)  e^{-K\mu'}.
\label{eq:simplified}
\end{equation}
To determine an upper bound on the $\gamma$-distinguishability threshold, it suffices to find the smallest integer $K$ for which \eqref{eq:simplified} becomes less than $\gamma$; that is, 
\begin{align}
& 2\left(  \sqrt{\frac{d_{\max}}{d_{\min_-}|\mathcal{L}_-|}}  +  \sqrt{\frac{d_{\max}}{d_{\min_+}|\mathcal{L}_+|}}   \right)  e^{-K\mu'} \leq \gamma.
\label{eq:ineq}
\end{align} 
Multiplying both sides of \eqref{eq:ineq} by the positive number $e^{K\mu'}/\gamma$, and taking logarithms yields 
\begin{displaymath}
\log \left[ \tfrac{2\sqrt{d_{\max}}}{\gamma}  \left(\sqrt{\tfrac{1}{d_{\min_-}|\mathcal{L}_-|}}  +  \sqrt{\tfrac{1}{d_{\min_+}|\mathcal{L}_+|}} \right)   \right] \leq K\mu'.
\end{displaymath}
Therefore, using as landing probabilities
\begin{displaymath}
\left\lceil\frac{1}{\mu'} \log \left[ \tfrac{2\sqrt{d_{\max}}}{\gamma}  \left(\sqrt{\tfrac{1}{d_{\min_-}|\mathcal{L}_-|}}  +  \sqrt{\tfrac{1}{d_{\min_+}|\mathcal{L}_+|}} \right)   \right]\right\rceil
\end{displaymath}
the $\ell_2$ distance between any two diffusion-based classifiers will be at most $\gamma$; and the proof is complete.
\subsection{Bound for PageRank}
\label{ap:PRHR}

Substituting PageRank's diffusion coefficients in the proof of Theorem 1, inequality \eqref{eq:ineq} becomes
\begin{equation}
2 (1-\alpha)\alpha^K\left(  \sqrt{\frac{d_{\max}}{d_{\min_-}|\mathcal{L}_-|}}  +  \sqrt{\frac{d_{\max}}{d_{\min_+}|\mathcal{L}_+|}}   \right)  e^{-K\mu'} \leq \gamma. 
\nonumber 
\label{eq:prop2_3}
\end{equation} 
Multiplying both sides by the positive number $e^{K\mu'}\alpha^{-K}/\gamma$ and taking logarithms yields	
\begin{displaymath}
\log \left[ \tfrac{2\sqrt{d_{\max}}}{\gamma/(1-\alpha)}  \left(\sqrt{\tfrac{1}{d_{\min_-}|\mathcal{L}_-|}}  +  \sqrt{\tfrac{1}{d_{\min_+}|\mathcal{L}_+|}} \right)   \right] \leq K(\mu'-\log\alpha)
\end{displaymath}
which results in the $\gamma$-distinguishability threshold bound
\begin{align}
K^{\mathrm{PR}}_\gamma & \leq
\tfrac{1}{\mu'-\log \alpha} \log \left[ \tfrac{2\sqrt{d_{\max}}}{\gamma/(1-\alpha)}  \left(\sqrt{\tfrac{1}{d_{\min_-}|\mathcal{L}_-|}}  +  \sqrt{\tfrac{1}{d_{\min_+}|\mathcal{L}_+|}} \right)   \right]. \nonumber
\end{align}




\begin{thebibliography}{10}
	\providecommand{\url}[1]{#1}
	\csname url@samestyle\endcsname
	\providecommand{\newblock}{\relax}
	\providecommand{\bibinfo}[2]{#2}
	\providecommand{\BIBentrySTDinterwordspacing}{\spaceskip=0pt\relax}
	\providecommand{\BIBentryALTinterwordstretchfactor}{4}
	\providecommand{\BIBentryALTinterwordspacing}{\spaceskip=\fontdimen2\font plus
		\BIBentryALTinterwordstretchfactor\fontdimen3\font minus
		\fontdimen4\font\relax}
	\providecommand{\BIBforeignlanguage}[2]{{%
			\expandafter\ifx\csname l@#1\endcsname\relax
			\typeout{** WARNING: IEEEtran.bst: No hyphenation pattern has been}%
			\typeout{** loaded for the language `#1'. Using the pattern for}%
			\typeout{** the default language instead.}%
			\else
			\language=\csname l@#1\endcsname
			\fi
			#2}}
	\providecommand{\BIBdecl}{\relax}
	\BIBdecl
		
\bibitem{argyriou2006combining}
	A.~Argyriou, M.~Herbster, and M.~Pontil, ``Combining graph laplacians for
	semi--supervised learning,'' in \emph{Proc. Advances in Neural Information
		Processing Systems}, Vancouver, Can., 2006, pp. 67--74.
	
\bibitem{atwood2016diffusion}
J.~Atwood and D.~Towsley, ``Diffusion-convolutional neural networks,'' in
\emph{ Proc. Advances in Neural Information Processing Systems}, Barcelona, Spain, 2016, pp.
1993--2001.
	
\bibitem{avrachenkov2012generalized}
K.~Avrachenkov, A.~Mishenin, P.~Gon{\c{c}}alves, and M.~Sokol, ``Generalized
optimization framework for graph-based semi-supervised learning,''
\emph{Proc. SIAM Int. Conf. on Data
	Mining}, Anaheim, CA, 2012, pp. 966--974.

\bibitem{FunctionalRankings}
\BIBentryALTinterwordspacing
R.~Baeza-Yates, P.~Boldi, and C.~Castillo, ``Generic damping functions for
propagating importance in link-based ranking,'' \emph{Internet Math.},
vol.~3, no.~4, pp. 445--478, 2006. 
\BIBentrySTDinterwordspacing

\bibitem{belkin2006manifold}
M.~Belkin, P.~Niyogi, and V.~Sindhwani, ``Manifold regularization: A geometric
	framework for learning from labeled and unlabeled examples,'' \emph{J. Mach. Learn. Res.}, no. 7, Nov, 2006, pp. 2399--2434.
	
\bibitem{label-propagation-and-quadratic-criterion}
	Y.~Bengio, O.~Delalleau, and N.~Le~Roux, ``Label propagation and quadratic
	criterion,'' in \emph{Semi-Supervised Learning}. Cambridge, MA, USA: MIT Press, 2006.
	
\bibitem{berberidis2018random}
D.~Berberidis, A.~N. Nikolakopoulos, and G.~B. Giannakis, ``Random walks with
restarts for graph-based classification: Teleportation tuning and sampling
design,'' in \emph{Proc. of IEEE Int. Conf. on Acoustics, Speech
	and Signal Processing}, Calgary, Can., April 2018.

\bibitem{AdadifBigData}
D. Berberidis, A. N. Nikolakopoulos, and G. B. Giannakis, "AdaDIF: Adaptive Diffusions for Efficient Semi-supervised Learning over Graphs," Proc. of IEEE Int. Conf. on Big Data, Seattle, Washington, Dec. 10-13, 2018. pp. 92--99.


\bibitem{brin2012reprint}
S.~Brin and L.~Page, ``Reprint of: The anatomy of a large-scale hypertextual
web search engine,'' \emph{Comput. Netw.}, vol.~56, no.~18, pp.
3825--3833, 2012.
	
\bibitem{buchnik2017bootstrapped}
E.~Buchnik and E.~Cohen, ``Bootstrapped graph diffusions: Exposing the power of
nonlinearity,'' \emph{arXiv preprint arXiv:1703.02618}, 2017.
	
\bibitem{SSL}
	O.~Chapelle, B.~Sch\"{o}lkopf, and A.~Zien, \emph{Semi-Supervised Learning}. Cambridge, MA, USA: MIT Press, 2006.
	
\bibitem{Kovac2014semi}
S.~Chen, F.~Cerda, P.~Rizzo, J.~Bielak, J.~H. Garrett, and J.~Kovacevic,
``Semi-supervised multiresolution classification using adaptive graph
filtering with application to indirect bridge structural health monitoring,''
\emph{IEEE Trans. Signal Process.}, vol.~62, no.~11, pp.
2879--2893, June 2014.


\bibitem{constantine2009}
\BIBentryALTinterwordspacing
P.~G. Constantine and D.~F. Gleich, ``Random alpha pagerank,'' \emph{Internet
	Math.}, vol.~6, no.~2, pp. 189--236, 2009. 
\BIBentrySTDinterwordspacing

\bibitem{leus2017filter}
\BIBentrySTDinterwordspacing
M. Contino, E. Isufi and G. Leus,``Distributed edge-variant graph filters,''
\emph{Proc. Int. Work. on Computational Advances in Multi-Sensor Adaptive Processing}, 
Curacao, Dutch Antilles, Dec. 2017, pp. 1-5.
\BIBentrySTDinterwordspacing

\bibitem{chung2007heat}
F.~Chung, ``The heat kernel as the pagerank of a graph,'' \emph{Proc. Natl. Acad. Sci.}, vol. 104, no.~50, pp. 19\,735--19\,740,
2007.

\bibitem{GleichBeyond}
\BIBentryALTinterwordspacing
D.~F. Gleich, ``Pagerank beyond the web,'' \emph{SIAM Rev.}, vol.~57, no.~3,
pp. 321--363, 2015. 
\BIBentrySTDinterwordspacing

\bibitem{gorski2007biconvex}
\BIBentryALTinterwordspacing
J.~Gorski, F.~Pfeuffer, and K.~Klamroth, ``Biconvex sets and optimization with
biconvex functions: a survey and extensions,'' \emph{Math. Methods of
	Oper. Res.}, vol.~66, no.~3, pp. 373--407, Dec. 2007. 
\BIBentrySTDinterwordspacing

\bibitem{grover2016node2vec}
A.~Grover and J.~Leskovec, ``node2vec: Scalable feature learning for
networks,'' in \emph{Proc. of ACM SIGKDD Int.
	Conf. on Knowledge Discovery and Data Mining}, San Francisco, CA, 2016, pp. 855--864.

\bibitem{joachims2003transductive}
T.~Joachims, ``Transductive learning via spectral graph partitioning,''
\emph{Proc. of Int. Conf. on
	Machine Learn.}, Washington DC, 2003, pp. 290--297.

\bibitem{keka2011sparse}
V.~Kekatos and G.~B. Giannakis, ``From sparse signals to sparse residuals for
robust sensing,'' \emph{IEEE Trans. Signal Process.}, vol.~59,
no.~7, pp. 3355--3368, July 2011.

\bibitem{kipf2016semi}
T.~N. Kipf and M.~Welling, ``Semi-supervised classification with graph
convolutional networks,'' \emph{arXiv preprint arXiv:1609.02907}, 2016.

\bibitem{kloster2014heat}
K.~Kloster and D.~F. Gleich, ``Heat kernel based community detection,'' in
\emph{Proc. of ACM SIGKDD Int. Conf. on
	Knowledge Discovery and Data Mining}, New York, NY,  2014, pp. 1386--1395.

\bibitem{kloumann2017block}
I.~M. Kloumann, J.~Ugander, and J.~Kleinberg, ``Block models and personalized
pagerank,'' \emph{Proc. Natl. Acad. Sci.}, vol. 114,
no.~1, pp. 33--38, 2017.

\bibitem{kondor2002diffusion}
R.~I. Kondor and J.~Lafferty, ``Diffusion kernels on graphs and other discrete
input spaces,'' in \emph{Proc. of Int. Conf. on
	Machine Learning}, Syndey, Australia, 2002, pp. 315--322.

\bibitem{kveton2010semi}
B.~Kveton, M.~Valko, A.~Rahimi, and L.~Huang, ``Semi-supervised learning with
max-margin graph cuts,'' in \emph{Proc. of. Int. Conf. on Artificial Intelligence and
	Statistics}, Sardinia, Italy, 2010, pp. 421--428.

\bibitem{LangvilleMeyer}
\BIBentryALTinterwordspacing
A.~N. Langville and C.~D. Meyer, ``Deeper inside pagerank,'' \emph{Internet
	Math.}, vol.~1, no.~3, pp. 335--380, 2004. 
\BIBentrySTDinterwordspacing

\bibitem{levin2017markov}
D.~A. Levin and Y.~Peres, \emph{Markov Chains and Mixing Times}. New York, NY, USA: Amer. Math. Soc., 2017.

\bibitem{lin2010semi}
F.~Lin and W.~W. Cohen, ``Semi-supervised classification of network data using
very few labels,'' in \emph{Proc. of Int. Conf. on Advances in Social Network 
	Analysis and Mining}, Odense, Denmark, 2010, pp.
192--199.

\bibitem{liu2012robust}
	W.~Liu, J.~Wang, and S.-F. Chang, ``Robust and scalable graph-based
	semisupervised learning,'' \emph{Proc. of the IEEE}, vol. 100, no.~9,
	pp. 2624--2638, 2012.

\bibitem{manning2008ir}
C.~D. Manning, P. Raghavan, and H. Schutze, \emph{Introduction to Information Retrieval}.    Cambridge, MA: Cambridge University Press, 2008.
	
	
	
\bibitem{merkurjev2016semi}
	E.~Merkurjev, A.~L. Bertozzi, and F.~Chung, ``A semi-supervised heat kernel
	pagerank MBO algorithm for data classification,'' Univ. of California
	Los Angeles, Los Angeles, US, Tech. Rep., 2016.

\bibitem{Nikolakopoulos:2013:NNR:2433396.2433415}
\BIBentryALTinterwordspacing
A.~N. Nikolakopoulos and J.~D. Garofalakis, ``Ncdawarerank: A novel ranking
method that exploits the decomposable structure of the web,''
\emph{Proc. ACM Int. Conf. on Web Search and
	Data Mining}, Rome, Italy, 2013, pp. 143--152. 
\BIBentrySTDinterwordspacing

\bibitem{7840666}
A.~N. Nikolakopoulos, A.~Korba, and J.~D. Garofalakis, ``Random surfing on
multipartite graphs,'' in \emph{Proc. of IEEE Int. Conf. on Big
	Data}, Washington DC, Dec. 2016, pp. 736--745.
	
\bibitem{perozzi2014deepwalk}
B.~Perozzi, R.~Al-Rfou, and S.~Skiena, ``Deepwalk: Online learning of social
representations,'' \emph{Proc. ACM SIGKDD Int.
	Conf. on Knowl. Disc. and Data Mining}, New York, NY, 2014, pp. 701--710.	

\bibitem{puig2011multidimensional}
A.~T. Puig, A.~Wiesel, G.~Fleury, and A.~O. Hero, ``Multidimensional
shrinkage-thresholding operator and group lasso penalties,'' \emph{IEEE Signal Process. Lett.}, vol.~18, no.~6, pp. 363--366, 2011.	
	
\bibitem{rosenfeld2017semi}
	N.~Rosenfeld and A.~Globerson, ``Semi-supervised learning with competitive
	infection models,'' \emph{arXiv preprint arXiv:1703.06426}, 2017.
	
\bibitem{sandryhaila2013discrete}
	A.~Sandryhaila and J.~M.~F. Moura, ``Discrete signal processing on graphs,''
	\emph{IEEE Trans. Signal Process.}, vol.~61, no.~7, pp.
	1644--1656, April 2013.


	
\bibitem{segarra2017filter}
\BIBentrySTDinterwordspacing
S. Segarra, A. Marques, and A. Ribeiro,``Optimal graph-filter
design and applications to distributed linear network operators,''
\emph{IEEE Trans. on Signal Process.}, 
vol. 65, no. 15, pp. 4117--4131, August 2017.
\BIBentrySTDinterwordspacing



\bibitem{talukdar2009new}
P.~P. Talukdar and K.~Crammer, ``New regularized algorithms for transductive
learning,'' in \emph{Proc. of Joint Eur. Conf. on Machine Learning
	and Knowledge Discovery in Databases}, 2009, pp. 442--457.
	
\bibitem{ugander2013balanced}
J.~Ugander and L.~Backstrom, ``Balanced label propagation for partitioning
massive graphs,'' in \emph{Proc. of ACM Int. Conf. on Web Search and Data
	Mining}, Rome, Italy, 2013, pp. 507--516.

\bibitem{wu2012learning}
	X.-M. Wu, Z.~Li, A.~M. So, J.~Wright, and S.-F. Chang, ``Learning with
	partially absorbing random walks,'' \emph{Proc. Adv. in Neural Inform.
		Proc. Systems}, Lake Tahoe, CA, Dec. 2012, pp. 3077--3085.
	
	
\bibitem{yang2016revisiting}
Z.~Yang, W.~W. Cohen, and R.~Salakhutdinov, ``Revisiting semi-supervised
learning with graph embeddings,'' \emph{arXiv preprint arXiv:1603.08861},
2016.
	
\bibitem{zhu2003semi}
X.~Zhu, Z.~Ghahramani, and J.~Lafferty, ``Semi-supervised learning using
{G}aussian fields and harmonic functions,'' in \emph{Proc. of Int. Conf. on
	Machine Learning}, Washington DC, Aug. 2003.

\bibitem{gleich_robust}
D.~F. Gleich, and M.~W. Mahoney, ``Using Local Spectral Methods to Robustify
Graph-Based Learning Algorithms,'' in \emph{Proc. of the Int. Conf. on Knowl. Disc. and Data Mining}, Sidney Australia, Aug. 2015.

\bibitem{losp}
K.~ He, P.~ Shi, J.~E. Hopcroft, and D.~ Bindel, ``Local Spectral Diffusion for Robust Community Detection,'' in \emph{Proc. of the SIGKDD workshop}, San Francisco CA, Aug. 2016.

\bibitem{aptrank}
B.~ Jiang, K.~ Kloster, D.~F. Gleich, and M.~ Gribskov, ``AptRank: an adaptive PageRank model for protein function prediction on bi-relational graphs,'' in \emph{Bioinformatics}, vol. 33, no. 12, pp. 1829--1836, Aug. 2017.

\bibitem{lemon}
K.~ He, Y.~ Sun, D.~ Bindel, J.~E. Hopcroft, and Y.~Li, ``Detecting overlapping communities from local spectral subspaces,'' in \emph{Proc. of the Int. Conf. on Data Mining }, Atlantic City NJ, Aug. 2015, pp. 769--774


\end{thebibliography}
\end{document}